%% file: main.tex
\documentclass[10pt, twocolumn, letterpaper]{article}

\usepackage[pagenumbers]{wacv}             %

\usepackage{booktabs}
\usepackage{times}
\usepackage{epsfig}
\usepackage{graphicx}
\usepackage{amsmath}
\usepackage{amssymb}
\usepackage{float}
\usepackage{subcaption}
\usepackage{multirow}
\usepackage{bigdelim}
\usepackage{rotating}
\usepackage{caption}
\usepackage{multirow}
\usepackage{tabularx}
\usepackage{array}
\usepackage[dvipsnames]{xcolor}
\usepackage{mathtools}
\usepackage{marvosym}
\usepackage[export]{adjustbox}

\usepackage[pagebackref=true, breaklinks=true, colorlinks=true, bookmarks=false]{hyperref}

\usepackage[capitalize]{cleveref}
\crefname{section}{Sec.}{Secs.}
\Crefname{section}{Section}{Sections}
\Crefname{table}{Table}{Tables}
\crefname{table}{Tab.}{Tabs.}

\def\wacvPaperID{67} %
\def\confName{WACV}
\def\confYear{2024}

\begin{document}

\title{TriPlaneNet: An Encoder for EG3D Inversion}

\author{Ananta R. Bhattarai \and Matthias Nie{\ss}ner \and Artem Sevastopolsky \and \\
Technical University of Munich (TUM)
}
\maketitle

\setlength{\abovedisplayskip}{8pt}
\setlength{\belowdisplayskip}{8pt}

\begin{abstract}
   \input{sections/abstract}
\end{abstract}

\input{sections/01_introduction}
\input{sections/02_related_works}
\input{sections/03_method}
\input{sections/04_experiments}
\input{sections/05_discussion}
\input{sections/06_acknowledgments}

{\small
\bibliographystyle{ieee_fullname}
\bibliography{main}
}

\clearpage
\appendix
\input{supplement}
\clearpage
\end{document}

%% file: sections/abstract.tex
Recent progress in NeRF-based GANs has introduced a number of approaches for high-resolution and high-fidelity generative modeling of human heads with a possibility for novel view rendering. 
At the same time, one must solve an inverse problem to be able to re-render or modify an existing image or video. 
Despite the success of universal optimization-based methods for 2D GAN inversion, those applied to 3D GANs may fail to extrapolate the result onto the novel view, whereas optimization-based 3D GAN inversion methods are time-consuming and 
can require at least several minutes per image. 
Fast encoder-based techniques, such as those developed for StyleGAN, may also be less appealing due to the lack of identity preservation. 
Our work introduces a fast technique that bridges the gap between the two approaches by directly utilizing the tri-plane representation presented for the EG3D generative model.
In particular, we build upon a feed-forward convolutional encoder for the latent code and extend it with a fully-convolutional predictor of tri-plane numerical offsets.
The renderings are similar in quality to the ones produced by optimization-based techniques and 
outperform the ones by encoder-based methods.
As we empirically prove, this is a consequence of directly operating in the tri-plane space, not in the GAN parameter space, while making use of an encoder-based trainable approach.
Finally, we demonstrate significantly more correct embedding of a face image in 3D than for all the baselines, further strengthened by a \emph{probably symmetric prior} enabled during training.

%% file: sections/01_introduction.tex
\section{Introduction}\label{introduction}
In recent years, numerous works~\cite{chan2021pi,chan2022efficient} have tackled the problem of multi-view consistent image synthesis with 3D-aware GANs. 
Such methods make generators aware of a 3D structure by modeling it with explicit voxel grids~\cite{henzler2019escaping, sitzmann2019deepvoxels, nguyen2020blockgan} or neural implicit representations~\cite{chan2021pi,or2022stylesdf}. 
Most notably, EG3D~\cite{chan2022efficient} introduced a 3D GAN framework based on a tri-plane 3D representation that is both efficient and expressive to enable high-resolution 3D-aware image synthesis. Moreover, they demonstrate state-of-the-art results for unconditional geometry-aware image synthesis.

\begin{table}[t!]
\centering
\setlength\tabcolsep{0pt}
\begin{tabular}{cc@{\hskip 0.1cm}ccc}
    \multirow{4}{*}[-5em]{
    \begin{subfigure}[h]{0.11\textwidth}
        \includegraphics[width=\textwidth]{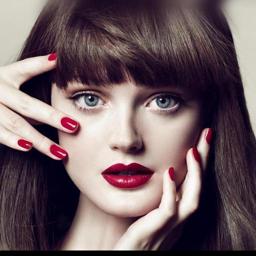}
        \caption*{Input}
    \end{subfigure}
    \hspace{0.1cm}
    } & 
    \raisebox{1\height}{\rotatebox[origin=c]{90}{$\mathcal{W}+$ \cite{karras2020analyzing}} }
    & \includegraphics[width=0.105\textwidth]{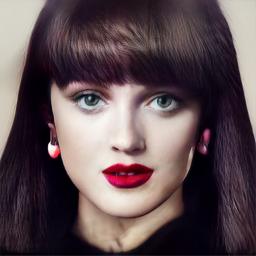}
    & \includegraphics[width=0.105\textwidth]{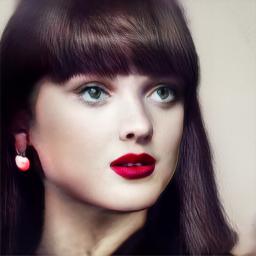}
    & \includegraphics[width=0.105\textwidth]{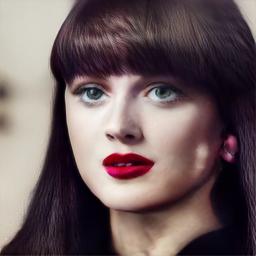} \\
& 
    \raisebox{1.2\height}{\rotatebox[origin=c]{90}{PTI~\cite{roich2022pivotal}}} 
    & \includegraphics[width=0.105\textwidth]{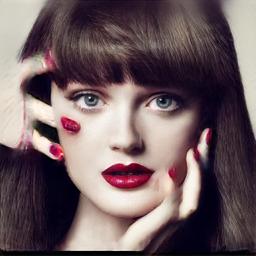}
    & \includegraphics[width=0.105\textwidth]{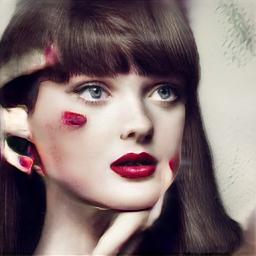}
    & \includegraphics[width=0.105\textwidth]{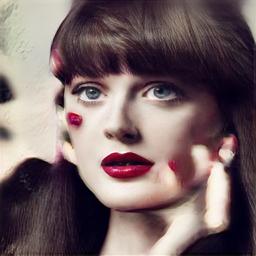} \\
& 
    \raisebox{1.3\height}{\rotatebox[origin=c]{90}{SPI~\cite{yin20223d}}} 
    & \includegraphics[width=0.105\textwidth]{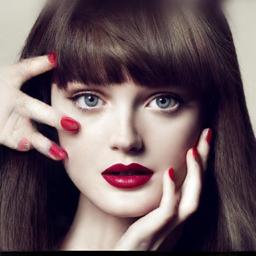}
    & \includegraphics[width=0.105\textwidth]{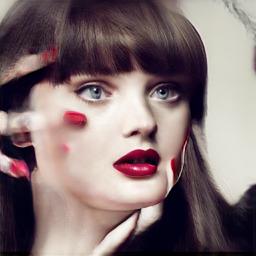}
    & \includegraphics[width=0.105\textwidth]{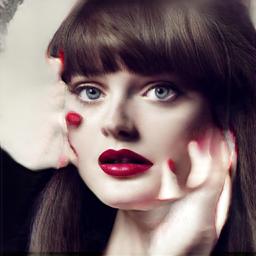} \\
&
    \raisebox{1.2\height}{\rotatebox[origin=c]{90}{pSp~\cite{richardson2021encoding}}} 
    & \includegraphics[width=0.105\textwidth]{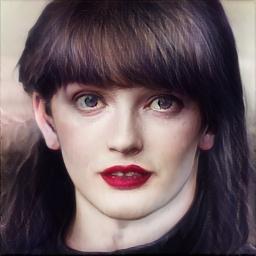}
    & \includegraphics[width=0.105\textwidth]{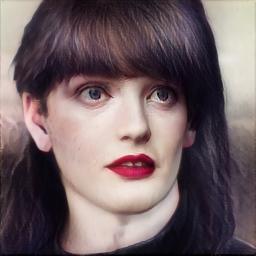}
    & \includegraphics[width=0.105\textwidth]{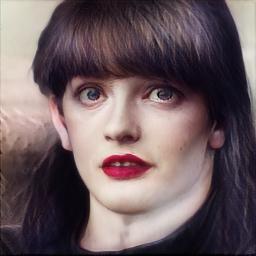} \\
&
    \raisebox{1.8\height}{\rotatebox[origin=c]{90}{\textbf{Ours}}} 
    & \includegraphics[width=0.105\textwidth]{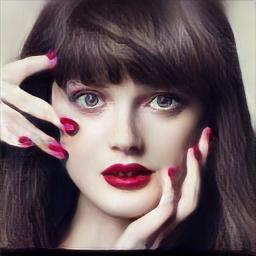}
    & \includegraphics[width=0.105\textwidth]{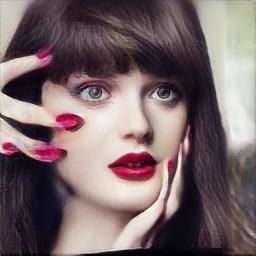}
    & \includegraphics[width=0.105\textwidth]{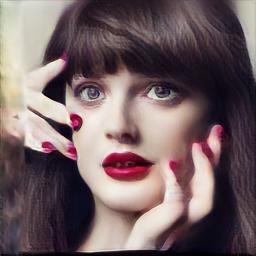} \\
\end{tabular}
\vspace{-0.1cm}
\captionof{figure}{
For a given picture, our method predicts the appropriate latent code and the tri-plane offsets for the EG3D generator in a feed-forward manner. This way, both the frontal view and the novel view rendering can be obtained with high fidelity and in close to real time.
\label{fig:introduction}
}
\vspace{-0.4cm}
\end{table}
\begin{figure*}[t!]
    \centering
    \includegraphics[trim={0 2.5cm 0 0},width=\textwidth]{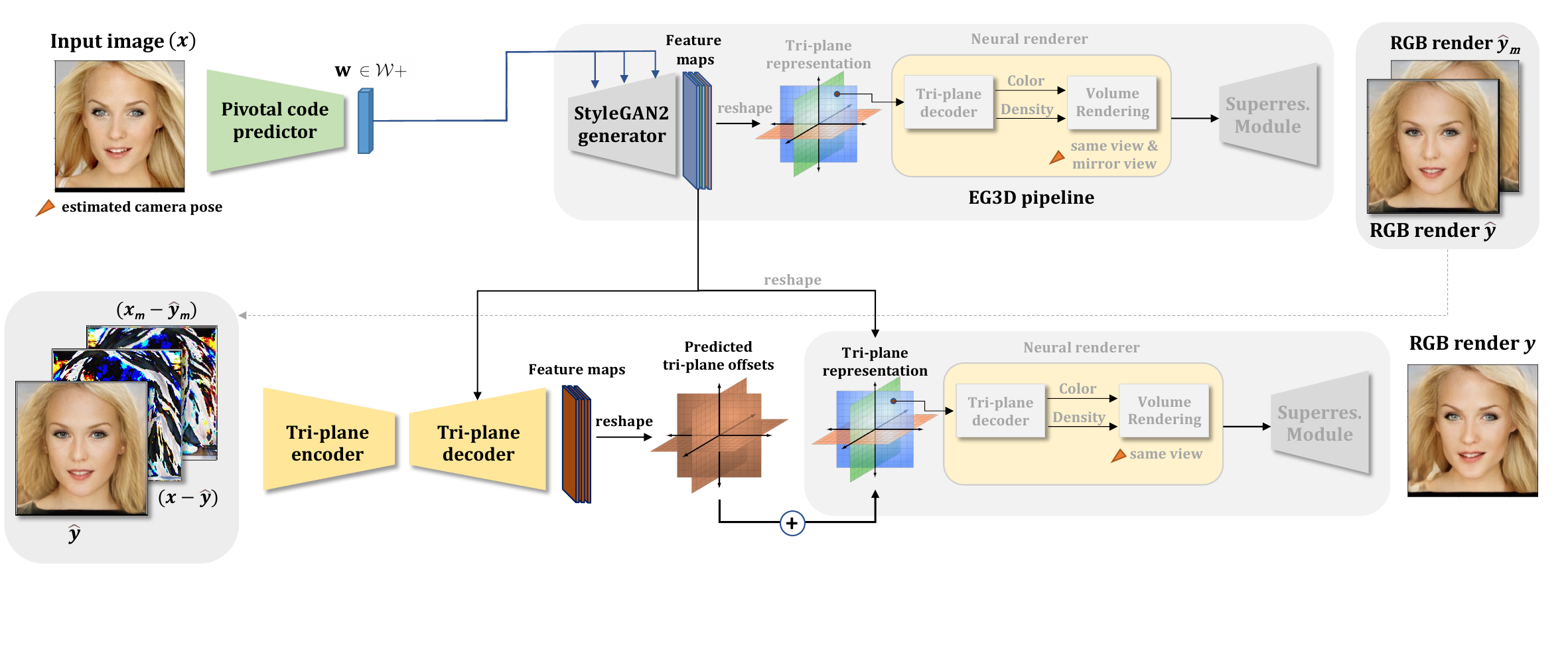}
    \caption{Overview of the proposed method. TriPlaneNet consists of two branches. The first branch (above) comprises the predictor $\hat{w} = \boldsymbol{\phi}(x)$ of the pivotal latent code $\hat{w} \in \mathcal{W}+$, which results in an input-view RGB image $\hat{y} = \mathcal{R}(\boldsymbol{G}(\hat{w}), \pi)$ and a mirror-view RGB image $\hat{y}_m = \mathcal{R}(\boldsymbol{G}(\hat{w}), \pi_m)$ with corresponding camera matrices $\pi$ and $\pi_m$ respectively after passing it through the EG3D tri-plane generator $\boldsymbol{G}(\cdot)$ and renderer block $\mathcal{R}(\cdot, \cdot)$ containing super-resolution module. The second branch (below) uses the first-stage approximation $\hat{y}$ its difference with the target $(x - \hat{y})$, difference between mirror image and rendered mirror view image $(x_m - \hat{y}_m)$ and the tri-planes features $\boldsymbol{G}(\hat{w})$ to predict the numerical offsets to the tri-planes $\Delta \boldsymbol{T}$ by a convolutional autoencoder $\Delta \boldsymbol{T} = \boldsymbol{\psi}(\hat{y}, x - \hat{y})$, which yields the final prediction $y = \mathcal{R}(\boldsymbol{G}(\hat{w}) + \Delta \boldsymbol{T}, \pi)$.}
    \label{method:fig}
\end{figure*}

The main applications of 3D GANs include human face inversion, including head tracking, reenactment, facial manipulation, and novel view synthesis of a given image or video. 
Oftentimes, the classical GAN formulation does not support trivial inversion, i.e.~finding the appropriate code in the learned GAN space for a given sample. A straightforward way to achieve this is by obtaining the latent code of the input image via optimization-based or encoder-based approaches, i.e.~applying 2D GAN inversion techniques. An existing branch of research studies 2D GAN inversion in high detail~\cite{abdal2019image2stylegan,richardson2021encoding,alaluf2021restyle,abdal2020image2stylegan++,zhu2020domain,tewari2020pie}, but nevertheless, the problem remains underexplored in 3D.

Optimization-based inversion methods are often superior to encoder-based approaches in terms of reconstruction quality. However, encoder-based techniques are orders of magnitude faster as they map a given image to the latent space of GAN in a single forward pass. 
Compared to 2D GAN inversion, 3D GAN inversion is a more challenging task as the inversion needs to both preserve the identity of an input image and plausibly embed the head in 3D space. 
In particular, optimization-based 2D GAN inversion methods that have no knowledge of the specific GAN architecture make sure to yield a high-quality rendering of the desired image from the same camera view, but the lack of any geometry information in the image may produce broken or stretched geometry when rendered from a novel camera. Optimization-based 3D GAN inversion techniques improve these shortcomings by adding 3D constraints in the optimization process. Even though these techniques prevent geometry collapse and offer high-fidelity reconstruction, they are slow and time-consuming. We improve the above-mentioned shortcomings in two separate ways. First, by predicting an input latent code for the EG3D generator with a convolutional encoder, we observe that the geometry is preserved better than by optimizing it. This can be attributed to the fact that the encoder, trained for the inversion task, is exposed to thousands of images under different poses and, in this way, learns to be 3D-aware. 
Second, we utilize the knowledge about the model and improve the details and consistency by predicting offsets to the tri-planes that constitute the 3D representation in EG3D. Unlike voxel grids or implicit representations, tri-planes can be naturally estimated by 2D convnets and, as demonstrated by our experiments, can realistically express object features beyond the capabilities of an input latent code, e.g., hands and long hair (see Fig.~\ref{fig:introduction}). This advantage is attained by recovering the object representation directly in the world space. Since the tri-plane offsets are fully predicted by convolutional layers, our inversion can run in close to real time on modern GPUs.

We propose the EG3D-specific inversion scheme in two stages. In the first stage, the initial inversion is obtained using the latent encoder that directly embeds the input image into the $\mathcal{W}+$ space of EG3D. In the second stage, we introduce another encoder, TriPlaneNet, that learns to refine the initial reconstruction. 
Conditioned on the input image and corresponding tri-plane features, it predicts a numerical offset for them. 
The system is trained with a combination of perceptual and photometric losses. In addition, we make use of the soft constraint based on the mirror image -- \emph{probably symmetric prior} inspired by~\cite{Wu2019UnsupervisedLO} -- that makes the encoder even more 3D-aware.

To summarize, our contributions are the following:
\begin{itemize}
    \item We propose a novel and fast inversion framework for EG3D that enables high-quality reconstruction and plausible geometric embedding of a head in 3D space by directly utilizing the tri-plane representation and a soft symmetry constraint.
    \item We demonstrate that our method achieves on-par reconstruction quality compared to optimization-based inversion methods and is an order of magnitude time faster. Our method is also more resilient towards harder cases, such as when a hat or accessories are featured.
\end{itemize}

%% file: sections/02_related_works.tex
\section{Related Work}\label{relatedwork}
\noindent \textbf{3D Generative Models for Human Faces.}
Representing and generating diverse 3D human faces and heads attracted increasing attention over the last decade \cite{nguyen2019hologan,chen2016infogan,huang2017beyond}, while the appearance of NeRF~\cite{mildenhall2021nerf} has sparked additional interest in that topic. 
The first generative models built upon NeRF-style volumetric integration \cite{schwarz2020graf,niemeyer2021giraffe} achieved generalization by conditioning the multi-layer perceptron on latent codes, representing the object's shape and appearance. 
Later introduced $\pi$-GAN \cite{chan2021pi} and StyleNeRF \cite{gu2021stylenerf} condition the generative network on the output of the StyleGAN-like generator~\cite{karras2020analyzing}, which amounted to the higher-quality rendering of faces and arbitrary objects with subtle details. 
As a next major improvement step, authors of EG3D~\cite{chan2022efficient} propose a tri-plane 3D representation that serves as a bridge between expressive implicit representations and spatially-restricting explicit representations. 
As a byproduct, methods such as EG3D and StyleSDF~\cite{or2022stylesdf} allow for the extraction of explicit, highly detailed geometry of the human faces, despite the fact that they are trained without any volumetric supervision.
Further, recently demonstrated abilities of diffusion models to generate highly accurate 2D images are currently being transferred onto 3D objects~\cite{zeng2022lion,muller2022diffrf} and 3D human heads~\cite{wang2022rodin,pan2023avatarstudio}.

\noindent \textbf{GAN Inversion.} Unlike other kinds of generative models, such as VAE or normalizing flows, inverting a GAN (finding the appropriate latent code for a given image) is oftentimes a tricky and computationally demanding task. 
Early attempts focused on the tuning of the latent code with the optimization-based approaches~\cite{creswell2018inverting,lipton2017precise,karras2020analyzing}. 
Various approaches exploited the idea of predicting latent representation by an encoder~\cite{guan2020collaborative,luo2017learning,perarnau2016invertible,zhu2016generative,park2019semantic}.
In~\cite{roich2022pivotal}, a universal PTI method is introduced, which comprises the optimization of a latent code and, consequently, fine-tuning parameters of the generator.
A recent survey on GAN inversion~\cite{xia2022gan} compares multiple generic techniques introduced since the appearance of GANs.

\noindent \textbf{Inversion of 2D GAN.} For StyleGAN, an important observation was made by the authors of~\cite{abdal2019image2stylegan} that operating in the extended $\mathcal{W}+$ space is significantly more expressive than in the restrictive $\mathcal{W}$ generator input space. 
The latter idea has been strengthened and better adapted for face editing with the appearance of pSp~\cite{richardson2021encoding} and e4e~\cite{tov2021designing}, as well as of their cascaded variant ReStyle~\cite{alaluf2021restyle} and other works~\cite{abdal2020image2stylegan++,zhu2020domain,tewari2020pie}.
Similarly to PTI but in an encoder-based setting, HyperStyle~\cite{alaluf2022hyperstyle} and HyperInverter~\cite{dinh2022hyperinverter} predict offsets to the StyleGAN generator weights in a lightweight manner in order to represent the target picture in a broader space of parameters.

\noindent \textbf{Inversion of 3D GAN.} 
Unlike the 2D case, the inversion of a 3D GAN is a significantly more advanced problem due to the arising ambiguity: the latent code must be both compliant with the target image and correspond to its plausible 3D representation. 
While PTI remains a universal method that solves this problem for an arbitrary generator, recent art demonstrates that the quality rapidly declines when the PTI inversion result is rendered from a novel view.
The suggested ways of resolving this fidelity-consistency trade-off for an arbitrary 3D GAN include incorporating multi-view consistency or geometry regularizers~\cite{li20223d,xie2023high}, augmenting training with surrogate mirrored images~\cite{yin20223d}, introducing local features~\cite{lan2023self}, or optimizing camera parameters and latent code simultaneously~\cite{ko20233d}. 
All of these approaches are still optimization-based and require at least a few minutes of inference time per image.
A concurrent encoder-based work Live 3D Portrait~\cite{Trevithick2023RealTimeRF} also leverages the tri-plane representation for high-fidelity reconstruction while relying on a self-constructed generator instead of EG3D and skipping the latent code prediction part.
Our work solves the inversion for the pretrained, frozen EG3D generator and addresses face manipulation due to the use of the latent space. 
Additionally, in~\cite{Trevithick2023RealTimeRF}, the training pipeline is reversed compared to ours. Starting from the random latent code, they generate synthetic images from EG3D to train the encoder. In contrast, we pass real and synthetic images through the encoder to generate latent codes.     
Another work, EG3D-GOAE~\cite{yuan2023make}, concurrent to ours, modifies the internal features of EG3D instead of tri-planes directly.

\begin{table*}[t]
\setlength{\tabcolsep}{0pt}
\renewcommand{\arraystretch}{0}
\begin{tabular}{c@{\hskip 0.3cm}ccccccc}
    \raisebox{2.5\height}{\rotatebox[origin=c]{90}{Input}} & 
    \includegraphics[width=0.138\textwidth]{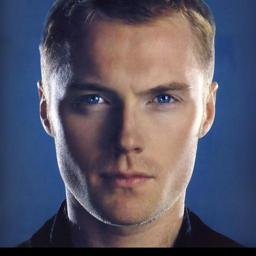} & 
    \includegraphics[width=0.138\textwidth]{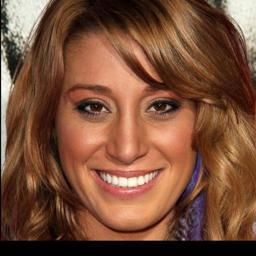} & 
    \includegraphics[width=0.138\textwidth]{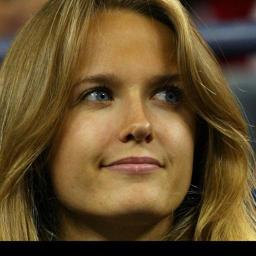} & 
    \includegraphics[width=0.138\textwidth]{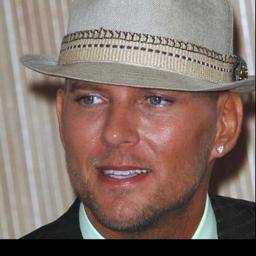} & 
    \includegraphics[width=0.138\textwidth]{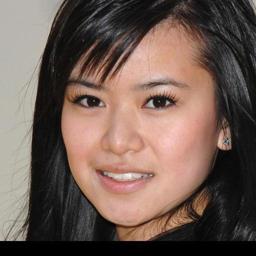} & 
    \includegraphics[width=0.138\textwidth]{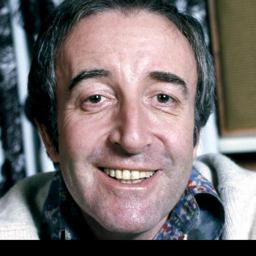} & 
    \includegraphics[width=0.138\textwidth]{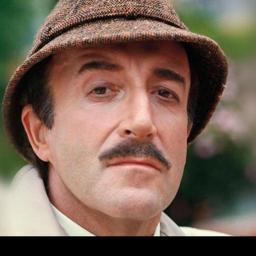}  \\[0.3cm]
\raisebox{1.2\height}{\rotatebox[origin=c]{90}{$\mathcal{W}+$ opt. \cite{karras2020analyzing}}} & 
    \includegraphics[width=0.138\textwidth]{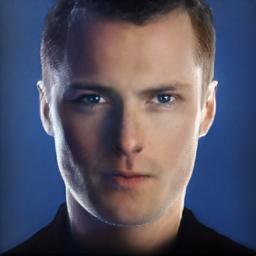} & 
    \includegraphics[width=0.138\textwidth]{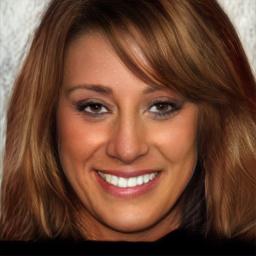} & 
    \includegraphics[width=0.138\textwidth]{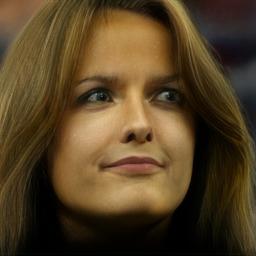} & 
    \includegraphics[width=0.138\textwidth]{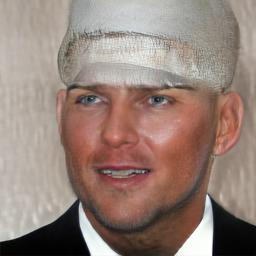} & 
    \includegraphics[width=0.138\textwidth]{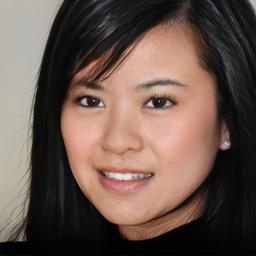} & 
    \includegraphics[width=0.138\textwidth]{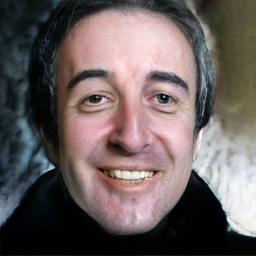} & 
    \includegraphics[width=0.138\textwidth]{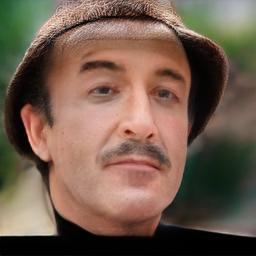}  \\
\raisebox{1.7\height}{\rotatebox[origin=c]{90}{PTI~\cite{roich2022pivotal}}} & 
    \includegraphics[width=0.138\textwidth]{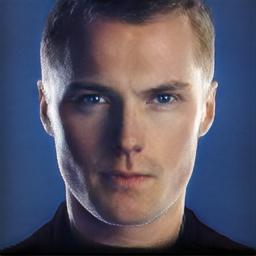} & 
    \includegraphics[width=0.138\textwidth]{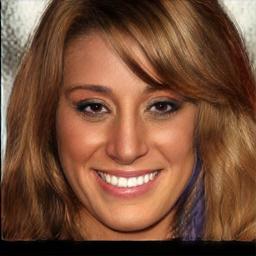} & 
    \includegraphics[width=0.138\textwidth]{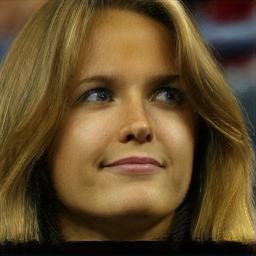} & 
    \includegraphics[width=0.138\textwidth]{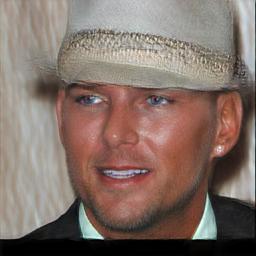} & 
    \includegraphics[width=0.138\textwidth]{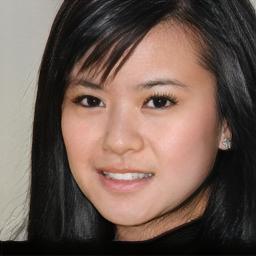} & 
    \includegraphics[width=0.138\textwidth]{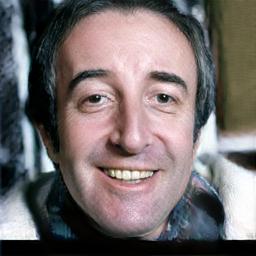} & 
    \includegraphics[width=0.138\textwidth]{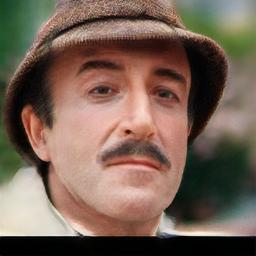}  \\
\raisebox{1.7\height}{\rotatebox[origin=c]{90}{pSp~\cite{richardson2021encoding}}} & 
    \includegraphics[width=0.138\textwidth]{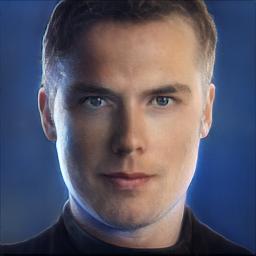} & 
    \includegraphics[width=0.138\textwidth]{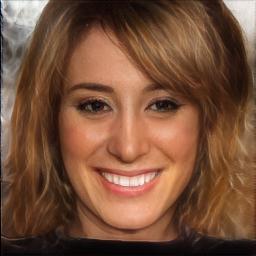} & 
    \includegraphics[width=0.138\textwidth]{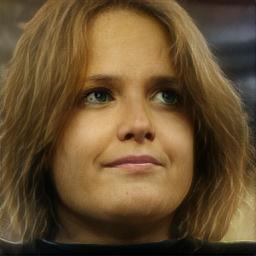} & 
    \includegraphics[width=0.138\textwidth]{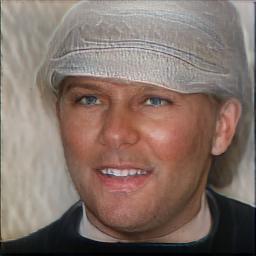} & 
    \includegraphics[width=0.138\textwidth]{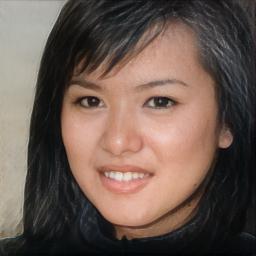} & 
    \includegraphics[width=0.138\textwidth]{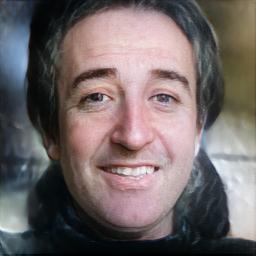} & 
    \includegraphics[width=0.138\textwidth]{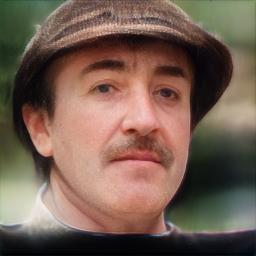}  \\
\raisebox{2.5\height}{\rotatebox[origin=c]{90}{\textbf{Ours}}} & 
    \includegraphics[width=0.138\textwidth]{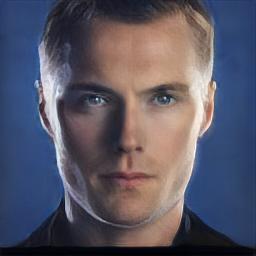} & 
    \includegraphics[width=0.138\textwidth]{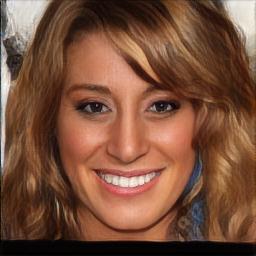} & 
    \includegraphics[width=0.138\textwidth]{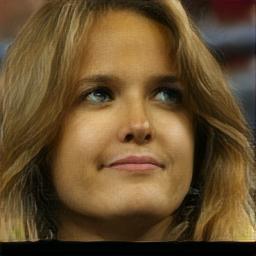} & 
    \includegraphics[width=0.138\textwidth]{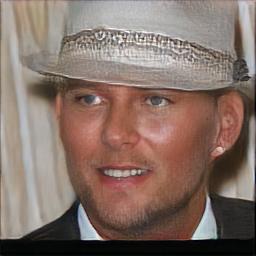} & 
    \includegraphics[width=0.138\textwidth]{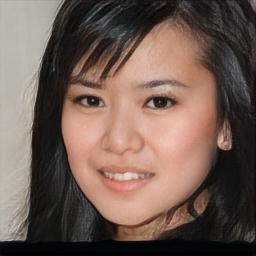} & 
    \includegraphics[width=0.138\textwidth]{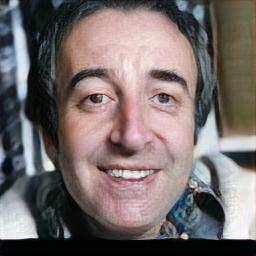} & 
    \includegraphics[width=0.138\textwidth]{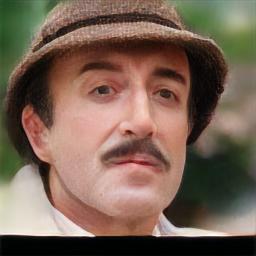} 
\end{tabular}
\captionof{figure}{Qualitative comparison for image reconstruction. Compared to other approaches, our method can reconstruct a face in the same view in more detail, especially introducing more detail for features such as hats, hair, and background.}
\label{visualcomp}
\vspace{-0.3cm}
\end{table*}

%% file: sections/03_method.tex
\section{Method}\label{method}

\begin{table*}[h!]
    \centering
    \setlength{\tabcolsep}{0cm}
    \resizebox{\textwidth}{!}{%
    \begin{tabular}{lllllllllll}
        \multicolumn{5}{l}{\hskip 4cm Novel Views} &
        \hskip 0.5cm Input
        & 
        \multicolumn{5}{l}{\hskip 2.5cm Novel Views}
        \\
        \raisebox{3.0\height}{$\mathcal{W}+$~\cite{karras2020analyzing}}
        &
        \includegraphics[width=0.099\textwidth]{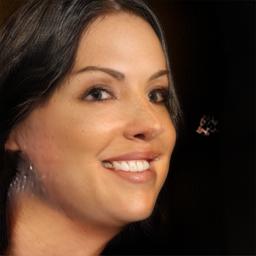}
        &
        \includegraphics[width=0.099\textwidth]{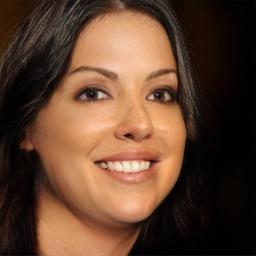}
        &
        \includegraphics[width=0.099\textwidth]{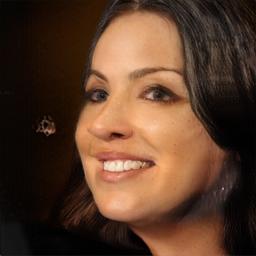}
        &
        \includegraphics[trim={2.2cm 2.2cm 2.2cm 2.2cm},clip,width=0.099\textwidth]{images/4_experiments/rotation/w_plus/10008_0.6.jpg}
        & 
        \multirow{3}{*}{
        \includegraphics[width=0.099\textwidth]{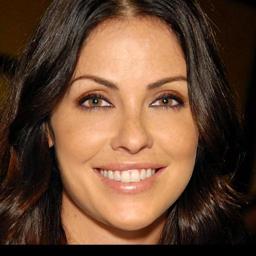}
        }
        &
        \includegraphics[width=0.099\textwidth]{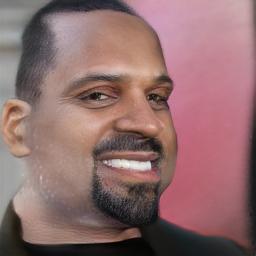}
        &
        \includegraphics[width=0.099\textwidth]{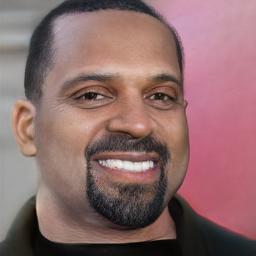}
        &
        \includegraphics[width=0.099\textwidth]{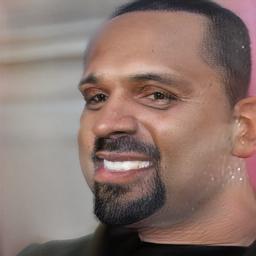}
        &
        \includegraphics[trim={2.2cm 2.2cm 2.2cm 2.2cm},clip,width=0.099\textwidth]{images/4_experiments/rotation/w_plus/5358_0.6.jpg}
        &
        \raisebox{3.0\height}{\hskip 0.1cm $\mathcal{W}+$~\cite{karras2020analyzing}}
        \\
        \raisebox{3.0\height}{PTI~\cite{roich2022pivotal}}
        &
        \includegraphics[width=0.099\textwidth]{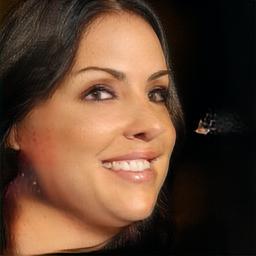}
        &
        \includegraphics[width=0.099\textwidth]{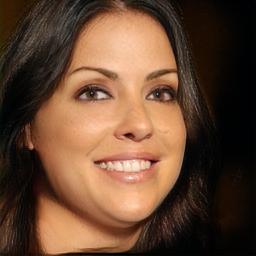}
        &
        \includegraphics[width=0.099\textwidth]{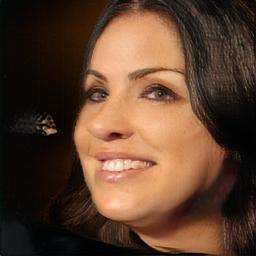}
        &
        \includegraphics[trim={2.2cm 2.2cm 2.2cm 2.2cm},clip,width=0.099\textwidth]{images/4_experiments/rotation/PTI/10008_0.6.jpg}
        &
        &
        \includegraphics[width=0.099\textwidth]{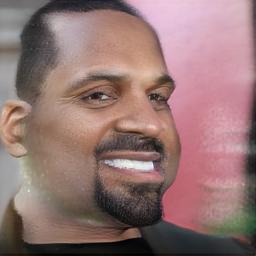}
        &
        \includegraphics[width=0.099\textwidth]{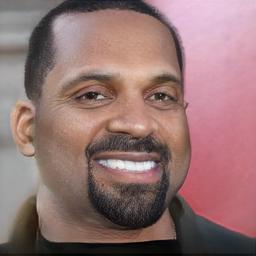}
        &
        \includegraphics[width=0.099\textwidth]{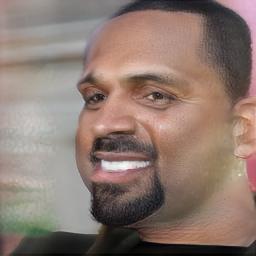}
        &
        \includegraphics[trim={2.2cm 2.2cm 2.2cm 2.2cm},clip,width=0.099\textwidth]{images/4_experiments/rotation/PTI/5358_0.6.jpg}
        &
        \raisebox{3.0\height}{\hskip 0.1cm PTI~\cite{roich2022pivotal}}
        \\
        \raisebox{3.0\height}{Pose Opt.~\cite{ko20233d}\,}
        &
        \includegraphics[width=0.099\textwidth]{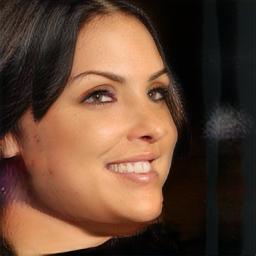}
        &
        \includegraphics[width=0.099\textwidth]{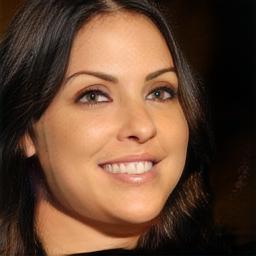}
        &
        \includegraphics[width=0.099\textwidth]{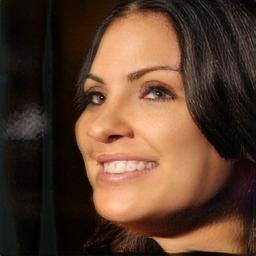}
        &
        \includegraphics[trim={2.2cm 2.2cm 2.2cm 2.2cm},clip,width=0.099\textwidth]{images/4_experiments/rotation/poseopt/10008_0.6.jpg}
        &
        &
        \includegraphics[width=0.099\textwidth]{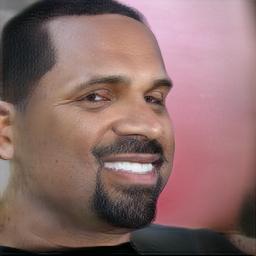}
        &
        \includegraphics[width=0.099\textwidth]{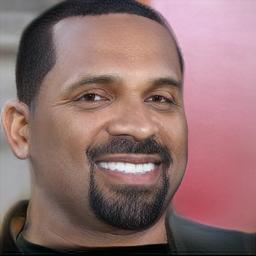}
        &
        \includegraphics[width=0.099\textwidth]{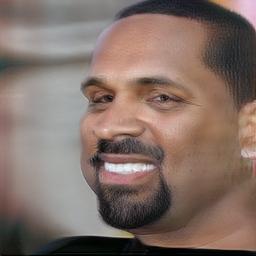}
        &
        \includegraphics[trim={2.2cm 2.2cm 2.2cm 2.2cm},clip,width=0.099\textwidth]{images/4_experiments/rotation/poseopt/5358_0.6.jpg}
        &
        \raisebox{3.0\height}{\hskip 0.1cm Pose Opt.~\cite{ko20233d}}
        \\
        \raisebox{3.0\height}{SPI~\cite{yin20223d}}
        &
        \includegraphics[width=0.099\textwidth]{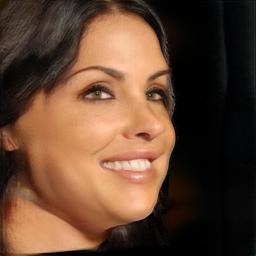}
        &
        \includegraphics[width=0.099\textwidth]{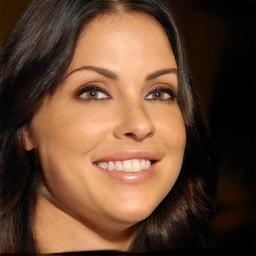}
        &
        \includegraphics[width=0.099\textwidth]{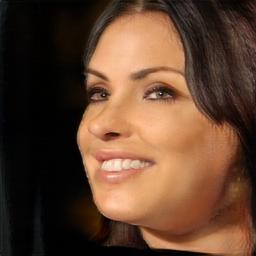}
        &
        \includegraphics[trim={2.2cm 2.2cm 2.2cm 2.2cm},clip,width=0.099\textwidth]{images/4_experiments/rotation/SPI/10008_0.6.jpg}
        & 
        \multirow{3}{*}{
        \includegraphics[width=0.099\textwidth]{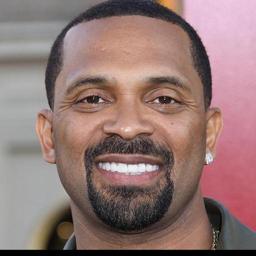}
        }
        &
        \includegraphics[width=0.099\textwidth]{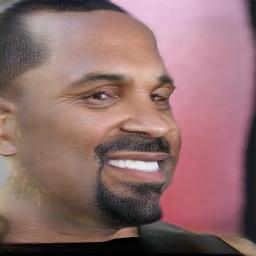}
        &
        \includegraphics[width=0.099\textwidth]{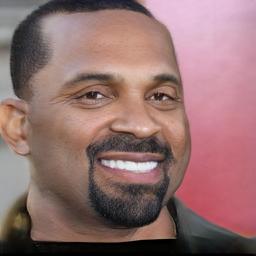}
        &
        \includegraphics[width=0.099\textwidth]{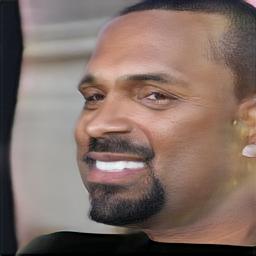}
        &
        \includegraphics[trim={2.2cm 2.2cm 2.2cm 2.2cm},clip,width=0.099\textwidth]{images/4_experiments/rotation/SPI/5358_0.6.jpg}
        &
        \raisebox{3.0\height}{\hskip 0.1cm SPI~\cite{yin20223d}}
        \\
        \raisebox{3.0\height}{pSp~\cite{richardson2021encoding}}
        &
        \includegraphics[width=0.099\textwidth]{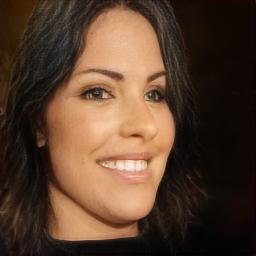}
        &
        \includegraphics[width=0.099\textwidth]{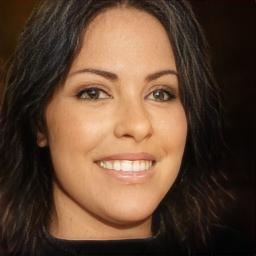}
        &
        \includegraphics[width=0.099\textwidth]{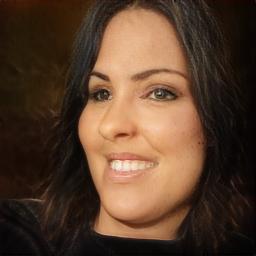}
        &
        \includegraphics[trim={2.2cm 2.2cm 2.2cm 2.2cm},clip,width=0.099\textwidth]{images/4_experiments/rotation/pSp/10008_0.6.jpg}
        &
        &
        \includegraphics[width=0.099\textwidth]{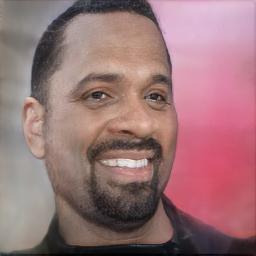}
        &
        \includegraphics[width=0.099\textwidth]{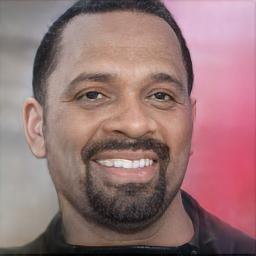}
        &
        \includegraphics[width=0.099\textwidth]{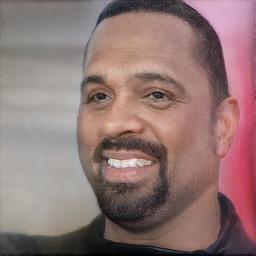}
        &
        \includegraphics[trim={2.2cm 2.2cm 2.2cm 2.2cm},clip,width=0.099\textwidth]{images/4_experiments/rotation/pSp/5358_0.6.jpg}
        &
        \raisebox{3.0\height}{\hskip 0.1cm pSp~\cite{richardson2021encoding}}
        \\
        \raisebox{3.0\height}{\textbf{Ours}}
        &
        \includegraphics[width=0.099\textwidth]{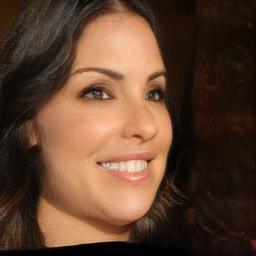}
        &
        \includegraphics[width=0.099\textwidth]{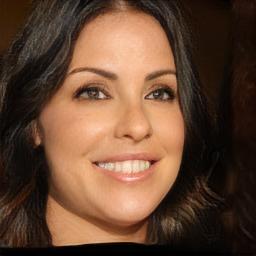}
        &
        \includegraphics[width=0.099\textwidth]{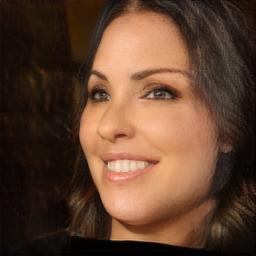}
        &
        \includegraphics[trim={2.2cm 2.2cm 2.2cm 2.2cm},clip,width=0.099\textwidth]{images/4_experiments/rotation/triplanenet/10008_0.6.jpg}
        &
        &
        \includegraphics[width=0.099\textwidth]{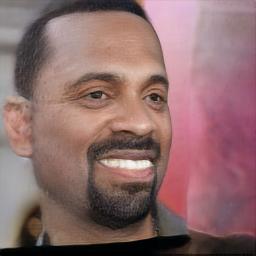}
        &
        \includegraphics[width=0.099\textwidth]{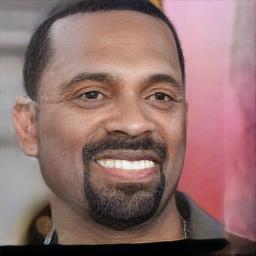}
        &
        \includegraphics[width=0.099\textwidth]{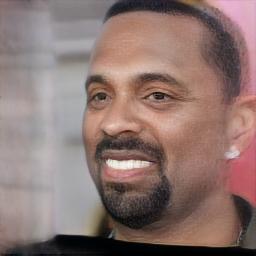}
        &
        \includegraphics[trim={2.2cm 2.2cm 2.2cm 2.2cm},clip,width=0.099\textwidth]{images/4_experiments/rotation/triplanenet/5358_0.6.jpg}
        &
        \raisebox{3.0\height}{\hskip 0.1cm \textbf{Ours}}
        \\
    \end{tabular}%
    }
    \vspace{-0.1cm}
    \captionof{figure}{Qualitative evaluation on novel view rendering of yaw angle -0.6, -0.3, and 0.6 radians (full and zoom-in). In comparison to others, our method preserves identity and multi-view consistency better when rendered from a novel view.}
    \label{novelview}
    \vspace{-0.3cm}
\end{table*}

\subsection{Preliminaries}

\noindent \textbf{GAN inversion.}
Given a target image $x$, the goal of GAN inversion is to find a latent code that minimizes the reconstruction loss between the synthesized image and the target image:
\begin{equation}
    \hat{w} = \operatorname*{argmin}_{w}\mathcal{L}(x, G(w; \theta))
    \label{gan-obj}
\end{equation}
where $G(w; \theta)$ is the image generated by a pre-trained generator $G$ parameterized by weights $\theta$, over the latent $w$. 
The problem in (\ref{gan-obj}) can be solved via optimization or encoder-based approaches. Encoder-based approaches utilize an encoder network $E$ to map real images into a latent code. The training of an encoder network is performed over a large set of images $\{x^{i}\}_{i=1}^{N}$ to minimize:
\begin{equation}
    \setlength{\abovedisplayskip}{3pt}
    \setlength{\belowdisplayskip}{3pt}
    \min_E \sum_{i=1}^N\mathcal{L}(x^i, G(E(x^i); \theta))
\end{equation}
During inference, an input image is inverted by $G(E(x); \theta)$. In the recent works~\cite{roich2022pivotal, alaluf2022hyperstyle, dinh2022hyperinverter}, a number of approaches are proposed to additionally estimate image-specific generator parameters $\theta(x)$ by a convolutional network.\newline

\noindent \textbf{EG3D.}
EG3D \cite{chan2022efficient} uses tri-plane 3D representation for geometry-aware image synthesis from 2D images. EG3D image generation pipeline consists of several modules: a StyleGAN2-based feature generator, a tri-plane representation, a lightweight neural decoder, a volume renderer, and a super-resolution module. To synthesize an image, a random latent code $z \in \mathbb{R}^{D}$ (typically, $D=512$) and camera parameters are first mapped to a pivotal latent code $w \in \mathcal{W}+$ using a mapping network. Then, $w$ is fed into the StyleGAN2 CNN generator $\boldsymbol{G}(\cdot)$ to generate a $H \times W \times 96$ feature map. This feature map is reshaped to form three 32-channel planes, thus forming a tri-plane feature representation $\boldsymbol{T}$ of the corresponding object. To sample from the tri-plane features, a position $p \in \mathbb{R}^3$ is first projected onto the three feature planes. Then, corresponding feature vectors ($F_{xy}(p), F_{xz}(p), F_{yz}(p)$) are retrieved using bilinear interpolation and aggregated. These aggregated features are processed by a lightweight neural decoder to transform the feature into the estimated color and density at the location $p$. Volume rendering is then performed to project 3D feature volume into a feature image. Finally, a super-resolution module is utilized to upsample the feature image to the final image size. For simplicity, we will later refer to the lightweight neural decoder, renderer, and the super-resolution block, all combined, as the rendering block $\mathcal{R}(\cdot, \cdot)$. The high efficiency and expressiveness of EG3D, as well as the ability to work with tri-planes directly, motivates the development of our model-specific inversion algorithm.\newline

\noindent \textbf{pSp.}
Richardson \textit{et al}.~\cite{richardson2021encoding} proposed a pSp framework based on an encoder that can directly map real images into $\mathcal{W}+$ latent space of StyleGAN. In pSp, an encoder backbone with a feature pyramid generates three levels of feature maps. The extracted feature maps are processed by a map2style network to extract styles. The styles are then fed into the generator network to synthesize an image $\hat{y}$:

\begin{equation}
\hat{y} = G(E(x) + \bar{w}),
\end{equation}
where $G(\cdot)$ and $E(\cdot)$ denote the generator and encoder networks respectively and $\bar{w}$ is the average style vector of the pretrained generator.

\subsection{TriPlaneNet}
Our TriPlaneNet inversion framework comprises two branches
(see Fig.~\ref{method:fig} for the overview).
The first branch employs a latent encoder following a design of pSp to embed an input image into $\mathcal{W}+$ space of EG3D. Specifically, given an input image $x$, we train a encoder $\phi$ to predict the pivotal latent $\hat{w} \in \mathcal{W}+$:
\begin{equation}
    \hat{w} = \phi(x) + \bar{w}
\end{equation}
where the dimension of $\hat{w}$ is $K \times D$ (for the output image resolution of 128, $K=14$, and $D=512$). The pivotal code is then fed into StyleGAN2 generator $\boldsymbol{G}(\cdot)$ in the EG3D pipeline to obtain initial tri-plane features $\boldsymbol{T}$. Then, the tri-plane representation is processed by the rendering block $\mathcal{R}(\cdot, \cdot)$ to generate initial reconstruction $\hat{y}$:
\begin{equation}
    \setlength{\abovedisplayskip}{5pt}
    \setlength{\belowdisplayskip}{5pt}
    \hat{y} = \mathcal{R}(\boldsymbol{G}(\hat{w}), \pi)
\end{equation}
where $\pi$ is the input-view camera matrix.

The second branch consists of a convolutional auto-encoder $\psi$ that learns to predict numerical offsets to the initial tri-plane features. The input to the encoder module of the autoencoder network is the channel-wise concatenation of initial reconstruction $\hat{y}$, the difference between an input image and initial input-view reconstruction ($x - \hat{y}$), and the difference between a mirrored input image and initial mirror-view reconstruction ($x_m - \hat{y}_m$). The decoder takes input from the encoder and first branch tri-plane features. Given these inputs, the autoencoder is tasked with computing tri-plane offsets $\Delta \boldsymbol{T}$ with respect to tri-plane features obtained in the first branch:
\begin{equation}
    \Delta \boldsymbol{T} = \boldsymbol{\psi}(\hat{y},\, x - \hat{y},\, x_m - \hat{y}_m,\, G(\hat{w}))
\end{equation}
The new tri-plane features corresponding to the inversion of the input image $x$ are then computed as an element-wise addition of tri-plane offsets $\Delta \boldsymbol{T}$ with initial tri-plane features $\boldsymbol{T} = G(\hat{w})$. This new tri-plane representation is similarly processed by the rendering block $\mathcal{R}(\cdot, \cdot)$ to obtain the final reconstructed image:
\begin{equation}
    y = \mathcal{R}(\boldsymbol{T} + \Delta \boldsymbol{T}, \pi)
\end{equation}

The autoencoder follows the typical U-Net \cite{Ronneberger2015UNetCN} architecture, consisting of a contracting path and an expansive path.
The decoder architecture is similar to that of RUNet \cite{Hu2019RUNet} with some minor modifications.
A detailed view of the architecture is presented in Appendix~\ref{supp:impdetails}.  

\subsection{Loss Functions}
The pipeline is trained by minimizing the loss function that decomposes into the separate loss expressions for two branches:
\begin{equation}
    \mathcal{L}_{\phi, \psi}(x, y, \hat{y}, y_m, \hat{y}_m) = \mathcal{L}_\phi(x, \hat{y}, \hat{y}_m) + \mathcal{L}_\psi(x, y, y_m)
\end{equation}

For training the encoder $\phi(\cdot)$ in the first branch, we employ pixel-wise $\mathcal{L}_2$ loss, LPIPS loss \cite{Zhang2018TheUE}, and ID loss \cite{Deng2018ArcFaceAA}. Therefore, the total loss formulation is given by
\begin{equation}
    \begin{aligned}
        \mathcal{L}_{\phi}(x, \hat{y}, \hat{y}_m) &= \lambda_1\mathcal{L}_2(x, \hat{y}) + \lambda_2\mathcal{L}_\textrm{LPIPS}(x, \hat{y}) 
        \\ & + \lambda_3\mathcal{L}_\textrm{id}(x, \hat{y}) + \lambda_m\mathcal{L}_{m}(x_m, \hat{y}_m)
    \label{encoder-loss}
    \end{aligned}
\end{equation}
where $x_m = \textrm{flip}(x)$, and $\mathcal{L}_{m}(x_m, \hat{y}_m)$ is a \textit{probably symmetric prior} defined as
\begin{equation}
    \vspace{-0.1cm}
    \begin{multlined}
        \mathcal{L}_{m}(x_m, \hat{y}_m) = \lambda_4\mathcal{L}_\textrm{symm}(x_m, \hat{y}_m, \sigma(x_m)) \\ + \lambda_5\mathcal{L}_\textrm{LPIPS}(x_m, \hat{y}_m) + \lambda_6\mathcal{L}_\textrm{id}(x_m, \hat{y}_m),
    \end{multlined}
    \label{encoder-loss1}
\end{equation}
where $\mathcal{L}_\textrm{symm}$ is a symmetric term inspired by~\cite{Wu2019UnsupervisedLO}. Since human faces are not perfectly symmetric, the symmetric term is based on a per-pixel Gaussian density with the pixel-wise uncertainty map $\sigma(x_m)$ that assigns lower confidence to the region in the mirrored image where the symmetry assumption fails.%

\setlength{\tabcolsep}{1.5pt}
\begin{table}[b!]
\caption{Quantitative comparison on save-view inversion. The inference time, including the EG3D pass, is given for a single RTX A100 Ti GPU. \textit{Ours} exceeds other encoder-based methods by photometric scores and embeds the head in 3D space significantly better than all the other methods (see \textit{Depth $\downarrow$}).}
\vspace{-0.4cm}
\begin{center}
\resizebox{0.49\textwidth}{!}{
    \begin{tabular}{ l l | c c c c | c | c}
        \hline
        \multicolumn{2}{l|}{Method} & MSE$\downarrow$ & LPIPS$\downarrow$ & ID $\uparrow$ & MS-SSIM$\uparrow$ & \,Depth$\downarrow$\, & Infer. time $\downarrow$\\
        \hline \hline
        \parbox[t]{2mm}{\multirow{4}{*}{\rotatebox[origin=c]{90}{\textcolor{RawSienna}{Optim.}}}}\,\,\ldelim\{{3.8}{*} &  $\mathcal{W}+$ \cite{karras2020analyzing} & \,0.071 & 0.17 & 0.55 & 0.80 & 0.086 & \textbf{77.07} s\\  
        & PTI \cite{roich2022pivotal} & \,0.013 & 0.07 & 0.76 & 0.89 & 0.087 & 119.34 s\\
        & P. Opt.~\cite{ko20233d} & \,0.014 & 0.08 & 0.67 & 0.88 & 0.119 & 110.86 s\\
        & SPI~\cite{yin20223d} & \,\underline{\textbf{0.005}} & \underline{\textbf{0.05}} & \underline{\textbf{0.94}} & \underline{\textbf{0.95}} & \textbf{0.078} & 258.84 s\\
        \hline
         \parbox[t]{2mm}{\multirow{4.5}{*}{\rotatebox[origin=c]{90}{\textcolor{blue}{Encoder}}}}\,\,\ldelim\{{4.7}{*}
         & e4e \cite{tov2021designing} & \,0.060 & 0.21 & 0.33 & 0.70 & 0.061 & \underline{\textbf{0.04 s}}\\
         & pSp \cite{richardson2021encoding} & \,0.045 & 0.18 & 0.40 & 0.73 & 0.076 & \underline{\textbf{0.04 s}}\\
        & EG3D-GOAE \cite{yuan2023make} & \,0.026 & 0.11 & 0.67 & 0.84 & 0.053 & 0.18 s \\
        & Ours \textit{(FFHQ)} & 0.016 & 0.07 & \textbf{0.78} & 0.89 & \underline{\textbf{0.042}} & 0.12 s\\
        & Ours & \textbf{0.015} & \textbf{0.06} & 0.77 & \textbf{0.90} & 0.047 & 0.12 s\\
        \hline
    \end{tabular}
}
\end{center}
\label{quant-same-view-table}
\vspace{-0.7cm}
\end{table}

The loss for the second branch $\mathcal{L}_\psi$ is constructed the same way as $\mathcal{L}_\phi$ by replacing $\mathcal{L}_2$ with $\mathcal{L}_1$ smooth loss in (\ref{encoder-loss}), inside $\mathcal{L}_{symm}$ in (\ref{encoder-loss1}) and first branch output $\hat{y}$ with the second branch output $y$. %
Appendix~\ref{supp:impdetails} contains more details about the loss functions.

\begin{table}[t!]
\caption{Quantitative comparison on novel view rendering of the inverted representation. We outperform all the other baselines on extreme novel view yaw angles. MSE and others are not suitable here due to spatial misalignment.}
\vspace{-0.5cm}
\begin{center}
\resizebox{0.48\textwidth}{!}{
    \begin{tabular}{l l|c c c c c c|c}
        \hline
        \multicolumn{2}{l|}{\multirow{2}{*}{Method}} & %
        \multicolumn{6}{c|}{ID $\uparrow$ for novel yaw angle (rad):} & \,ID $\uparrow$\,\\
        \cline{3-9}
& & \,\,\,\,-0.8\,\,\,\, & \,\,\,\,-0.6\,\,\,\, & \,\,\,\,-0.3\,\,\,\, & \,\,\,\,0.3\,\,\,\, & \,\,\,\,0.6\,\,\,\, & \,\,\,\,0.8\,\,\,\,\, & Avg \\
        \hline \hline
        \parbox[t]{2mm}{\multirow{4}{*}{\rotatebox[origin=c]{90}{\textcolor{RawSienna}{Optim.}}}}\,\,\ldelim\{{3.8}{*} & $\mathcal{W}+$ \cite{karras2020analyzing} & 0.26 & 0.32 & 0.42 & 0.42 & 0.33 & 0.28 & 0.33\\ 
        & PTI \cite{roich2022pivotal} & 0.34 & 0.41 & 0.55 & 0.56 & 0.43 & 0.35 & 0.44\\
        & P. Opt.~\cite{ko20233d} & 0.30 & 0.36 & 0.50 & 0.50 & 0.38 & 0.31 & 0.39\\
        & SPI~\cite{yin20223d} & \textbf{0.43} & \textbf{0.52} & \underline{\textbf{0.70}} & \underline{\textbf{0.71}} & \textbf{0.54} & \textbf{0.44} & \underline{\textbf{0.55}}\\
        \hline
        \parbox[t]{2mm}{\multirow{4}{*}{\rotatebox[origin=c]{90}{\textcolor{blue}{Encoder}}}}\,\,\ldelim\{{4.7}{*}
        & e4e \cite{tov2021designing} & 0.19 & 0.23 & 0.28 & 0.28 & 0.24 & 0.21 & 0.23\\
        & pSp \cite{richardson2021encoding} & 0.24 & 0.28 & 0.35 & 0.36 & 0.29 & 0.25 & 0.29\\
        & EG3D-GOAE \cite{yuan2023make} & 0.38 & 0.46 & 0.57 & 0.57 & 0.48 & 0.40 & 0.47\\
        & Ours \textit{(FFHQ)} & \underline{\textbf{0.45}} & \underline{\textbf{0.54}} & \textbf{0.66} & \textbf{0.67} & \underline{\textbf{0.56}} & \underline{\textbf{0.47}} & \underline{\textbf{0.55}}\\
        & Ours & 0.44 & 0.53 & \textbf{0.66} & \textbf{0.67} & 0.55 & 0.46 & \underline{\textbf{0.55}}\\
        \hline
    \end{tabular}
}
\end{center}
\label{novel-view-table}
\vspace{-0.7cm}
\end{table}

%% file: sections/04_experiments.tex
\section{Experiments}

\begin{table}[b!]
{\footnotesize
    \vspace{-0.3cm}
    \centering
    \setlength\tabcolsep{0pt}
        \begin{tabular}{cccccc}
            &
            Input
            &
            w/o $\mathcal{L}_m$
            &
            No $2^{nd}$ branch
            &
            $\mathcal{L}_m=0.5$
            &
            Ours
            \\
            \raisebox{1.0\height}{\rotatebox[origin=c]{90}{Same View}}\,\,
            &
            \includegraphics[width=0.09\textwidth]{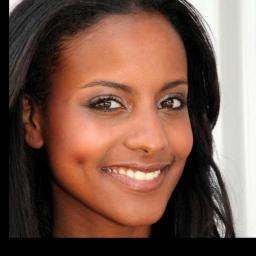}
            &
            \includegraphics[width=0.09\textwidth]{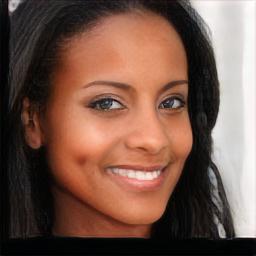}
            &
            \includegraphics[width=0.09\textwidth]{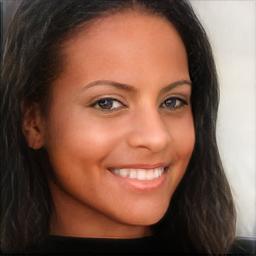}
            &
            \includegraphics[width=0.09\textwidth]{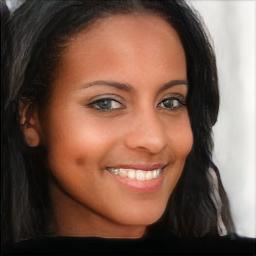}
            &
            \includegraphics[width=0.09\textwidth]{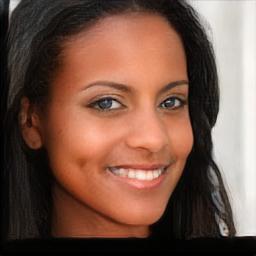}
            \\
            &
            \multicolumn{1}{r}{\raisebox{3.5\height}{\small Novel View}}\,\,
            &
            \includegraphics[width=0.09\textwidth]{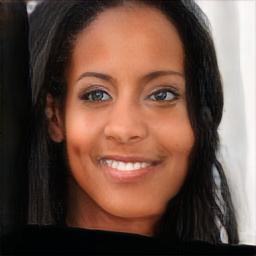}
            &
            \includegraphics[width=0.09\textwidth]{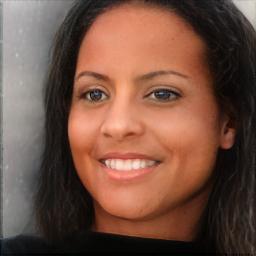}
            &
            \includegraphics[width=0.09\textwidth]{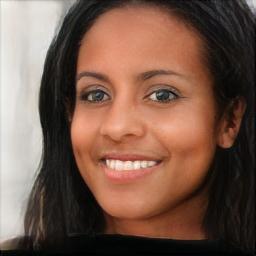}
            &
            \includegraphics[width=0.09\textwidth]{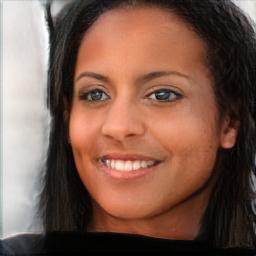}
            \\
            &
            \multicolumn{1}{r}{\raisebox{3.5\height}{\small Geometry}}\,\,
            &
            \includegraphics[width=0.09\textwidth]{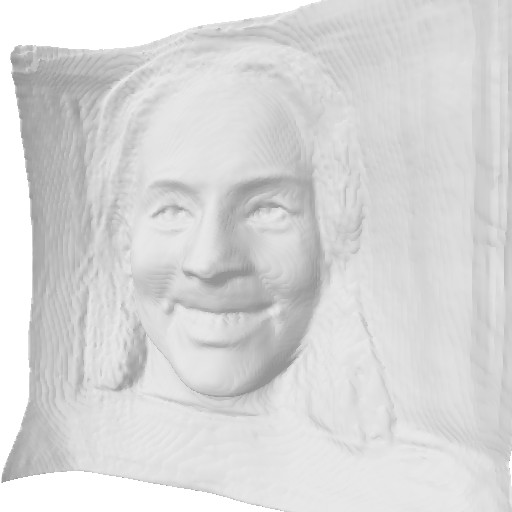}
            &
            \includegraphics[width=0.09\textwidth]{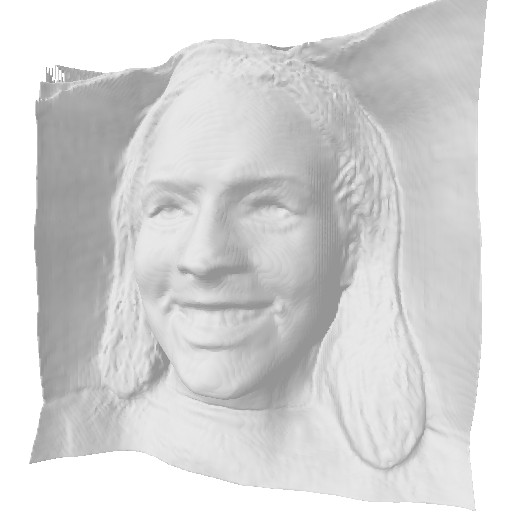}
            &
            \includegraphics[width=0.09\textwidth]{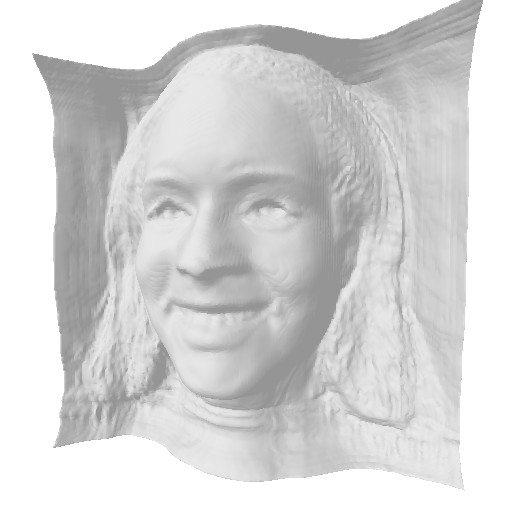}
            &
            \includegraphics[width=0.09\textwidth]{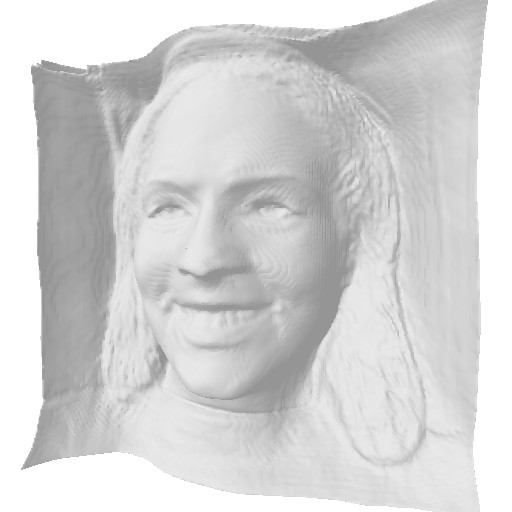}
            \\
        \end{tabular}%
}
\captionof{figure}{
        Qualitative ablation study for the loss, dataset, and architecture changes. \textit{Electronic zoom-in recommended.}
    }
\label{fig:ablation}
\vspace{-0.3cm}
\end{table}

\subsection{Training procedure}
\noindent \textbf{Datasets.} Since our focus is on the human facial domain, we use FFHQ \cite{Karras2018ASG} dataset and 100K generated images from EG3D pre-trained on FFHQ for training and perform the evaluation on the CelebA-HQ ~\cite{Liu2014DeepLF, Karras2017ProgressiveGO} test set. We extract the camera pose and pre-process the FFHQ and synthetic data in the same way as in \cite{chan2022efficient}. Since the pre-processing technique could not identify the camera poses of 4 images, we skipped the quantitative evaluation of 4 images for all the methods presented in the paper. We also augment the training dataset by mirroring it.  

\begin{table*}[h!]
\setlength{\tabcolsep}{0pt}
{\footnotesize
\begin{tabular}{clllcllll}
                     & \multicolumn{3}{c}{\textcolor{RawSienna}{$\mathcal{W}+$ opt.} + \dots} & \multicolumn{5}{c}{\textcolor{blue}{$\mathcal{W}+$ pred.} + \dots}                              \\
\vspace{0.5cm}       & \multicolumn{3}{c}{\multirow{8}{*}{\includegraphics[trim={6.5cm 5.5cm 6.5cm 0.2cm},clip,height=10cm]{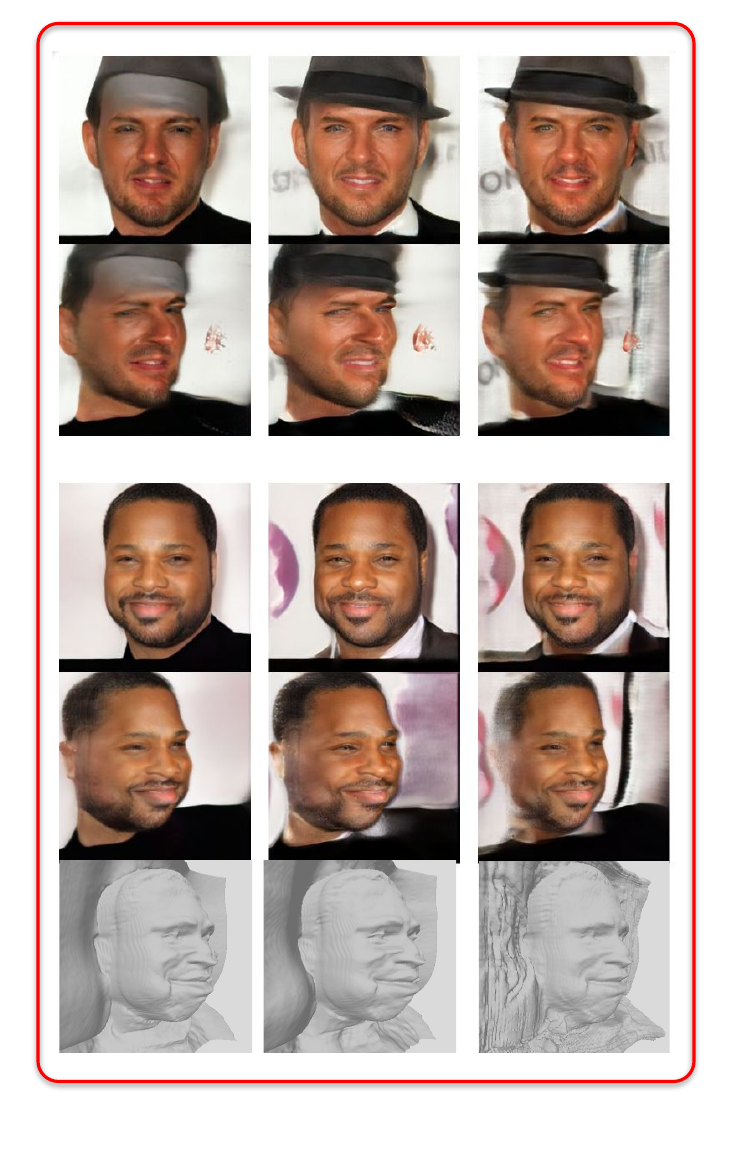}}}                          & \multicolumn{5}{c}{\multirow{8}{*}{\includegraphics[trim={3.5cm 5.5cm 3cm 0.2cm},clip,height=10cm]{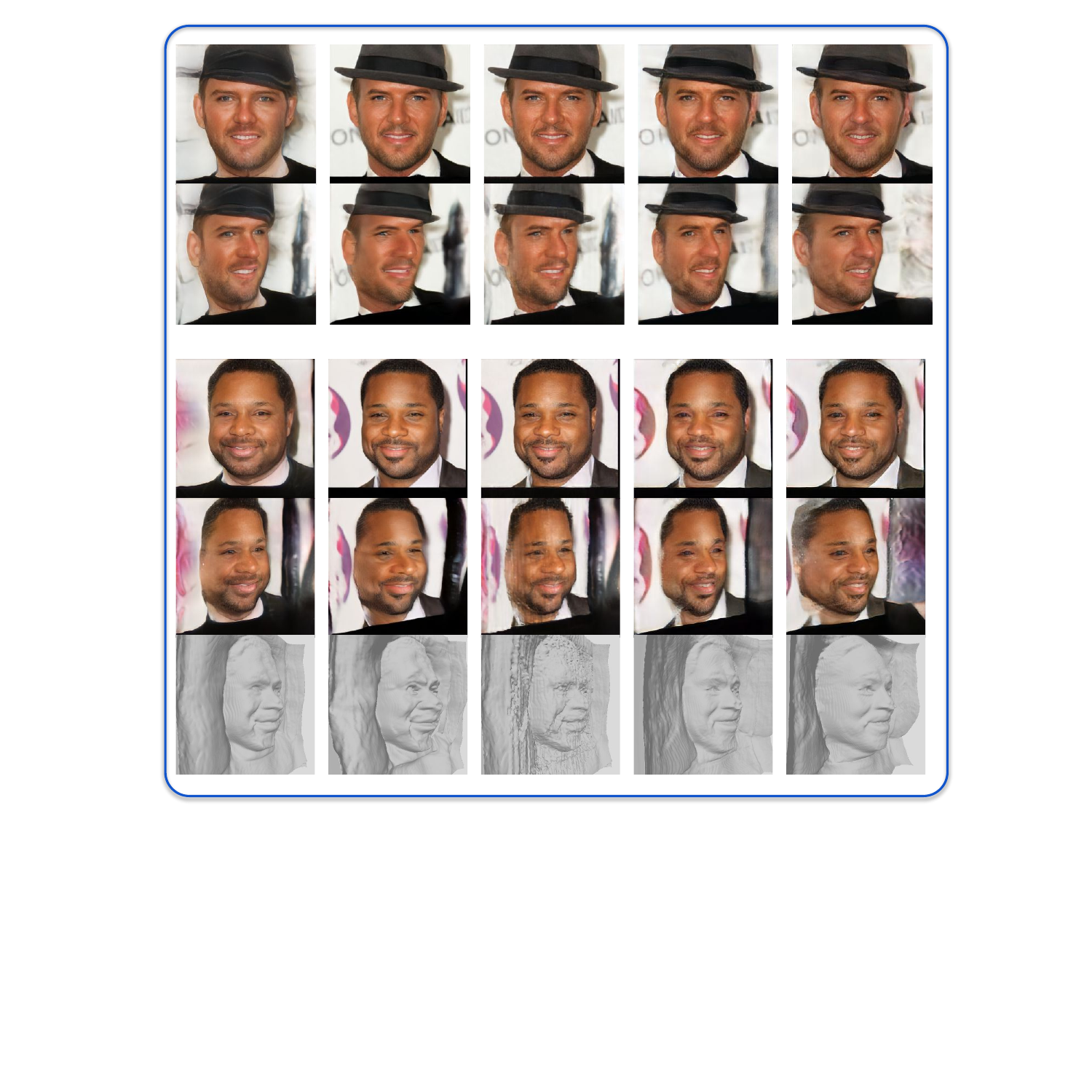}}}                                                        \\
\includegraphics[width=0.1\textwidth]{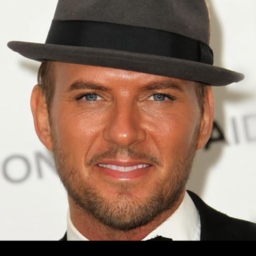}
                    & \multicolumn{3}{c}{}                                           & \multicolumn{5}{c}{}                                                                         \\
Input                & \multicolumn{3}{c}{}                                           & \multicolumn{5}{c}{}                                                                         \\
                     & \multicolumn{3}{c}{}                                           & \multicolumn{5}{c}{}                                                                         \\
\vspace{1.5cm}       & \multicolumn{3}{c}{}                                           & \multicolumn{5}{c}{}                                                                         \\
\includegraphics[width=0.1\textwidth]{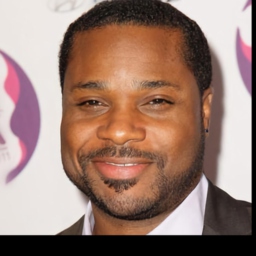}
                     & \multicolumn{3}{c}{}                                           & \multicolumn{5}{c}{}                                                                         \\
Input                & \multicolumn{3}{c}{}                                           & \multicolumn{5}{c}{}                                                                         \\
\vspace{1.8cm}       & \multicolumn{3}{c}{}                                           & \multicolumn{5}{c}{}                                                                         \\
                     &  \hspace{2.2cm}                      & {\footnotesize + \textcolor{RawSienna}{EG3D params}\,\,\,}   & {\footnotesize \textcolor{blue}{+ tri-plane}}   & \hspace{0.3cm}{\footnotesize (pSp)} & {\footnotesize \textcolor{RawSienna}{+ EG3D}\,\,\,\, \hspace{0.6cm}} & {\footnotesize \textcolor{RawSienna}{\,\,\,\,+ tri-plane}\,\,\,\,\,\,\,\,} & {\footnotesize + \textcolor{blue}{tri-plane pred.}} & \,\,{\footnotesize + \textcolor{blue}{tri-plane pred.}}
                     \\
\multicolumn{1}{l}{} & \multicolumn{1}{l}{}   & {\footnotesize \textcolor{RawSienna}{opt. (PTI)}}      &  {\footnotesize \textcolor{blue}{pred.}}                   & \hspace{2.1cm}                       & {\footnotesize \textcolor{RawSienna}{params opt.}}          & {\,\,\,\,\footnotesize \textcolor{RawSienna}{opt.}}        &  & \,\,{\footnotesize + symm. prior}
\vspace{-0.2cm}
\end{tabular}
}

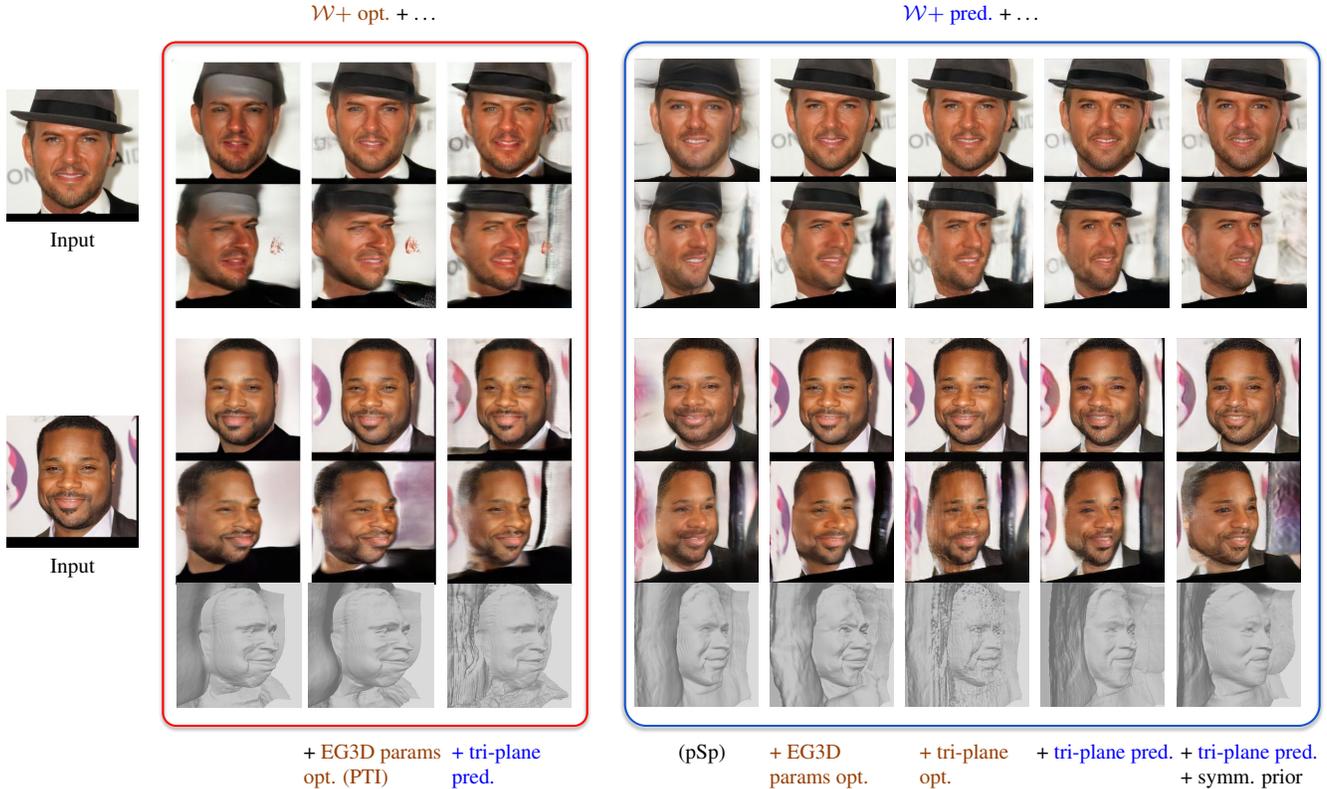
\captionof{figure}{Comparison of hybrid approaches on CelebA-HQ test dataset. We refer to the computation done via optimization as \textcolor{RawSienna}{opt.} and via an encoder as \textcolor{blue}{pred.} We observe that the methods starting from \textcolor{RawSienna}{$\mathcal{W}+$ opt.} yield elongated head geometry, whereas subsequent \textcolor{blue}{tri-plane pred.} can partially alleviate it. Experiments starting from \textcolor{blue}{$\mathcal{W}+$ pred.} demonstrate that the tri-plane space is more spatially restrictive than the EG3D parameters space. \textit{Ours} = \textcolor{blue}{$\mathcal{W}+$ pred.} + \textcolor{blue}{tri-plane pred.} + symmetry prior. \textit{Electronic zoom-in recommended.}}
\vspace{-0.3cm}
\label{analysis:fig}
\end{table*}

\noindent \textbf{Training details.}
Our pre-trained EG3D generator is also trained on the FFHQ dataset \cite{Karras2018ASG}. We train two versions of the same model: \textit{Ours}, trained on FFHQ and synthetic data, and $\textit{Ours (FFHQ)}$, trained only on FFHQ data. We discuss the motivation to use synthetic samples in Sec.~\ref{results} and describe the training procedure details in Appendix~\ref{supp:impdetails}.

\noindent \textbf{Baselines.} We compare our approach with both optimization- and encoder-based inversion methods. Among optimization-based methods, we compare to universal $\mathcal{W}+$ optimization~\cite{karras2020analyzing} and PTI~\cite{roich2022pivotal}, as well as to Pose Opt.~ \cite{ko20233d} and SPI~\cite{yin20223d}, recently introduced for 3D GANs. Among encoder-based methods, we compare to e4e~\cite{tov2021designing}, pSp~\cite{richardson2021encoding} and EG3D-GOAE~\cite{yuan2023make}. 
For $\mathcal{W}+$ optimization, we optimize the latent code for 1K steps. For PTI, we first optimize the latent code $\hat{w} \in \mathcal{W}+$ for 1K steps and then fine-tune the generator for 1K steps. For Pose Opt. and SPI, we re-run their official implementation. For pSp, we employ the original training configuration from \cite{richardson2021encoding} with a batch size of 3. We train the pSp encoder on both FFHQ and synthetic data, similarly to our method. For EG3D-GOAE, we take the released checkpoint and run the inference on our dataset.

\begin{table}[h!]
    {\footnotesize
    \vspace{0cm}
    \centering
    \setlength\tabcolsep{0pt}
        \begin{tabular}{ccccc}
            Input
            &
            PTI
            &
            SPI
            &
            \textbf{Ours}
            &
            SfM
            \\
            
            &
            (119.34 s)
            &
            (258.84 s)
            &
            \textbf{(0.12 s)}
            &
            reconstruction
            \\
            \includegraphics[width=0.09\textwidth]{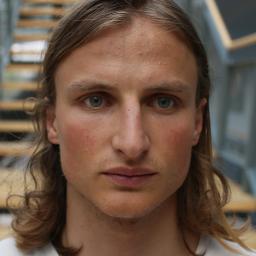}
            &
            \includegraphics[width=0.09\textwidth]{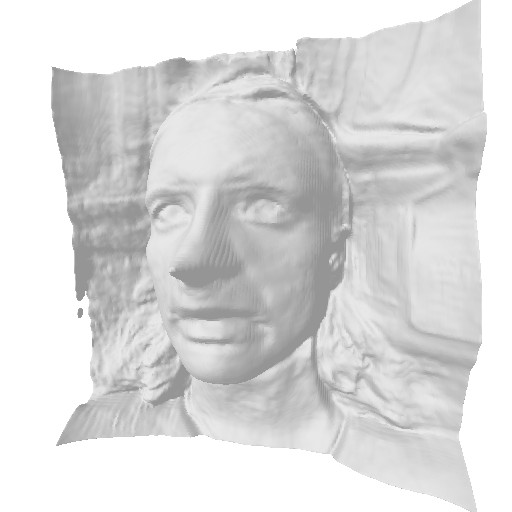}
            &
            \includegraphics[width=0.09\textwidth]{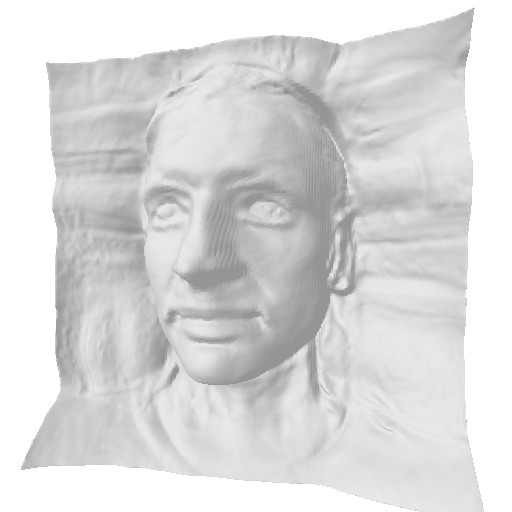}
            &
            \includegraphics[width=0.09\textwidth]{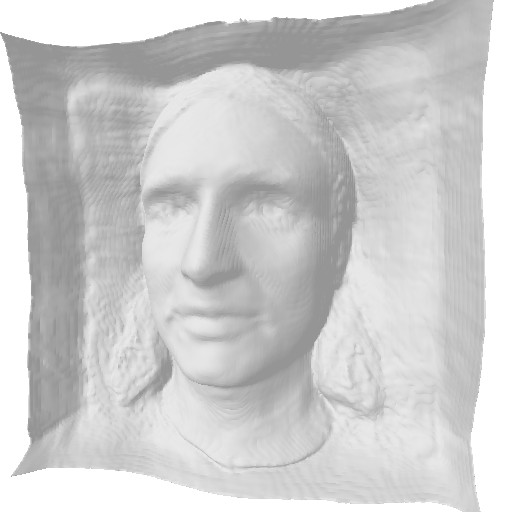}
            &
            \includegraphics[width=0.09\textwidth]{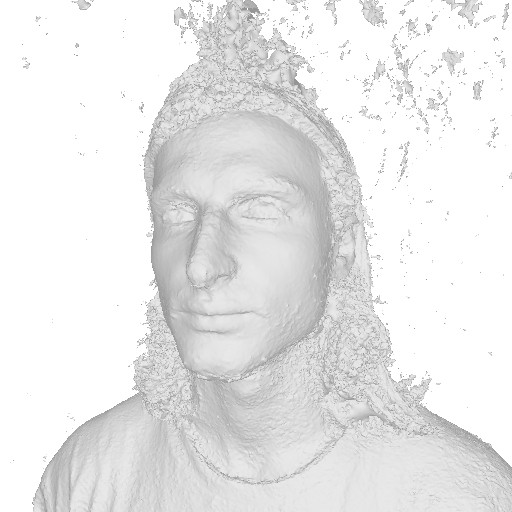}
            \\
            \includegraphics[width=0.09\textwidth]{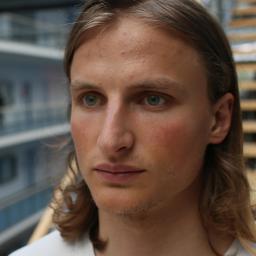}
            &
            \includegraphics[width=0.09\textwidth]{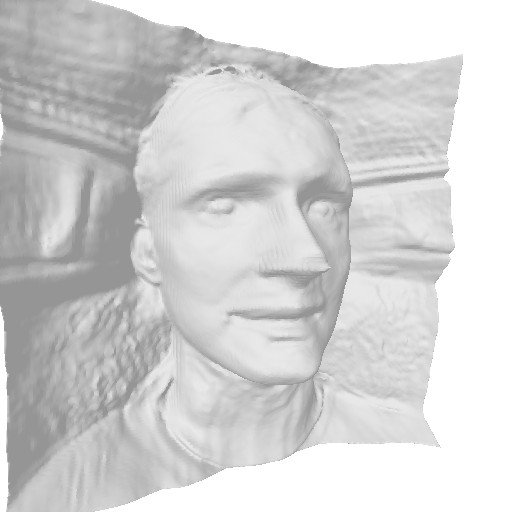}
            &
            \includegraphics[width=0.09\textwidth]{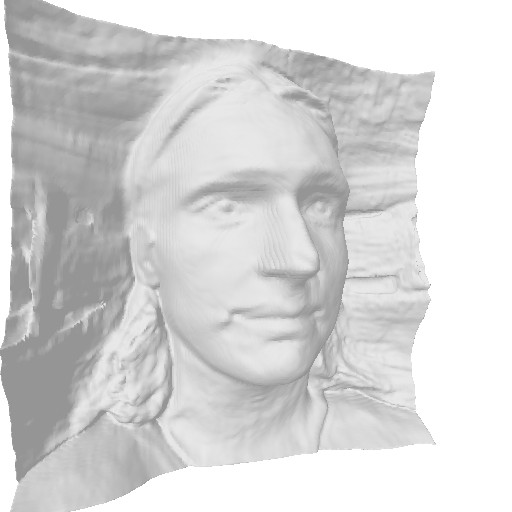}
            &
            \includegraphics[width=0.09\textwidth]{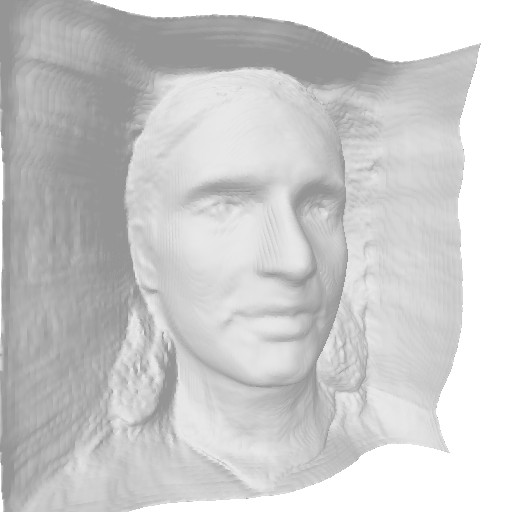}
            &
            \includegraphics[width=0.09\textwidth]{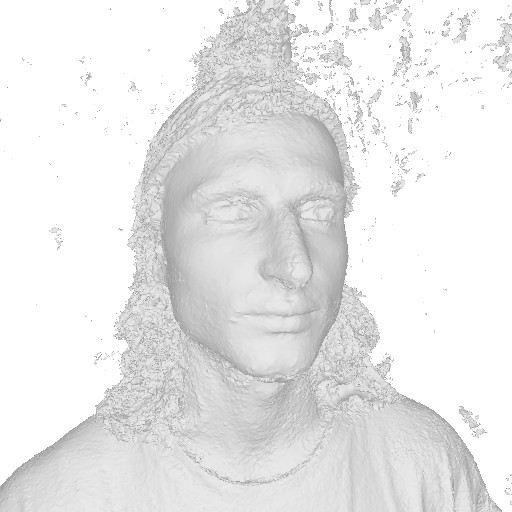}
            \\
        \end{tabular}%
    }
\captionof{figure}{
Comparison of the estimated 3D geometry w.r.t. the "ground-truth" reconstruction by Structure-from-Motion (SfM). Our method estimates the view-consistent embedding of a head in 3D from a single image. \textit{Electronic zoom-in recommended.}
}
\label{fig:ablation-geometry}
\vspace{-0.3cm}
\end{table}

\subsection{Results}\label{results}

\noindent \textbf{Comparison to the state-of-the-art.} We present the evaluation of our approach w.r.t.~the baselines in Fig.~\ref{visualcomp} and Table~\ref{quant-same-view-table}. 
Commonly used metrics MSE, LPIPS \cite{Zhang2018TheUE}, MS-SSIM \cite{ms-ssim}, and ID similarity \cite{Huang2020CurricularFaceAC} (measured by the pre-trained face recognition network not used in training) have been selected to analyze various aspects of perceptual similarity between inputs and corresponding reconstructions. To assess the quality of 3D geometry, we measure Depth MSE following a similar protocol from~\cite{Shi2020Lifting2S}. 
Among optimization-based techniques, SPI achieves the best results for same-view reconstruction. We attribute this to the almost three times longer optimization time than PTI and Pose Opt. 
However, our method outperforms all the baselines according to the Depth metric while being an order of magnitude times faster than optimization-based approaches. 
Effectively, this means that the head shape is closer to the one estimated by a parametric face prior model than for the other methods.

Furthermore, we demonstrate the identity preservation quality of input image re-rendering from a novel view in Table~\ref{novel-view-table} and in Fig.~\ref{novelview}.
We outperform all the baselines on extreme novel view yaw angles, and our method embeds the head in 3D space in a much more plausible way while preserving fine details and not relying on any explicit face or head priors.
For all methods in general, the ID score declines faster above a certain value of the yaw angle due to the uneven angle distribution in the FFHQ dataset.

\begin{table*}[h!]
    \begin{adjustbox}{minipage=0.63\linewidth,scale=1.0}
    {\fontsize{8pt}{0.35cm}
    \selectfont
    \begin{center}
        \caption{Quantitative ablation study for the loss, dataset, and architecture changes.}
        \vspace{-0.2cm}
        \begin{tabular}{l|c c c c |c| c c c | c c c}
            \multirow{3}{*}{Method} & \multirow{3}{*}{\,\,MSE $\downarrow$\,\,} & \multirow{3}{*}{LPIPS $\downarrow$} & \multirow{3}{*}{MS-SSIM $\uparrow$} &  \multirow{3}{*}{Depth $\downarrow$} & \multicolumn{6}{c}{ID $\uparrow$}\\
            \cline{6-12}
            & & & & & \,Same\, & \multicolumn{6}{c}{Novel View (Yaw angle in radians)} \\
            \cline{7-12}
            & & & & & View &\,\,-0.8\,\, & \,\,-0.6\,\, & \,\,-0.3\,\, & \,\,0.3\,\, & \,\,0.6\,\, & \,\,0.8\,\, \\ %
            \hline \hline
            Ours & 0.019 & \textbf{0.08} &  0.87 & 0.051 & 0.68 & 0.39 & 0.47 & 0.58 & 0.59 & 0.48 & 0.40 \\ %
            \dots\, \textit{(FFHQ)} & 0.022 & 0.09 & 0.86 & \textbf{0.044} & \textbf{0.70} & \textbf{0.41} & \textbf{0.48} & \textbf{0.60} & \textbf{0.60} & \textbf{0.50} & \textbf{0.42} \\ %
            \dots\, w/o $\mathcal{L}_m$ & \textbf{0.017} & \textbf{0.08} & \textbf{0.88} & 0.082 & 0.69 & 0.39 & 0.47 & 0.59 & 0.59 & 0.48 & 0.40 \\ %
            \dots\, $\mathcal{L}_m = .005$ & 0.018 & \textbf{0.08} & 0.87 & 0.068 & 0.69 & 0.38 & 0.46 & 0.59 & 0.60 &  0.48 & 0.40 \\ %
            \dots\, $\mathcal{L}_m = 0.5$ & 0.028 & 0.11 & 0.83 & 0.049 & 0.64 & 0.37 & 0.44 & 0.55 & 0.56 & 0.45 & 0.37 \\ %
            No $2^{nd}$ branch &  0.047 &  0.18 &  0.73 & 0.056 &  0.41 &  0.25 & 0.30 & 0.36 & 0.37 &  0.31 & 0.26 \\ %
            \dots\, \textit{(FFHQ)} & 0.047 & 0.18 & 0.74 & 0.051 & 0.44 & 0.28 & 0.33 & 0.39 & 0.40 & 0.34 & 0.29 \\ %
            \dots\, w/o $\mathcal{L}_m$ & 0.045 & 0.18 & 0.73 & 0.076 & 0.40 & 0.24 & 0.28 & 0.35 & 0.36 & 0.29 & 0.25 \\ %
            \dots\, $\mathcal{L}_m = .005$ & 0.045 & 0.18 & 0.73 & 0.068 & 0.39 & 0.24 & 0.28 & 0.34 & 0.36 & 0.30 & 0.25 \\ %
            \dots\, $\mathcal{L}_m = 0.5$ & 0.057 & 0.20 & 0.70 & 0.053 & 0.41 & 0.25 & 0.29 & 0.37 & 0.37 & 0.30 & 0.25 \\ %
            \hline
        \end{tabular}
        \label{ablation-table}
    \end{center}
    }
    \end{adjustbox}
    \hfill
    \hskip 0.5cm
    \begin{adjustbox}{minipage=0.30\linewidth,scale=1.0}
        \setlength{\tabcolsep}{0pt}
        \renewcommand{\arraystretch}{0}
        \begin{tabularx}{\columnwidth}{cccc}
            & {\footnotesize Input}   & {\footnotesize Ours (FFHQ)}  & {\footnotesize Ours} \\
            \raisebox{1.5\height}{\rotatebox[origin=c]{90}{\footnotesize Original}}\, & \includegraphics[height=1.7cm]{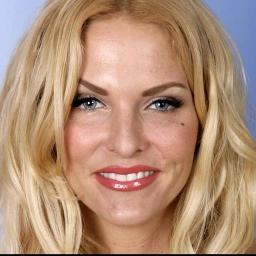} &
            \includegraphics[height=1.7cm]{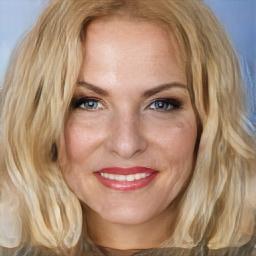} &
            \includegraphics[height=1.7cm]{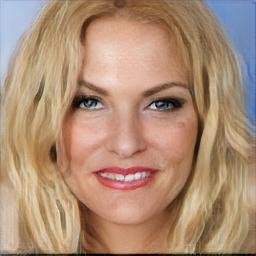} \\
            \raisebox{1.2\height}{\rotatebox[origin=c]{90}{\footnotesize Shifted up}}\, & \includegraphics[height=1.7cm]{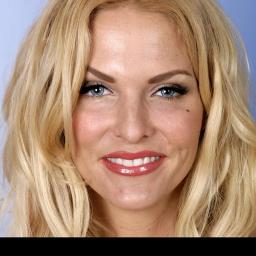} &
            \includegraphics[height=1.7cm]{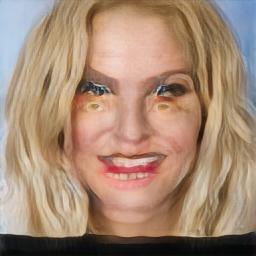} &
            \includegraphics[height=1.7cm]{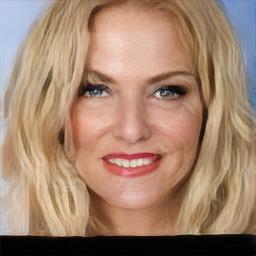} \\
        \end{tabularx}
        \captionof{figure}{Adding synthetic samples to training (\textit{Ours}) makes the model robust to the input image shifts compared to trained on FFHQ only (\textit{Ours (FFHQ)}).}%
        \label{fig:ablation-dataset}
    \end{adjustbox}
\end{table*}

\noindent \textbf{Ablation study.}
In Fig.~\ref{fig:ablation} and Table~\ref{ablation-table}, we ablate over the possible differences in our model design, such as loss functions %
weights and the presence of the second branch.
As some of those were introduced to handle occluded regions in the input view, we demonstrate visually how the incorporated symmetric prior affects the novel view and 3D geometry quality. 
All models in this ablation except the first branch encoder are trained for 600K steps. We observed that including symmetric prior significantly improves novel-view quality and geometric consistency. Despite the model trained on FFHQ achieving a higher ID score, in Fig.~\ref{fig:ablation-dataset}, we show that adding synthetic EG3D samples in training allows us to correctly canonicalize face for a slightly shifted input, while our model trained with only real images is very sensitive to minor image misalignment.

\noindent \textbf{Geometry evaluation for a multi-view sequence.} In Fig.~\ref{fig:ablation-geometry}, we present a qualitative evaluation of the geometry obtained using our method against PTI~\cite{roich2022pivotal}, SPI~\cite{yin20223d}, and ground truth geometry obtained using 
a third-party SfM software~\cite{agisoft}. We notice that the estimated geometry is highly view-consistent and is closer to the ground truth.

We present more evaluations and visual results in the Appendix, such as novel view synthesis for a talking head video and face manipulation capabilities.

\subsection{PTI and tri-plane offsets behavior}

Both our method and optimization- and encoder-based baselines can be decomposed into two stages: estimating the latent code and the delta for the generator parameters. In Fig.~\ref{analysis:fig}, we show how combining these steps, each performed either by optimization (\textcolor{RawSienna}{opt.}) or an encoder (\textcolor{blue}{pred.}), influences the inversion behavior. 

\textcolor{RawSienna}{$\mathcal{W}+$ opt.} inverts a single image and cannot account for 3D geometry due to the lack of supervision from other views, which results in incorrectly stretched geometry.
Accordingly, the same happens with PTI = (\textcolor{RawSienna}{$\mathcal{W}+$ opt.} + \textcolor{RawSienna}{EG3D params opt.}) method. Interestingly, tri-plane prediction, applied on top of \textcolor{RawSienna}{$\mathcal{W}+$ opt.}, can alleviate the damage to the geometry caused by \textcolor{RawSienna}{$\mathcal{W}+$ opt}.

\textcolor{blue}{$\mathcal{W}+$ pred.} by a pSp encoder, on the contrary, embeds the head in 3D more plausibly due to the supervision from images under different poses during training. At the same time, the same-view quality is marginally worse than PTI. Applying the PTI's second step (\textcolor{RawSienna}{EG3D params opt.}) helps improve it significantly; however, it incorrectly modifies head proportions, similar to the \textcolor{RawSienna}{$\mathcal{W}+$ opt.} behavior. To investigate this effect further, instead of optimizing EG3D parameters after \textcolor{blue}{$\mathcal{W}+$ pred.}, we try optimizing the tri-plane offsets directly, and this fully cancels the incorrect stretching of geometry while preserving high fidelity in the same view. Since both \textcolor{RawSienna}{EG3D params opt.} and \textcolor{RawSienna}{tri-plane opt.} are performed for a single image (i.e.~without multi-pose supervision during training), this may indicate that offsetting the tri-planes is more spatially restrictive and thus stable. 
Therefore, we base our method on directly leveraging the tri-plane representation.

We further improve the checkerboard artifacts in novel view, noticeable for \textcolor{RawSienna}{tri-plane opt.}, by \textcolor{blue}{tri-plane prediction}, and improve the embedding in 3D space by a symmetric prior. 

%% file: sections/05_discussion.tex
\section{Conclusion}
We present a novel approach for EG3D inversion that achieves high-quality reconstructions with view consistency and can be run in close to real time on modern GPUs. 
We also show that directly utilizing tri-plane representation better estimates 3D structure compared to other approaches while preserving identity in the novel view.
Although our method achieves compelling results and is on par with optimization-based approaches, %
both visually and quantitatively, it has certain limitations.
For instance, it is limited by the range of yaw angles shown to EG3D during training and cannot model the background depth.
In addition, there is room for improvement of the temporal consistency and for supporting input images with extreme head poses.

%% file: sections/06_acknowledgments.tex
\section*{Acknowledgments}
This work was supported by the ERC Starting Grant Scan2CAD (804724) and the German Research Foundation (DFG) Research Unit "Learning and Simulation in Visual Computing." We also thank Yawar Siddiqui, Guy Gafni, Shivangi Aneja, and Simon Giebenhain for participating in the sample videos and helpful discussions.

%% file: supplement.tex
\section{Implementation Details}
\label{supp:impdetails}
\noindent \textbf{Dataset Details.} Our model is trained on a combination of real images from FFHQ and generated samples from EG3D. 
As shown in the main text, adding EG3D samples makes the model robust to the input image shifts. 
The synthetic training samples are generated via sampling latent codes $z$ for EG3D with no truncation ($\psi = 1$) often applied for large-scale GANs~\cite{brock2018large}, thus including the hard samples. 
In order to match the camera pose distribution in the FFHQ, we generate EG3D samples with randomly sampled camera poses from FFHQ and their flipped versions. 
In Table~\ref{eg3d-samples-table}, we demonstrate the dependence of the reconstruction quality on CelebA-HQ on the number of synthetic samples, created by EG3D in advance added to the dataset.\\ 

\noindent \textbf{Experiment settings.} For training our models, we adopt the same training configuration from \cite{richardson2021encoding} except for some minor modifications. In particular, we train the second branch only after 20K steps and then train both branches until 500K. Afterward, we freeze the first branch and fine-tune the second branch until 1.5M steps. In each training step, we re-render the batch of input images from the same view and the mirror view. Then, we compute input view reconstruction losses using same-view rendered images and mirror-view losses $\mathcal{L}_m$ using mirror-view rendered images. We operate in the resolution of $256 \times 256$ except for the calculation of $\mathcal{L}_{id}$. Specifically, the region around the face is cropped and resized to $112 \times 112$ before feeding into the face recognition network \cite{Deng2018ArcFaceAA} to calculate $\mathcal{L}_{id}$. The models are trained with a batch size of 3. We use the Ranger optimizer that combines Rectified Adam \cite{Liu2019OnTV} with the Lookahead technique \cite{Zhang2019LookaheadOK} and set a learning rate to 0.0001. The models are trained using a single NVIDIA GeForce RTX A6000 GPU.\\

\noindent \textbf{Loss functions.} As outlined in the main text, we train our models using same-view reconstruction and mirror-view losses. The loss function for our first branch latent encoder $\phi(.)$ is defined as:
\begin{equation}
    \mathcal{L}_{\phi}(x, x_m, \hat{y}, \hat{y}_m) = \mathcal{L}_{rec}(x, \hat{y}) + \lambda_m\mathcal{L}_{m}(x_m, \hat{y}_m) 
\end{equation}
where $x_m = \textrm{flip}(x)$, $\mathcal{L}_{rec}(x, \hat{y})$ is defined as
\begin{equation}
    \begin{aligned}
        \mathcal{L}_{rec}(x, \hat{y}) = \lambda_1\mathcal{L}_2(x, \hat{y}) + \lambda_2\mathcal{L}_\textrm{LPIPS}(x, \hat{y}) + \\
        \lambda_3\mathcal{L}_\textrm{id}(x, \hat{y})
    \end{aligned}
    \label{supp:loss1}
\end{equation}
and $\mathcal{L}_{m}(x_m, \hat{y}_m)$ is a \textit{probably symmetric prior} defined as
\begin{equation}
    \begin{aligned}
        \mathcal{L}_{m}(x_m, \hat{y}_m) = \lambda_4\mathcal{L}_\textrm{symm}(x_m, \hat{y}_m, \sigma(x_m)) + \\ \lambda_5\mathcal{L}_\textrm{LPIPS}(x_m, \hat{y}_m)  
        + \lambda_6\mathcal{L}_\textrm{id}(x_m, \hat{y}_m)
    \end{aligned}
    \label{supp:loss2}
\end{equation}
\begin{table}[t!]
    \centering
    \renewcommand{\arraystretch}{0}
    \resizebox{0.49\textwidth}{!}{%
        \begin{tabular}{cc@{\hskip 0.6cm}cc}
            Input
            &
            Uncertainty
            &
            Input
            &
            Uncertainty
            \\
            &
            Map
            &
            &
            Map
            \\[0.3cm]
            \includegraphics[width=0.16\textwidth]{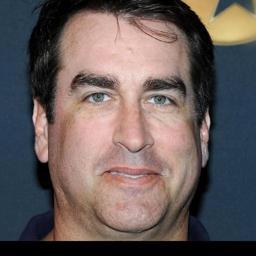}
            &
            \includegraphics[width=0.16\textwidth]{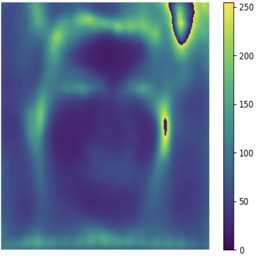}
            &
            \includegraphics[width=0.16\textwidth]{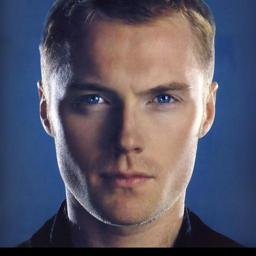}
            &
            \includegraphics[width=0.16\textwidth]{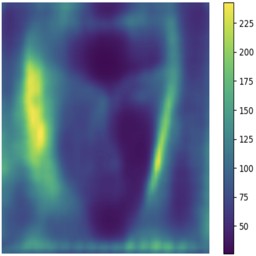}
            \\
        \end{tabular}%
}
\captionof{figure}{Uncertainty map $\sigma$ predicted by the pre-trained network from~\cite{Wu2019UnsupervisedLO} for the given input. The network assigns higher uncertainty to regions such as ears, hair, and background where the symmetry assumption fails.}
\label{fig:methodsigmamap}
\vspace{-0.3cm}
\end{table}
The main text shows that $\mathcal{L}_{m}$ significantly improves the embedding in 3D space. However, directly applying $\mathcal{L}_2$ between the mirror-view image and the surrogate mirrored image is not applicable since human faces are not perfectly symmetric.         
Therefore, following the practice outlined in~\cite{Wu2019UnsupervisedLO}, we construct $\mathcal{L}_\textrm{symm}'(x_m, \hat{y}_m, \sigma(x_m))$ as a penalty between mirrored image $x_m$ and reconstruction for the mirror image $\hat{y}_m$ weighted by a pixel-wise uncertainty map $\sigma(x_m)$ computed for each pixel and taking an average. Mathematically,
\begin{equation}\label{eq:unsup}
\begin{aligned}
    & \mathcal{L}_\textrm{symm}'(x_m, \hat{y}_m, \sigma(x_m)) \\
    & = -\frac{1}{|\Omega|}\sum_{uv \in \Omega}\log\frac{1}{\sqrt{2}(\sigma(x_m))_{uv}}\exp-\frac{\sqrt{2}\ell_{\textrm{1}, uv}}{(\sigma(x_m))_{uv}}\\
    & = \log(\sqrt{2}) +
    \frac{1}{|\Omega|}\sum_{uv \in \Omega}\log(\sigma(x_m))_{uv} + \frac{\sqrt{2}\ell_{\textrm{1}, uv}}{(\sigma(x_m))_{uv}}
\end{aligned}
\end{equation}
where $\ell_{\textrm{1}, uv}$ is the $\mathcal{L}_1$ distance between the intensity of pixels at location $uv$, and $\sigma(x_m)$ is estimated by the neural network for image $x_m$. We can interpret the loss function as the negative log-likelihood of a factorized Laplacian distribution on the reconstruction residuals. We take pre-trained network from~\cite{Wu2019UnsupervisedLO} for predicting uncertainty map $\sigma(x_m)$ and replace $\mathcal{L}_1$ distance with $\mathcal{L}_2$ in (\ref{eq:unsup}). Therefore, we reformulate $\mathcal{L}_\textrm{symm}'(x_m, \hat{y}_m, \sigma(x_m))$ as
\begin{equation}\label{eq:unsup1}
    \mathcal{L}_\textrm{symm}(x_m, \hat{y}_m, \sigma(x_m)) = \frac{1}{|\Omega|}\sum_{uv \in \Omega}\frac{\ell_{\textrm{2}, uv}}{(\sigma(x_m))_{uv}}
\end{equation}
However, we can also train the prediction network from scratch, optimizing the likelihood in (\ref{eq:unsup}). $\sigma(x_m)$ assigns lower confidence to the region in the mirrored image $x_m$ where the symmetry assumption fails (see Fig.~\ref{fig:methodsigmamap}). The uncertainty map predictor network is an encoder-decoder architecture that operates in the 64 $\times$ 64 resolution. Therefore, we resize the image to 64 $\times$ 64 before feeding into this network and upsample the output back to 256 $\times$ 256 for calculating $\mathcal{L}_\textrm{symm}$.

We use AlexNet \cite{Krizhevsky2012CNN} to extract features for the $\mathcal{L}_\textrm{LPIPS}$ loss. Similarly, $\mathcal{L}_\textrm{id}$ is computed by measuring the cosine similarity between the input image and the output with a pre-trained ArcFace~\cite{Deng2018ArcFaceAA} network. We set the weight of each component in the loss function as follows: $\lambda_m=0.1$, $\lambda_1=\lambda_4=1.0$, $\lambda_2=\lambda_5=1.0$ and $\lambda_3=\lambda_6=0.1$. Analogously, we construct the loss for the second branch $\mathcal{L}_{\psi}$ by replacing $\mathcal{L}_2$ with $\mathcal{L}_1$ smooth loss in (\ref{supp:loss1}), inside ${L}_\textrm{symm}$ in (\ref{supp:loss2}) and first branch outputs $\hat{y}$ and $\hat{y}_m$ with the second branch outputs $y$ and $y_m$. We use the same weight for each component.\\

\begin{table*}[h!]
    \caption{Quantitative ablation study over the number of synthesized EG3D samples in the training set.}
    \vspace{-0.4cm}
    \begin{center}
    \resizebox{\linewidth}{!}{
        \begin{tabular}{l|c c c c |c| c c c | c c c}
            \hline
            \multirow{3}{*}{Our Method} & \multirow{3}{*}{\,\,MSE $\downarrow$\,\,} & \multirow{3}{*}{LPIPS $\downarrow$} & \multirow{3}{*}{MS-SSIM $\uparrow$} &  \multirow{3}{*}{Depth $\downarrow$} & \multicolumn{6}{c}{ID $\uparrow$}\\
            \cline{6-12}
            & & & & & \,Same\, & \multicolumn{6}{c}{Novel View (Yaw angle in radians)} \\
            \cline{7-12}
            & & & & & View &\,\,-0.8\,\, & \,\,-0.6\,\, & \,\,-0.3\,\, & \,\,0.3\,\, & \,\,0.6\,\, & \,\,0.8\,\, \\ %
            \hline \hline
            \dots\, w/ 0 EG3D samples\,\, & 0.022 & 0.09 & 0.86 & \textbf{0.044} & \textbf{0.70} & \textbf{0.41} & \textbf{0.48} & \textbf{0.60} & \textbf{0.60} & \textbf{0.50} & \textbf{0.42}\\
            \dots\, w/ 10K EG3D samples & 0.020 & 0.09 & 0.86 & 0.048 & 0.68 & 0.39 & 0.46 & 0.58 & 0.58 & 0.48 & 0.40\\
            \dots\, w/ 50K EG3D samples & 0.021 & 0.09 & 0.86 & 0.047 & 0.68 & 0.39 & 0.47 & 0.58 & 0.59 & 0.48 & 0.40\\
            \dots\, w/ 100K EG3D samples & \textbf{0.019} & \textbf{0.08} & \textbf{0.87} & 0.051 & 0.68 & 0.39 & 0.47 & 0.58 & 0.59 & 0.48 & 0.40 \\
            \dots\, w/ 150K EG3D samples & 0.021 & 0.09 & 0.86 & 0.053 & 0.66 & 0.37 & 0.44 & 0.56 & 0.57 & 0.47 & 0.40\\
            \hline
        \end{tabular}
    }
    \end{center}
    \label{eg3d-samples-table}
    \vspace{-0.5cm}
\end{table*}

\noindent \textbf{First branch architecture.} To implement the latent encoder, we adopt the design of the pSp encoder from ~\cite{richardson2021encoding}. As the EG3D generator expects 14 styles vectors for the selected resolution, we modify the pSp architecture to output 14 styles vectors instead of 18. We employ IR-SE-50 \cite{Deng2018ArcFaceAA} pre-trained for face recognition for the backbone network.\\

\noindent \textbf{Second branch architecture.} The tri-plane offsets predictor consists of an encoder and a decoder network, a typical U-Net \cite{Ronneberger2015UNetCN} architecture. The encoder backbone is an IR-SE-50 \cite{Deng2018ArcFaceAA} pre-trained on face recognition, accelerating convergence. We adopt the design of the RUNet \cite{Hu2019RUNet} for the decoder with some minor modifications. Instead of using ReLU as in RUNet, we use PReLU \cite{He2015DelvingDI} with a separate $\alpha$ for each input channel and an initial value of 0.25. Like RUNet, every step in the decoder path consists of upsampling, concatenation, and convolution operations. Upsamping of the feature map is performed with a PyTorch nearest neighbor upsample layer (\textit{torch.nn.Upsample}). Then, it is followed by a concatenation with the intermediate feature maps from the encoder path. The intermediate features are extracted from the encoder's 3rd, 7th, 21st, and 22nd layers. Finally, batch normalization, $3 \times 3$ convolution, PReLU, $3 \times 3$ convolution, and PReLU are applied sequentially. The final step in the decoder path takes the concatenation of first branch tri-plane features with upsampled features from the previous step as an input and outputs $256 \times 256 \times 96$ tri-plane offsets. The final step applies $3 \times 3$ convolution, PReLU, $3 \times 3$ convolution, PReLU, and $1 \times 1$ convolution operations sequentially.

\section{Novel view rendering of videos}
\label{supp:talkinghead}
We demonstrate an application of our method to render in-the-wild videos from a novel view. In Fig.~\ref{video:novel-view-supp}, frames of a video with a person talking and their rendering from a fixed novel view in the EG3D space are presented. The background in the video was removed by a matting network \cite{ke2022modnet}. The encoder is capable of representing tiny details of in-the-wild portrait imagery in 3D and supports complex facial expressions.

\section{Facial Manipulation}
\label{supp:facialmanipulation}
To perform image editing, we first obtain the latent code $w \in \mathcal{W}$ of the input image via optimization. Since $\mathcal{W}$ space offers more editing power than $\mathcal{W+}$~\cite{tov2021designing}, we select $\mathcal{W}$ space for our experiments. We then obtain the final inversion using our second branch by replacing first branch components with components obtained using optimization. Given the latent code $w \in \mathcal{W}$, tri-plane features ($\boldsymbol{G}(w) + \Delta \boldsymbol{T}$) and edited latent code $w_{edit}$, we can render edited image with camera matrix $\pi$ by $\mathcal{R}(\boldsymbol{T}_{edit}, \pi)$. Inspired by \cite{xuyao2022}, we perform following operation to obtain $\boldsymbol{T}_{edit}$:
\begin{equation}
    \boldsymbol{T}_{edit} = (\boldsymbol{G}(w) + \Delta \boldsymbol{T}) + \boldsymbol{G}(w_{edit}) - \boldsymbol{G}(w) 
\end{equation}
We take two editing directions, smile, and age, from the official implementation of \cite{ko20233d}, obtained using GANspace \cite{Hrknen2020GANSpaceDI} and show editing results in Figs.~\ref{supp:editing1} and \ref{supp:editing2}. Note that our first branch latent encoder could be modified slightly to embed the input image in $\mathcal{W}$ instead of space $\mathcal{W}+$ as done in \cite{dinh2022hyperinverter} and \cite{alaluf2022hyperstyle}.
We demonstrate, however, that modification of the tri-plane features that would correspond to a certain semantic direction is possible and leave the research on the most plausible face manipulation for both input images and videos as a suggestion for future work.

\section{Discussion of the baselines design}
\begin{figure*}[!t]
    \centering
    \begin{subfigure}[t]{0.49\textwidth}
        \centering
        \includegraphics[width=\textwidth]{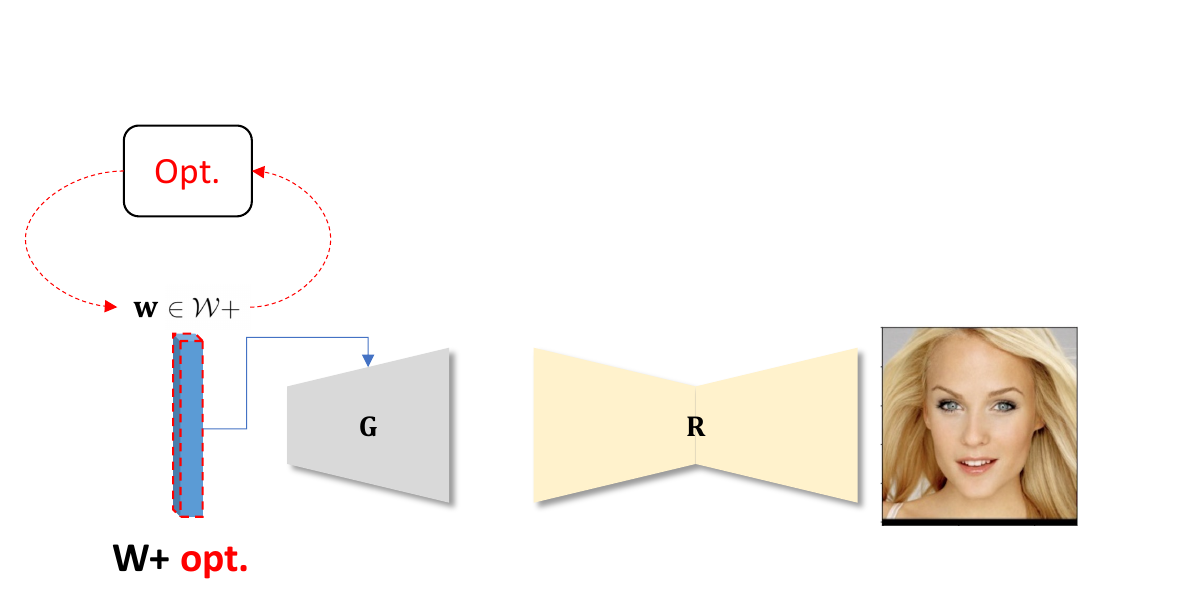}
        \caption{\textcolor{RawSienna}{$\mathcal{W}+$ opt.}}
    \end{subfigure}
    \hfill
    \begin{subfigure}[t]{0.49\textwidth}
        \centering
        \includegraphics[width=\textwidth]{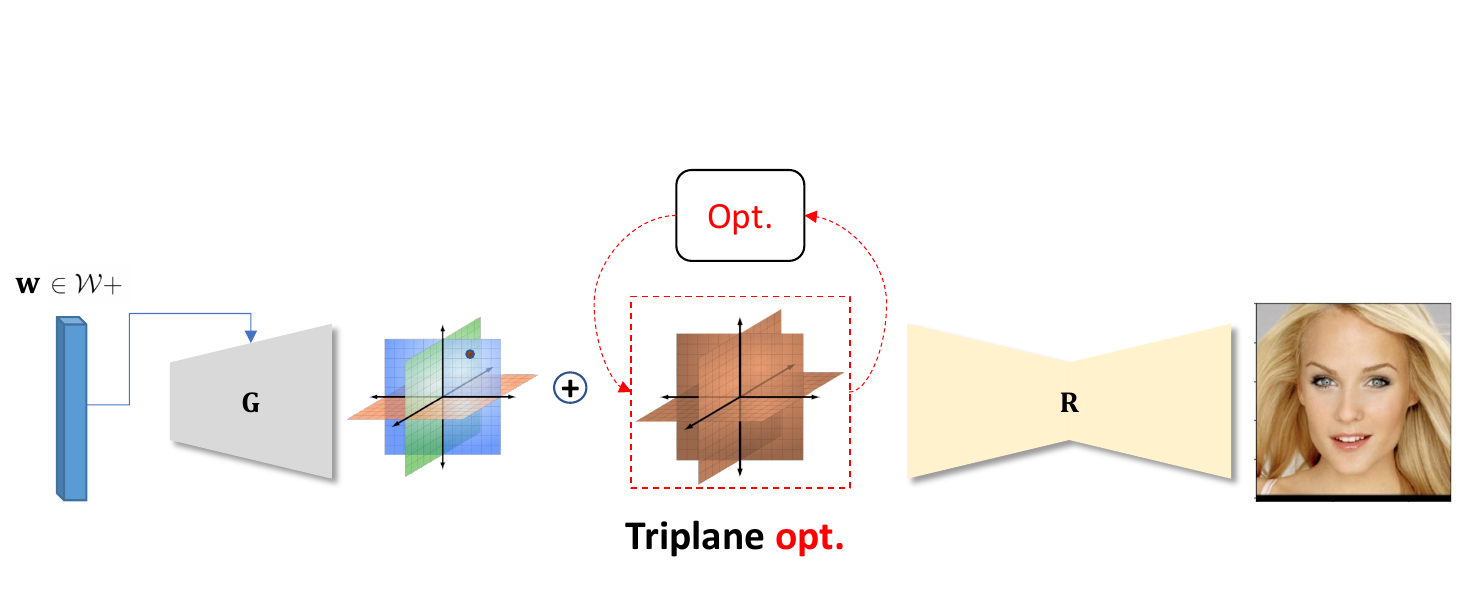}
        \caption{\textcolor{RawSienna}{Tri-plane opt.}}
    \end{subfigure}

    \begin{subfigure}[t]{0.49\textwidth}
        \centering
        \includegraphics[width=\textwidth]{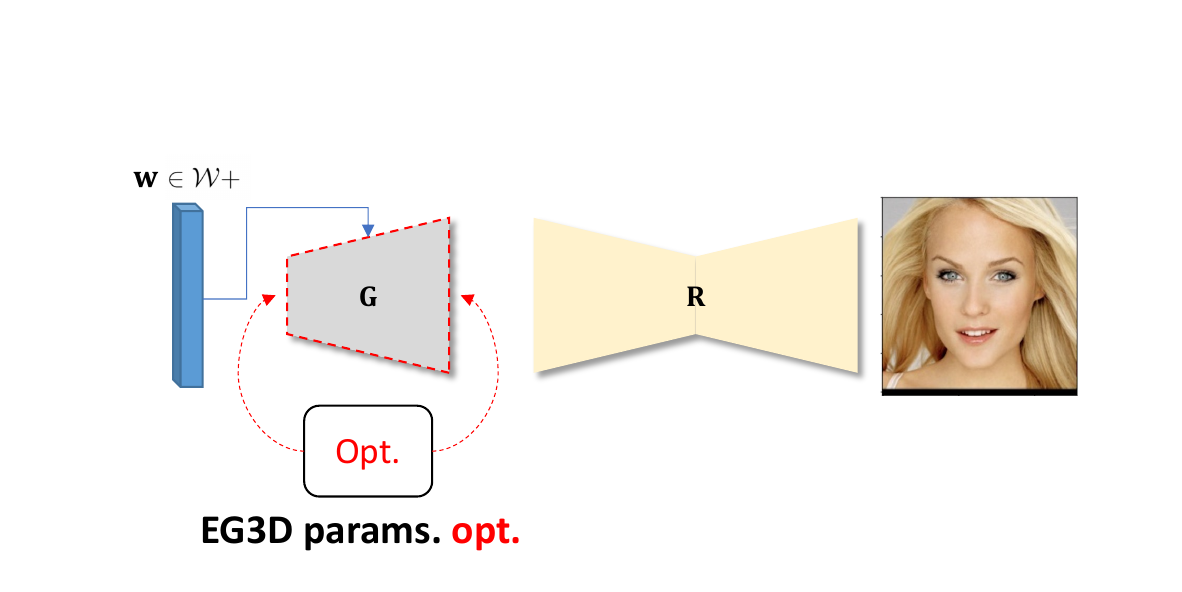}
        \caption{\textcolor{RawSienna}{EG3D params opt.}}
    \end{subfigure}
    \hfill
    \begin{subfigure}[t]{0.49\textwidth}
        \centering
        \includegraphics[width=\textwidth]{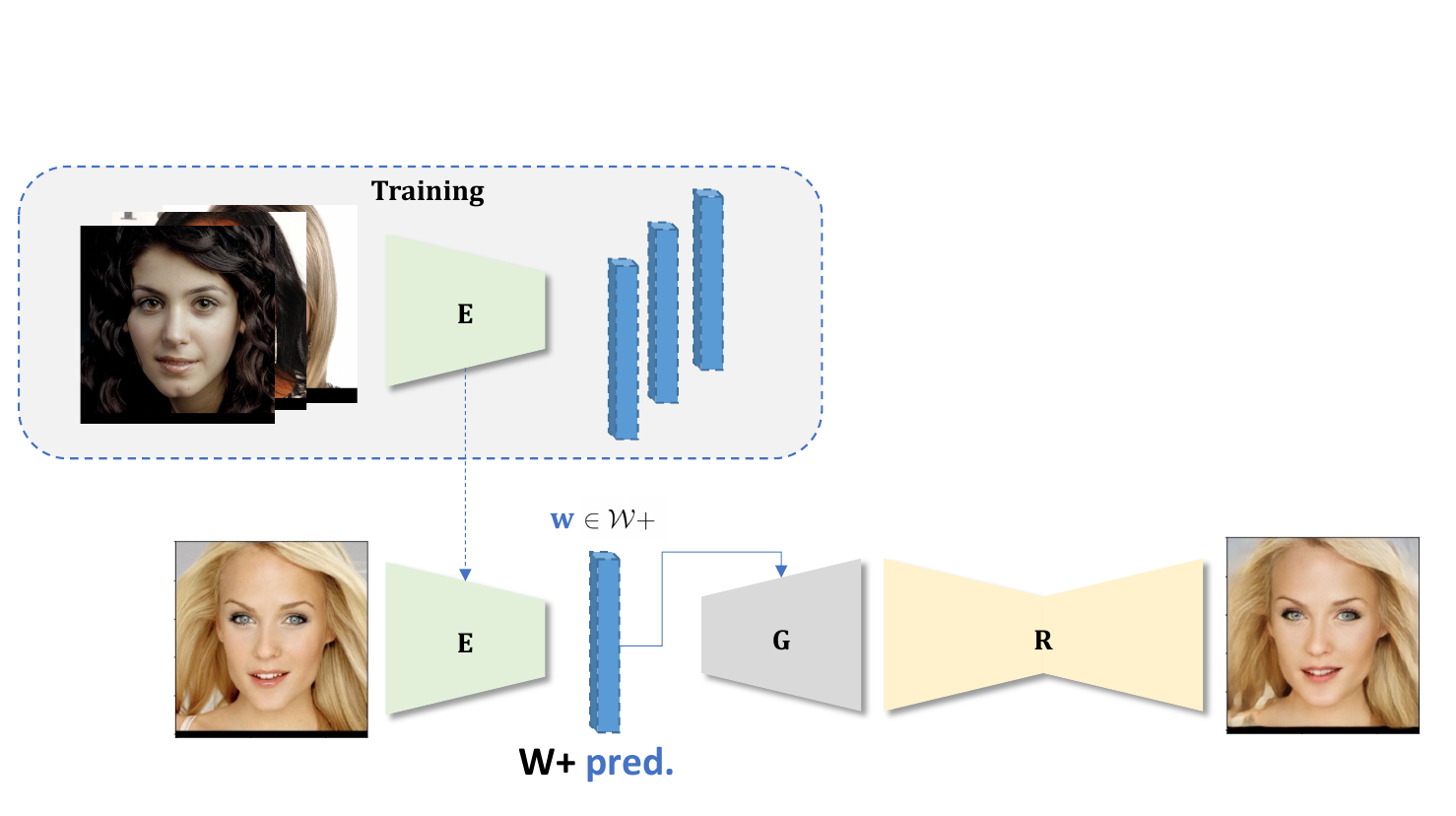}
        \caption{\textcolor{blue}{$\mathcal{W}+$ pred.} (pSp~\cite{richardson2021encoding})}
    \end{subfigure}
    \caption{Overview of our baseline approaches. $\boldsymbol{G}(\cdot)$ and $\mathcal{R}(\cdot)$ stand for EG3D generator and renderer blocks respectively. Hybrid approaches described in the main text constitute the combination of the techniques shown above, applied sequentially one after another. For instance, PTI~\cite{roich2022pivotal} sequentially performs (a) and then (c)., and \textcolor{blue}{$\mathcal{W}+$ pred.} + \textcolor{RawSienna}{tri-plane opt.} sequentially performs (d) and then (b).}
    \label{fig:baselines}
\end{figure*} In Fig.~\ref{fig:baselines}, we provide a visual overview of the baseline designs used for the analysis of PTI \cite{roich2022pivotal} and tri-plane offsets behavior. One-stage inversion techniques can be divided into two approaches: optimization-based and based on an encoder prediction. Two-stage inversion techniques involve inference followed by fine-tuning of some of the generator parameters. Similarly, these parameters can be fine-tuned via optimization or by encoder prediction. Since our method involves the prediction of the tri-plane offsets and avoids fine-tuning the generator parameters, we also consider the baseline where the tri-plane offsets in the second stage are optimized. 
In the main paper text, we demonstrate the design of different hybrid two-stage inversion approaches that combine both optimization and prediction. 

For \textcolor{RawSienna}{$\mathcal{W}+$ opt.}, we optimize the latent code $w \in \mathcal{W}+$ for 1K steps following \cite{karras2020analyzing}. The \textcolor{blue}{$\mathcal{W}+$ pred.} constitutes the baseline with the latent code $w \in \mathcal{W}+$ predicted by pSp encoder~\cite{richardson2021encoding}. For \textcolor{RawSienna}{EG3D params opt.}, we apply the second stage of PTI from \cite{roich2022pivotal} and optimize for 1K steps. To optimize for the tri-plane offsets (\textcolor{RawSienna}{tri-plane opt.}), we use L-BFGS~\cite{liu1989limited} as the optimizer and employ combination of $L_2$ or LPIPS~\cite{Zhang2018TheUE} with regularization term ($L_2$ and LPIPS discrepancy with the first branch prediction) as a loss objective. We run the optimization for 50 steps. As L-BFGS approximates the Hessian by calculating several estimates in a single step, 50 steps take %
equivalently 1K gradient evaluations.

\section{Geometric evaluation}

In order to evaluate how well the method embeds a head into 3D without any information about the head's geometry, we compare the prediction to the true head geometry constructed by a Structure-from-Motion method (see paragraph "\textbf{Geometry evaluation for a multi-view sequence}" in Sec.~4.2) for two subjects. The reconstruction is based on a $360^\circ$ DSLR capture, while the methods make predictions for the image of the sequence with the head pose closest to the straight frontal. The sequence for subject \#1 is the same as demonstrated in Fig.~7 in the main text.
We ridigly align the meshes by 5 eyes, nose, and mouth landmarks to the SfM mesh and analyze the proximity of each predicted mesh to the SfM mesh in the face region (the bounding region is defined as an ellipsoid in 3D with the same location and size for all methods). We deliberately only select 5 landmarks for alignment to analyze the shape correctness of the parts not fully visible in the frontal image, such as cheeks. In Fig.~\ref{fig:meshlab}, we demonstrate the overlay of the mesh predicted by TriPlaneNet and true SfM mesh, as well as a comparison to the meshes obtained from other methods. As shown in the view behind, parts with only partial presence in the frontal view get predicted more correctly by our method; Fig.~\ref{fig:AD}, and Table~\ref{table:AD} demonstrate that analytically via pixel-wise proximity to the SfM mesh.

\begin{table}[h!]
    \centering
    \resizebox{.48\textwidth}{!}{
        \begin{tabular}{l|ccccc}
                    & \multicolumn{5}{c}{AD $\downarrow$}                                                                                         \\ \cline{2-6} 
                    & \multicolumn{1}{l}{} & \multicolumn{1}{l}{} & \multicolumn{1}{l}{} &       Ours w/o    & \multicolumn{1}{l}{} \\
                    & PTI                  & SPI                  & pSp                  & symm. prior & Ours                 \\ \hline
        Subject \#1\,\, & \,\,0.971\,\,                & \,\,0.360\,\,                & \,\,0.115\,\,                & \,\,0.167\,\,       & \textbf{0.090}       \\
        Subject \#2\,\, & \,\,0.032\,\,                & \,\,0.024\,\,                & \,\,0.016\,\,                & \,\,0.019\,\,            &  \textbf{0.009}                    \\ \hline
        \end{tabular}
    }
    \caption{Comparison of the average absolute distance (AD) from the mesh for various methods' predictions to the true geometry from SfM.}
    \label{table:AD}
    \vspace{-0.2cm}
\end{table}

\begin{table}[h!]
    \setlength{\tabcolsep}{0pt}
    \setlength\extrarowheight{-14pt}
    \begin{tabular}{cccc}
    \multicolumn{2}{c}{True geometry from SfM} & \multicolumn{2}{c}{Ours} \\
    \multicolumn{2}{c}{\includegraphics[width=0.24\textwidth,trim={0 3.5cm 0 2.3cm},clip]{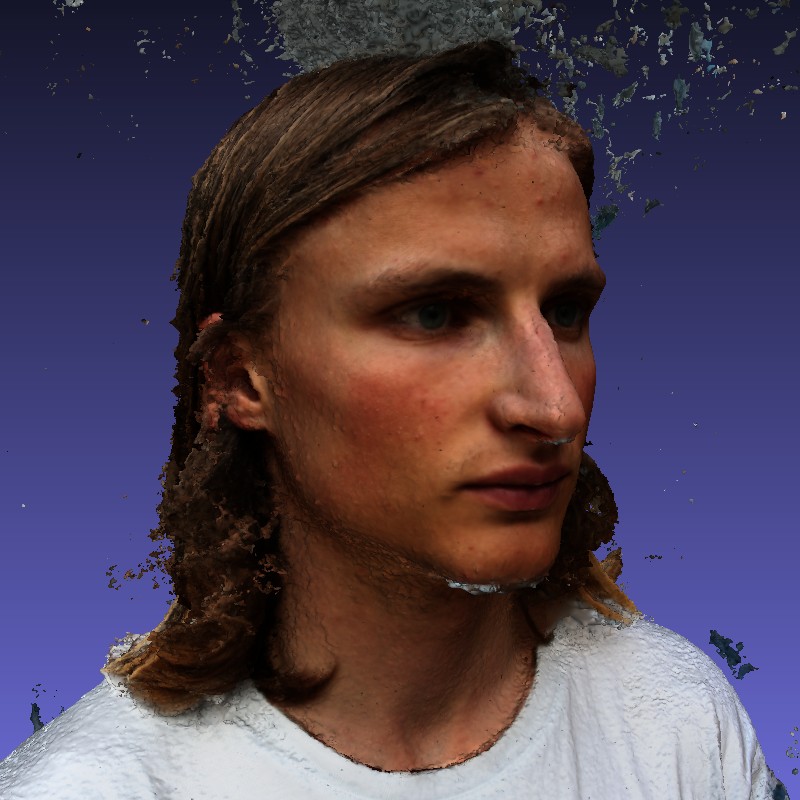}} &
    \multicolumn{2}{c}{\includegraphics[width=0.24\textwidth,trim={0 3.5cm 0 2.3cm},clip]{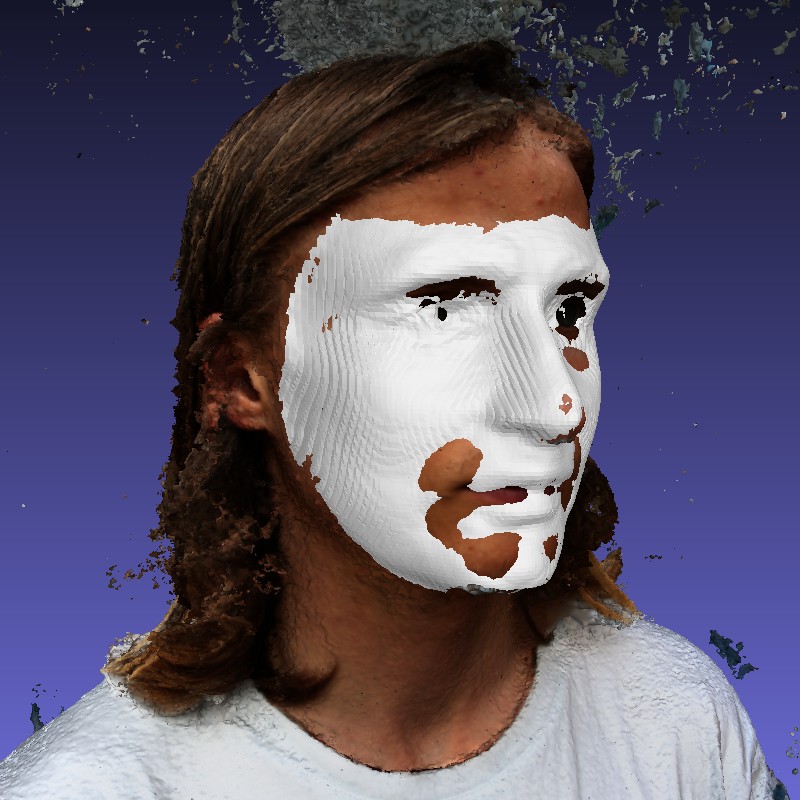}} \\
     \\
    {\footnotesize SfM} & {\footnotesize pSp} & {\footnotesize Ours w/o symm} & {\footnotesize Ours} \\
    \includegraphics[width=0.12\textwidth,trim={15cm 2cm 3cm 2cm},clip]{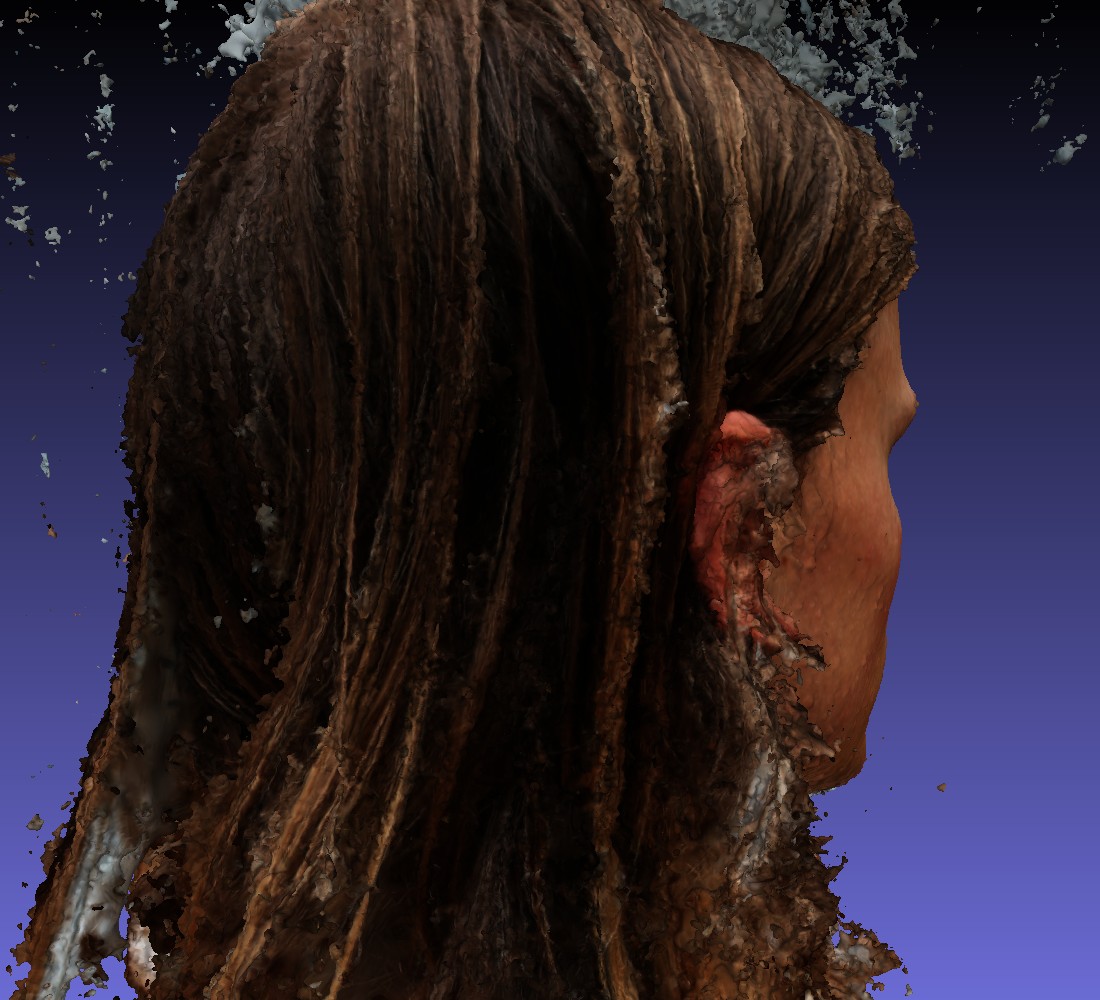} &
    \includegraphics[width=0.12\textwidth,trim={15cm 2cm 3cm 2cm},clip]{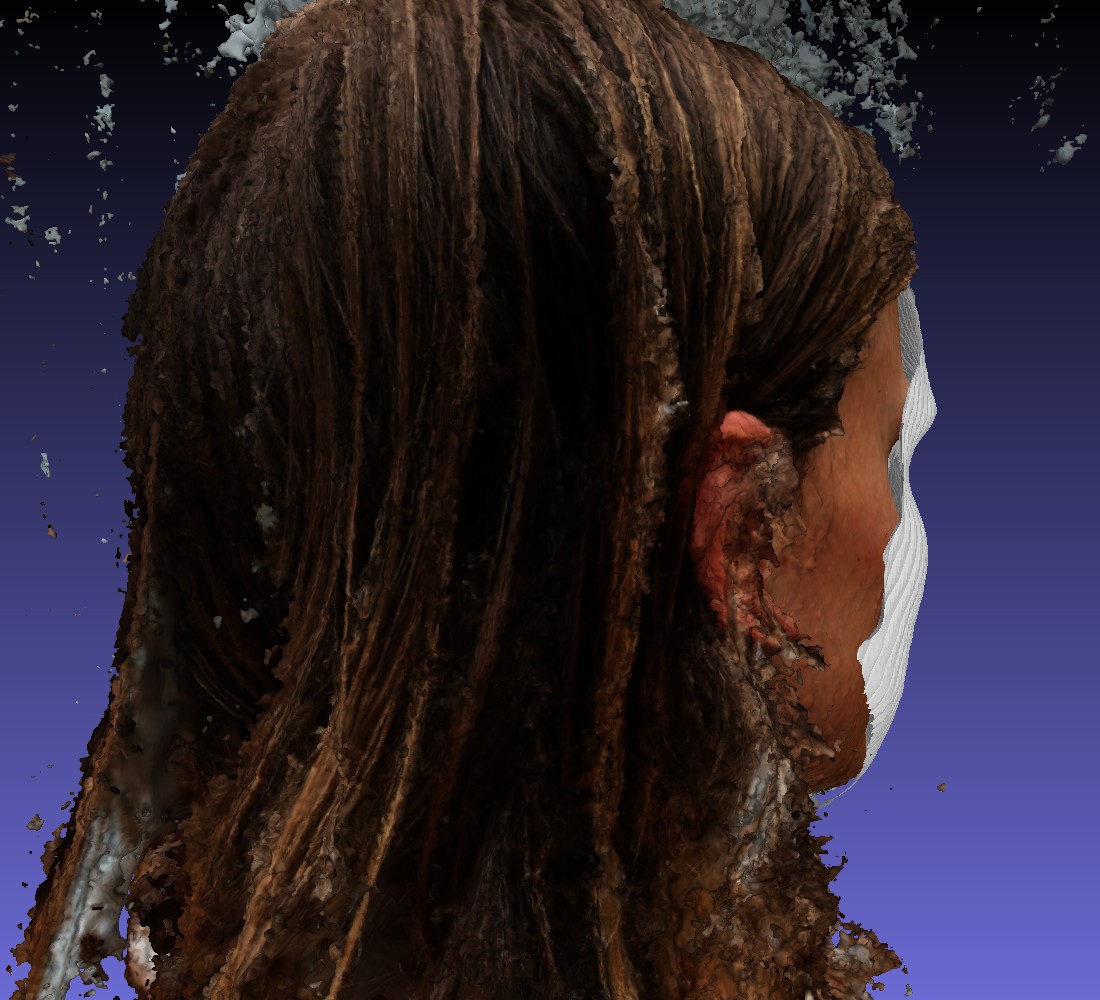} &
    \includegraphics[width=0.12\textwidth,trim={15cm 2cm 3cm 2cm},clip]{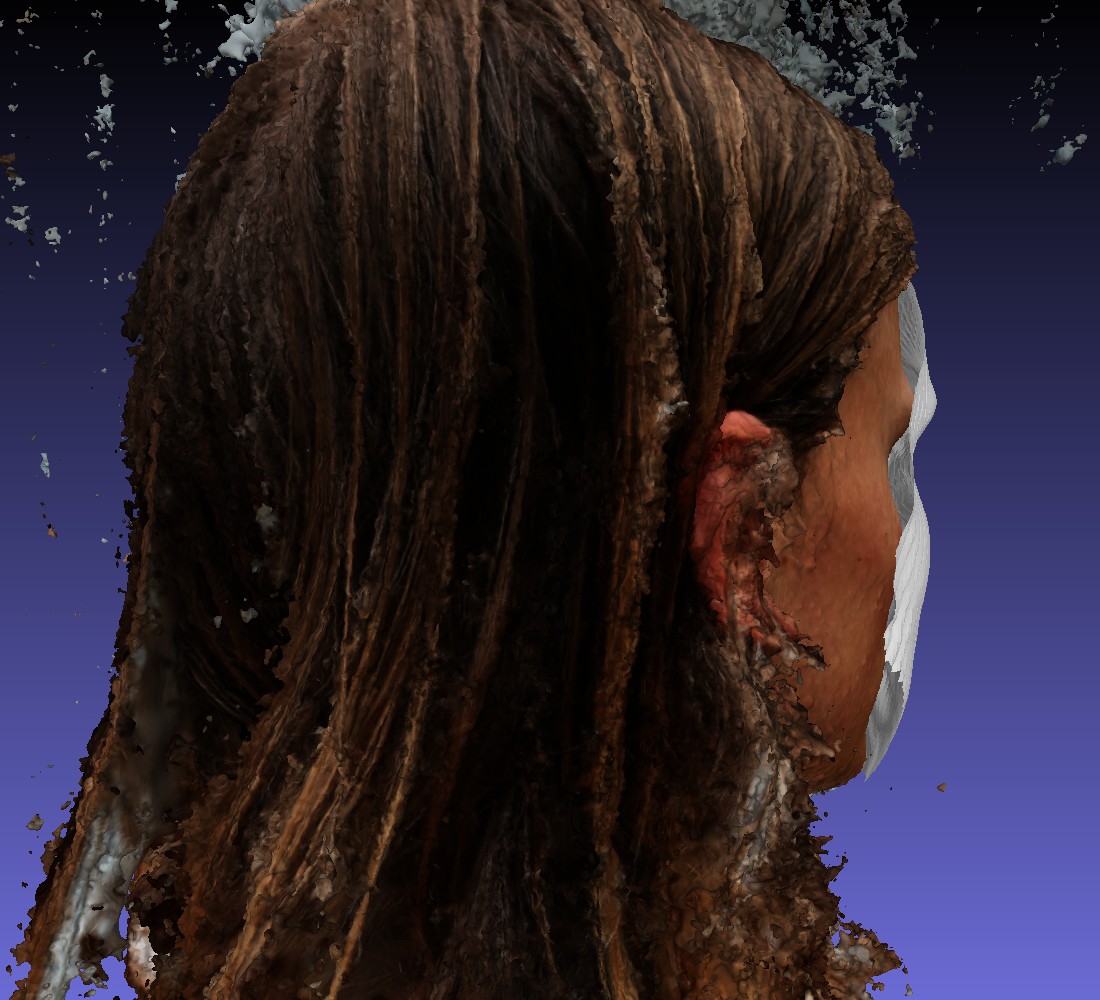} &
    \includegraphics[width=0.12\textwidth,trim={15cm 2cm 3cm 2cm},clip]{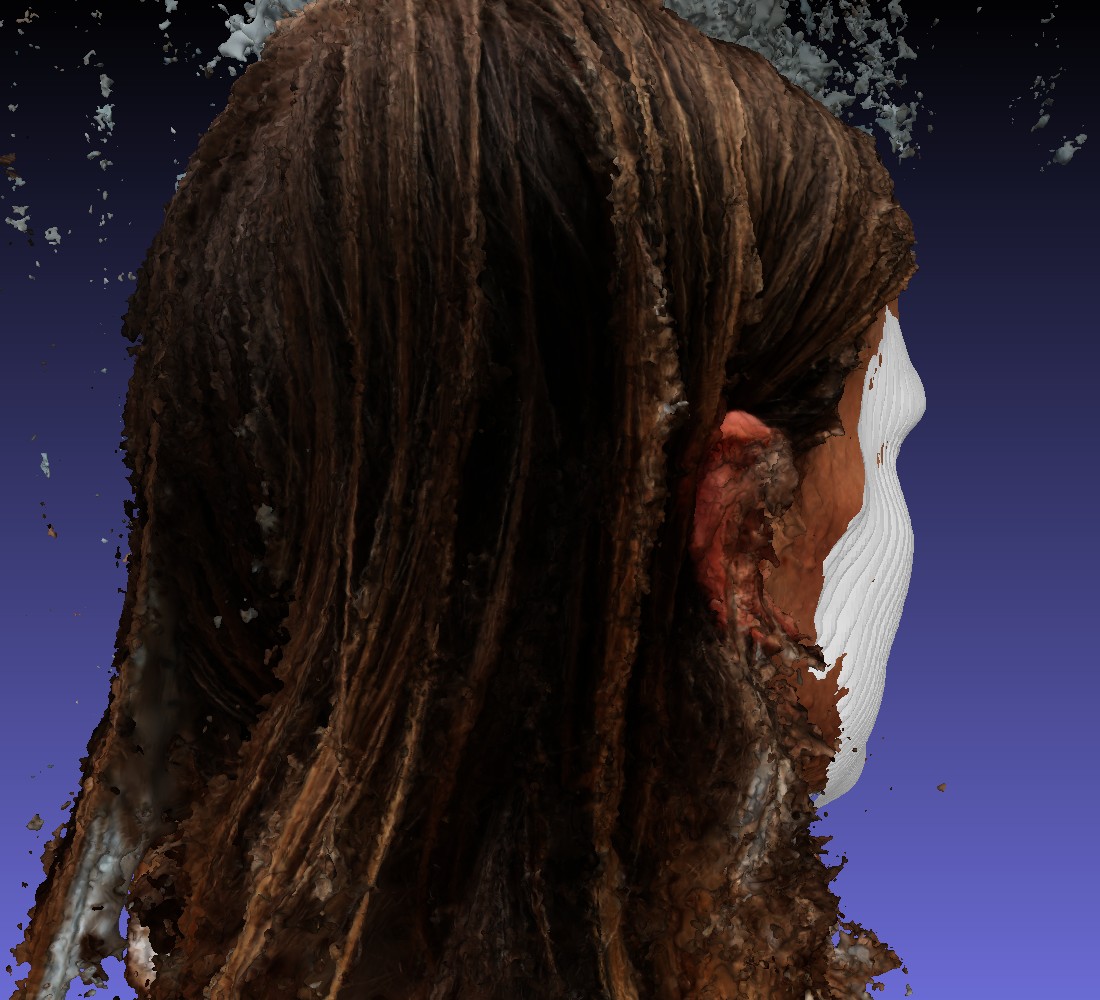} \\
    \end{tabular}
    \vspace{-0.2cm}
    \captionof{figure}{Overlay of the ground truth SfM mesh and predicted meshes. As observed in the bottom row (view from behind), \textit{Ours} provides the tightest fit to the ground truth SfM mesh. PTI and SPI produce unnaturally wide meshes that mostly lie inside the true geometry, except for the nose region (the discrepancy can be better observed in Fig.~\ref{fig:AD}).}
    \label{fig:meshlab}
\end{table}

\begin{table}[h!]
    \centering
    \setlength{\tabcolsep}{0pt}
    \setlength\extrarowheight{-14pt}
    \resizebox{0.45\textwidth}{!}{%
    \begin{tabular}{cccccc}
    {\footnotesize PTI} & {\footnotesize SPI} & {\footnotesize pSp} 
    & {\footnotesize Ours} & {\footnotesize Ours} & 
    \multirow{3}{*}{\includegraphics[width=0.07\textwidth,trim={13cm 2cm 1cm 1cm},clip]{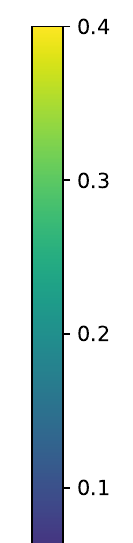}} \\
        &     &     & {\footnotesize w/o symm.} &   & \\
    \\
    \includegraphics[width=0.075\textwidth,trim={4.3cm 0cm 5.3cm 1.5cm},clip]{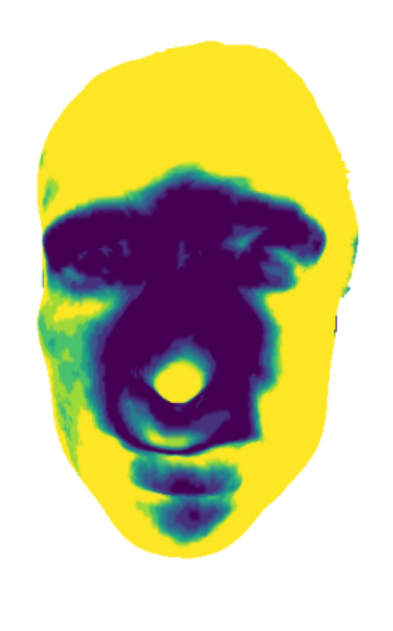} &
    \includegraphics[width=0.075\textwidth,trim={4.3cm 0cm 5.3cm 1.5cm},clip]{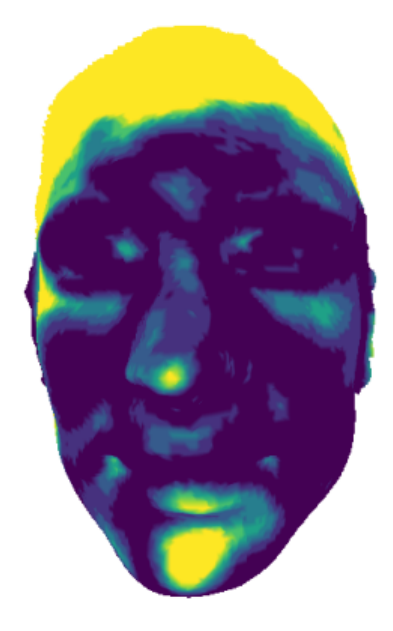} &  
    \includegraphics[width=0.075\textwidth,trim={4.3cm 0cm 5.3cm 1.5cm},clip]{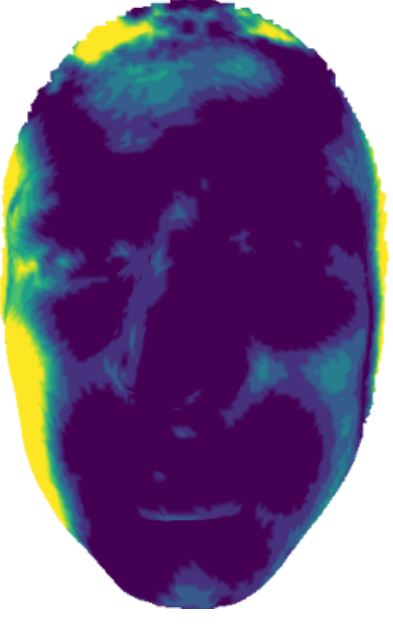} &  
    \includegraphics[width=0.075\textwidth,trim={4.3cm 0cm 5.3cm 1.5cm},clip]{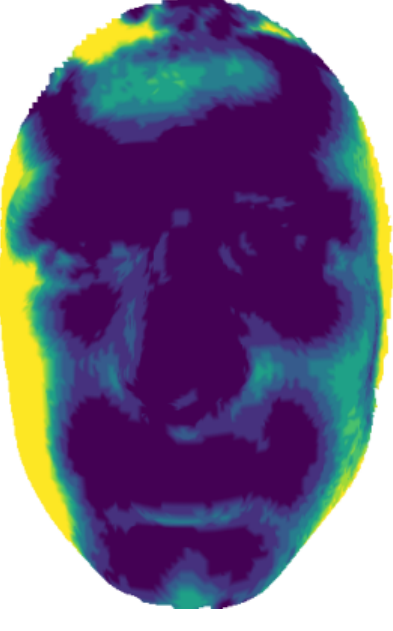} &  
    \includegraphics[width=0.075\textwidth,trim={4.3cm 0cm 5.3cm 1.5cm},clip]{images/rebuttal/flatness/seq1_triplanenet.pdf} &  \\
    & & & & & \\
    \\
    \includegraphics[width=0.075\textwidth,trim={7.5cm 0cm 3cm 2cm},clip]{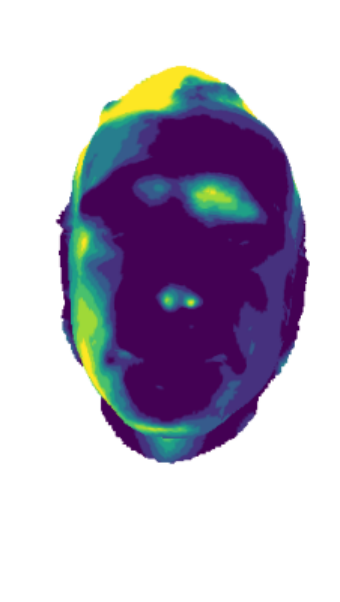} &
    \includegraphics[width=0.075\textwidth,trim={7.5cm 0cm 3cm 2cm},clip]{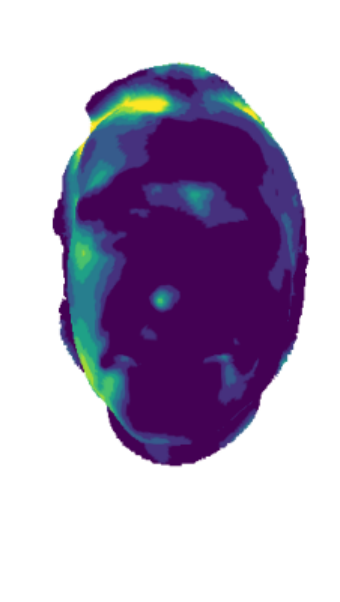} &  
    \includegraphics[width=0.075\textwidth,trim={7.5cm 0cm 3cm 2cm},clip]{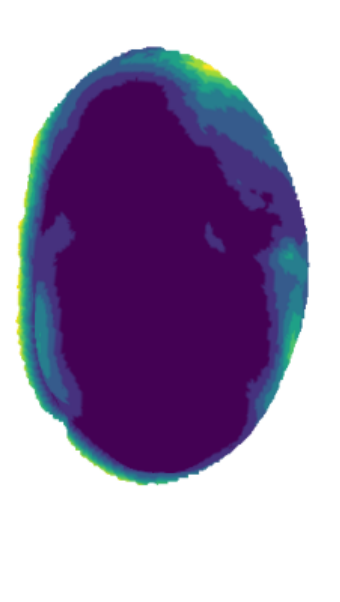} &  
    \includegraphics[width=0.075\textwidth,trim={7.5cm 0cm 3cm 2cm},clip]{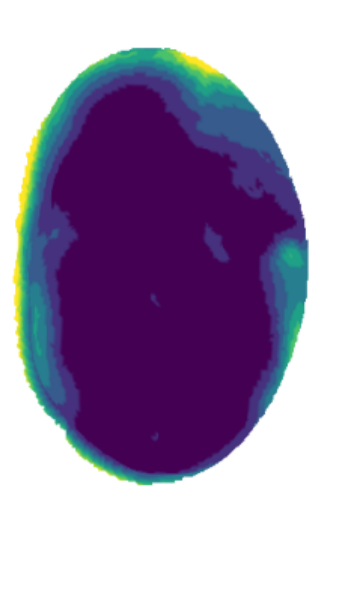} &  
    \includegraphics[width=0.075\textwidth,trim={7.5cm 0cm 3cm 2cm},clip]{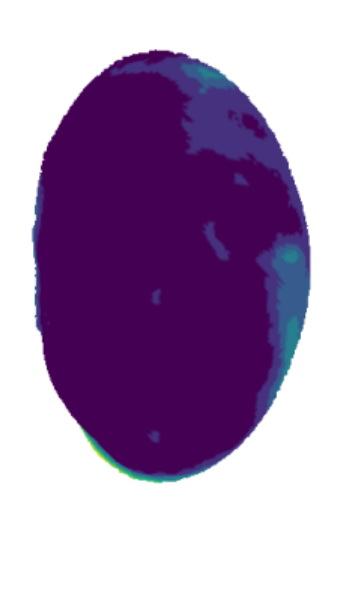} &  
    \end{tabular}
    }
    \captionof{figure}{Absolute distance (fraction of inter-ocular dist.) from the predicted meshes points to their nearest neighbors in the SfM mesh (top row: Subject \#1; bottom row: Subject \#2). Lower values (dark blue) indicate that the predicted mesh is closer to the true geometry, while higher values (yellow) indicate that the predicted mesh is too far (inside or outside) from the true geometry.}
    \label{fig:AD}
\end{table}

\begin{table}[h!]
    \centering
    \setlength{\tabcolsep}{0pt}
    \setlength\extrarowheight{-14pt}
    \resizebox{0.41\textwidth}{!}{%
    \begin{tabular}{cc}
         \includegraphics[width=0.2075\textwidth,trim={0 0 0 0},clip]{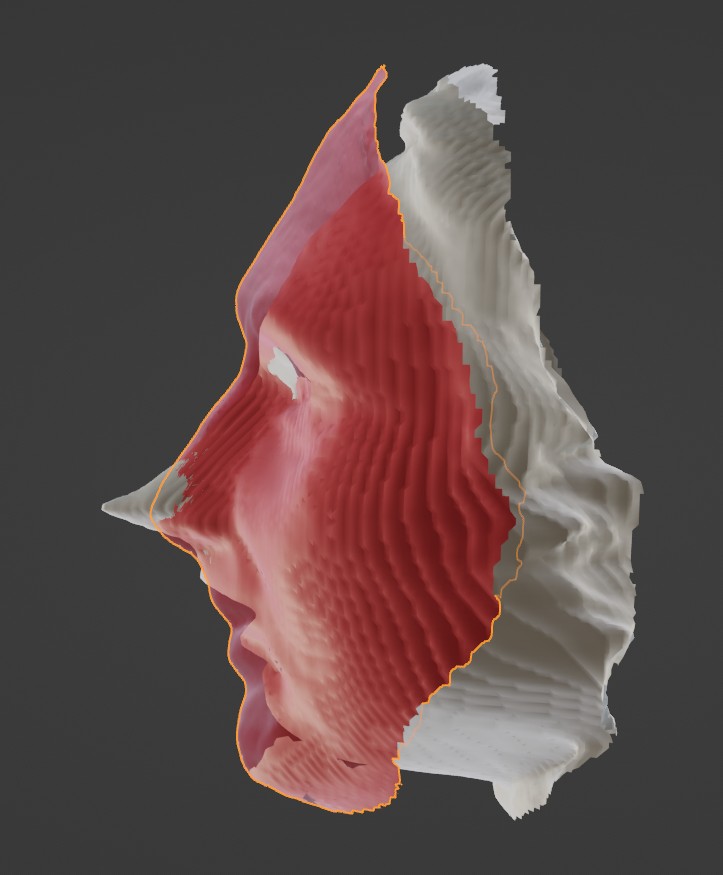} \hspace{0.1cm} &
         \includegraphics[width=0.2\textwidth,trim={0 0 0 1.2cm},clip]{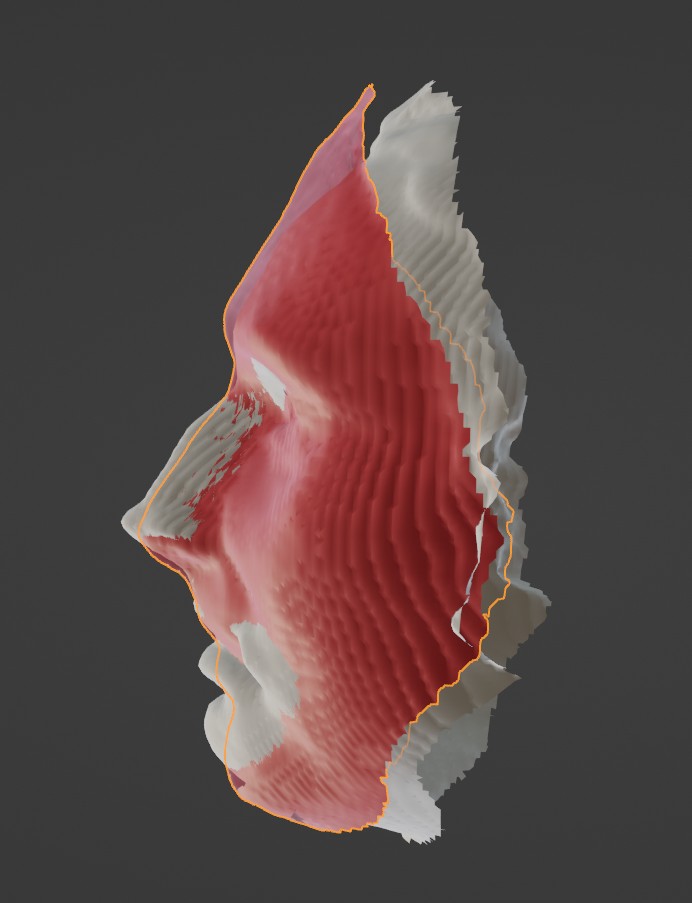} \\
    \end{tabular}
    }
    \captionof{figure}{Overlay of the shape predicted by our method (red) and: PTI (left image, gray) and SPI (right image, gray). Here we demonstrate that these shapes, especially for PTI, are typically wider in the side projection than for our method, which introduces a discrepancy with the true geometry (see Fig.~\ref{fig:AD}). The same effect is demonstrated in Fig.~6 in the main text.}
    \label{fig:PTI_SPI_transparent_overlay}
\end{table}

\begin{table}[h!]
    \centering
    \setlength{\tabcolsep}{0pt}
    \setlength\extrarowheight{-14pt}
    \resizebox{0.41\textwidth}{!}{
    \begin{tabular}{cc}
         \includegraphics[width=0.2075\textwidth,trim={0 0 0 0},clip]{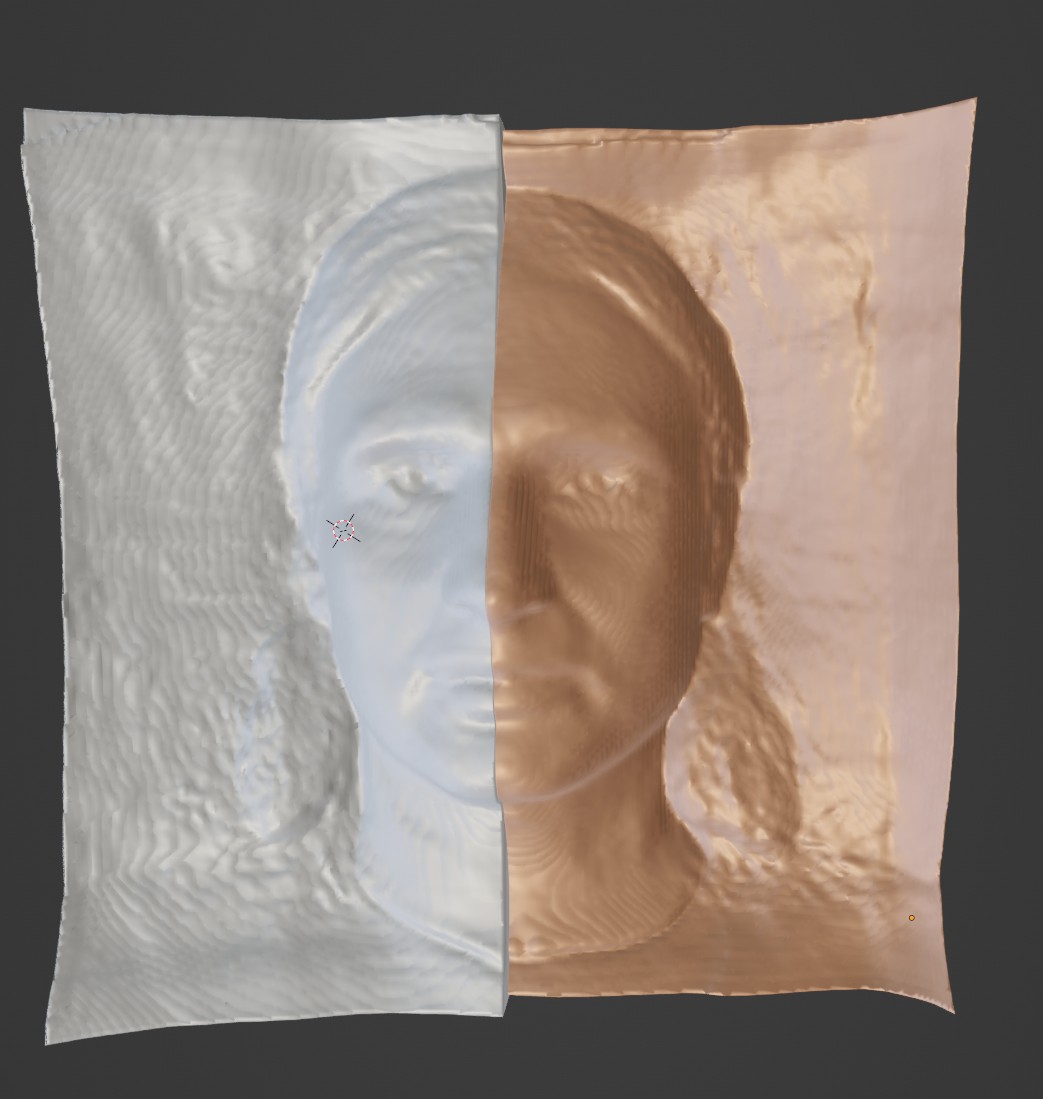} \hspace{0.1cm} &
         \includegraphics[width=0.2\textwidth,trim={0 0 0 0},clip]{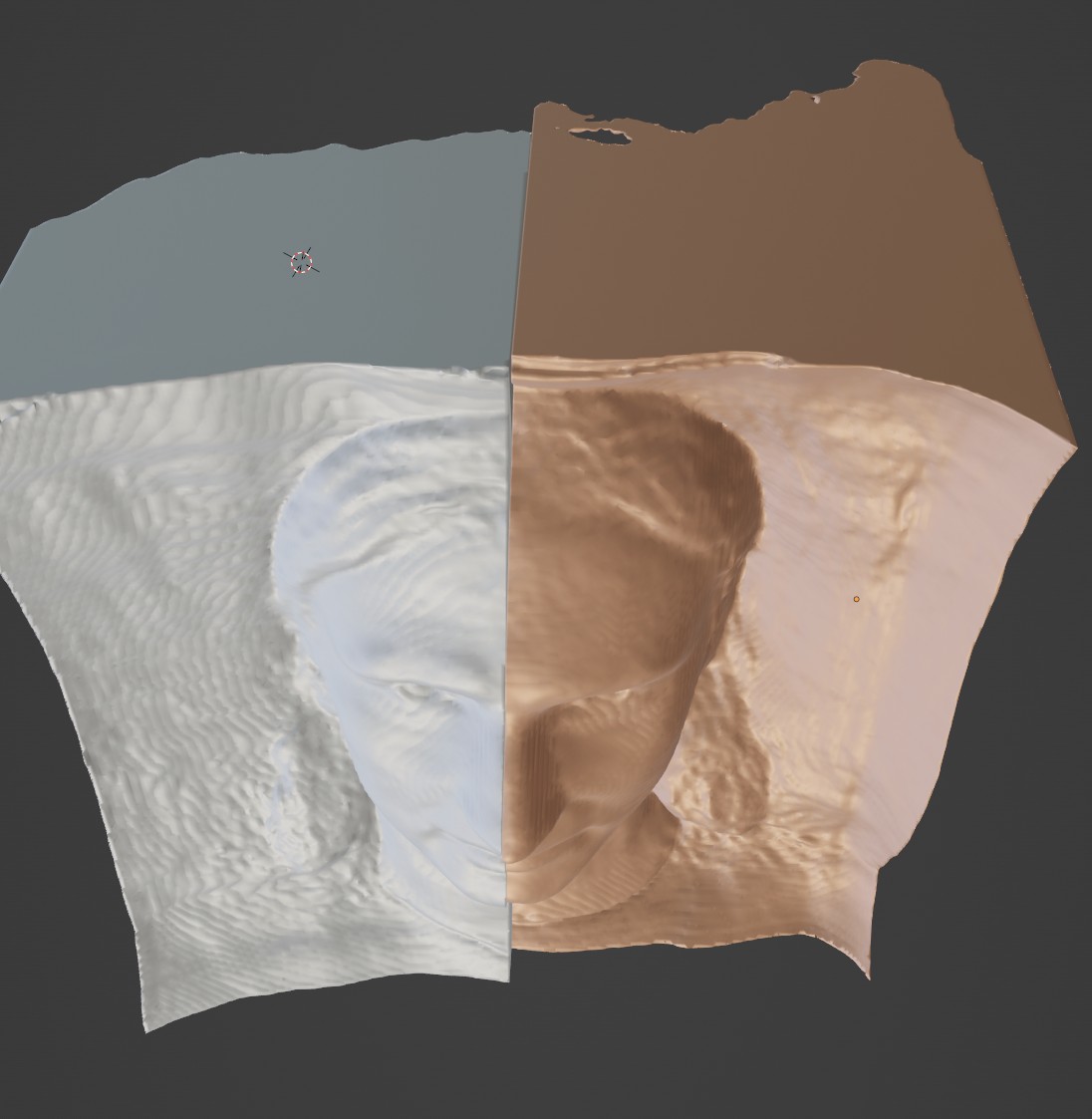} \\
    \end{tabular}
    }
    \captionof{figure}{Side-by-side comparison of \textit{Ours} (golden, right half) and \textit{Ours w/o symm. prior} (gray, left half). We observe that the methods produce similar geometry in the facial region, while the background ``cutting plane'' of the mesh is closer to the face w/o symm. prior. This way, the symm. prior allows to extend the modeled region. \textit{Zoom-in recommended.}}
    \label{fig:ours_wo_no_symm_prior_overlay}
\end{table}

\section{Additional Qualitative Results}
In this section, we present additional qualitative results on same-view inversion, novel view rendering and ablation for the loss, and architecture change.

\begin{itemize}
    \item Figs. \ref{supp:visualcomp1}, \ref{supp:visualcomp2}, \ref{refinement-hard:fig} and \ref{refinement-easy:fig} provide qualitative comparison of our approach with existing state-of-the-art inversion techniques on image reconstruction.
    
    \item Figs. \ref{supp:visualcomp3}, \ref{supp:visualcomp4}, \ref{supp:visualcomp5}, \ref{supp:visualcomp6}, and \ref{supp:visualcomp7} demonstrate qualitative comparison of our approach with existing state-of-the-art inversion techniques on novel-view rendering.
    \item Figs. \ref{fig:ablation-supp1} and \ref{fig:ablation-supp2} reflect extensive qualitative ablation study for the loss, and architecture changes.
\end{itemize}

\begin{table*}[h!]
\setlength{\tabcolsep}{0cm}
\renewcommand{\arraystretch}{0}
\begin{tabular}{c@{\hskip 0.2cm}cccccccc}
    \raisebox{2\height}{\rotatebox[origin=c]{90}{Input}}
    & 
    \includegraphics[width=0.12\textwidth]{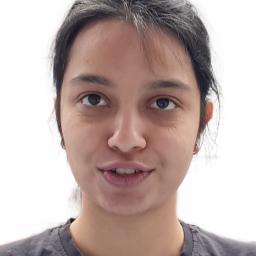} & 
    \includegraphics[width=0.12\textwidth]{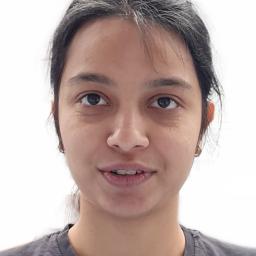} &
    \includegraphics[width=0.12\textwidth]{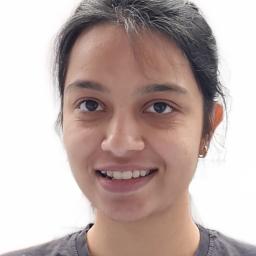} &
    \includegraphics[width=0.12\textwidth]{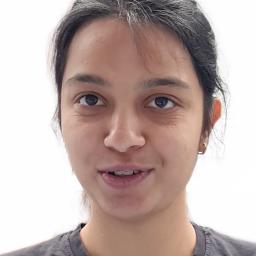} &
    \includegraphics[width=0.12\textwidth]{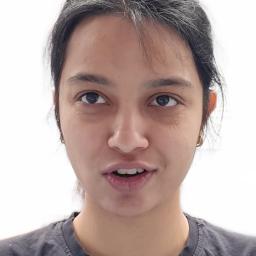} & 
    \includegraphics[width=0.12\textwidth]{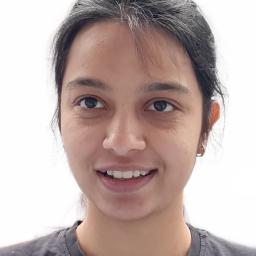} &
    \includegraphics[width=0.12\textwidth]{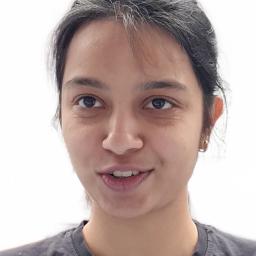} &
    \includegraphics[width=0.12\textwidth]{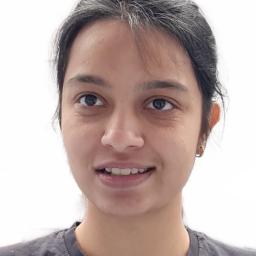}
    \\
    \raisebox{1.0\height}{\rotatebox[origin=c]{90}{Novel view}}  & 
    \includegraphics[width=0.12\textwidth]{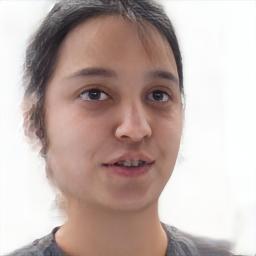} & 
    \includegraphics[width=0.12\textwidth]{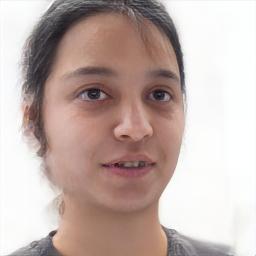} &
    \includegraphics[width=0.12\textwidth]{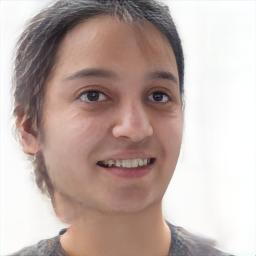} &
    \includegraphics[width=0.12\textwidth]{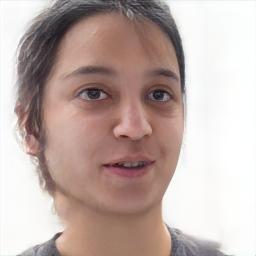} &
    \includegraphics[width=0.12\textwidth]{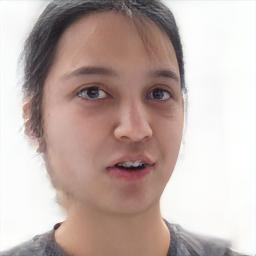} & 
    \includegraphics[width=0.12\textwidth]{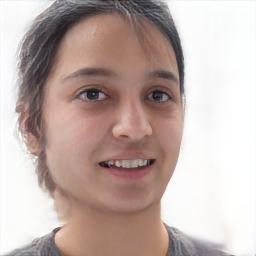} &
    \includegraphics[width=0.12\textwidth]{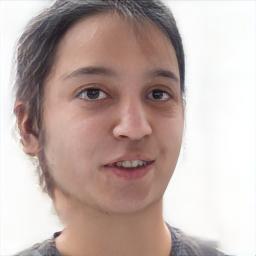} &
    \includegraphics[width=0.12\textwidth]{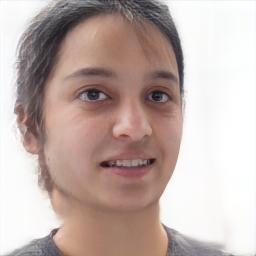}
    
    \\
\end{tabular}
\vspace{-0.25cm}
\captionof{figure}{Novel view synthesis of frames extracted from a talking head video. The novel view yaw angles relative to the frontal view are as follows: second row (-0.3 radians), fourth row (0.0 radians), sixth row (-0.3 radians), eighth row (0.6 radians), and last row (0.8 radians). \textit{Electronic zoom-in recommended.}}
\label{video:novel-view-supp}
\vspace{-0.4cm}
\end{table*}

\begin{table*}[h!]
    \centering
    \setlength{\tabcolsep}{0pt}
    \renewcommand{\arraystretch}{0}
    \resizebox{\textwidth}{!}{%
        \begin{tabular}{c@{\hskip 0.3cm}c@{\hskip 0.1cm}c@{\hskip 0.3cm}c@{\hskip 0.1cm}ccc}
            \raisebox{1.0\height} {Input}
            &
            &
            \raisebox{1.0\height}{Inversion}
            & 
            &
            \raisebox{1.0\height}{Edit}
            &
            \raisebox{1.0\height}{Novel View}
            &
            \raisebox{1.0\height}{Novel View}
            \\
            \multirow{2}{*}[4em]{
                \includegraphics[width=0.15\textwidth]{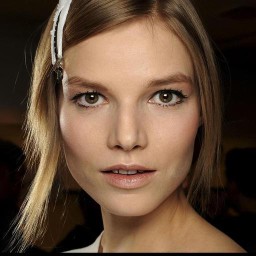}
            }
            & 
            \raisebox{1.5\height}{\rotatebox[origin=c]{90}{SG W \cite{karras2020analyzing}}}
            & 
            \includegraphics[width=0.15\textwidth]
            {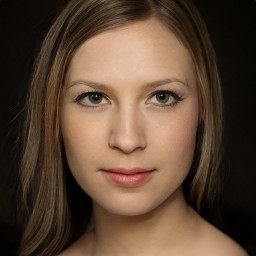}
            &
            \raisebox{1.5\height}{\rotatebox[origin=c]{90}{+Smile}}
            &
            \includegraphics[width=0.15\textwidth]
            {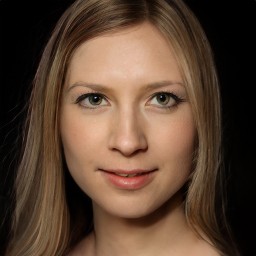}
            &
            \includegraphics[width=0.15\textwidth]
            {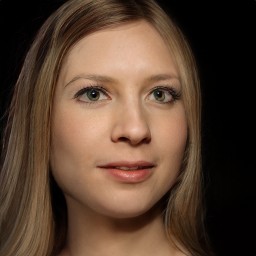}
            &
            \includegraphics[width=0.15\textwidth]
            {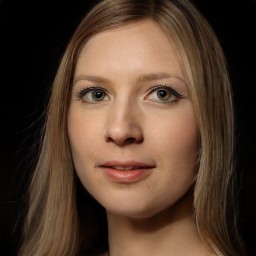}
            \\
            &
            \raisebox{2.5\height}{\rotatebox[origin=c]{90}{\textbf{Ours}}}
            &
            \includegraphics[width=0.15\textwidth]
            {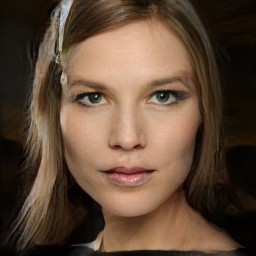}
            &
            \raisebox{1.5\height}{\rotatebox[origin=c]{90}{+Smile}}
            &
            \includegraphics[width=0.15\textwidth]
            {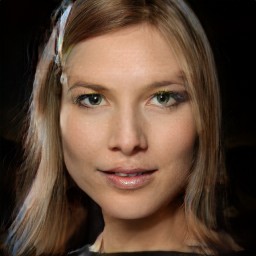}
            &
            \includegraphics[width=0.15\textwidth]
            {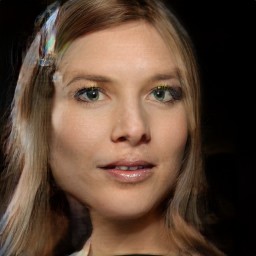}
            &
            \includegraphics[width=0.15\textwidth]
            {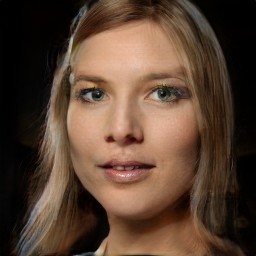}
            \\
\end{tabular}
    }
    \captionof{figure}{Editing result for smile attribute. By utilizing tri-plane features, our method preserves identity more and also generates both realistic and view-consistent editing.}
    \label{supp:editing1}
\end{table*}

\begin{table*}[h!]
    \centering
    \setlength{\tabcolsep}{0pt}
    \renewcommand{\arraystretch}{0}
    \resizebox{\textwidth}{!}{%
        \begin{tabular}{c@{\hskip 0.3cm}c@{\hskip 0.1cm}c@{\hskip 0.3cm}c@{\hskip 0.1cm}ccc}
            \raisebox{1.0\height} {Input}
            &
            &
            \raisebox{1.0\height}{Inversion}
            & 
            &
            \raisebox{1.0\height}{Edit}
            &
            \raisebox{1.0\height}{Novel View}
            &
            \raisebox{1.0\height}{Novel View}
            \\
            \multirow{4}{*}[-5em]{
                \includegraphics[width=0.15\textwidth]{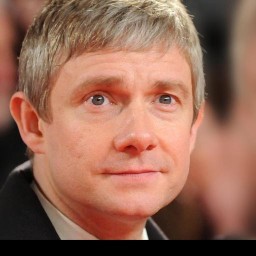}
            }
            & 
            \multirow{2}{*}[2.0em]{
                \raisebox{0.1\height}{\rotatebox[origin=c]{90}{SG W \cite{karras2020analyzing}}}
            }
            & 
            \multirow{2}{*}[3em]{
                \includegraphics[width=0.15\textwidth]
            {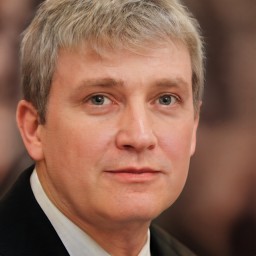}
            }
            &
            \raisebox{2.5\height}{\rotatebox[origin=c]{90}{+Age}}
            &
            \includegraphics[width=0.15\textwidth]
            {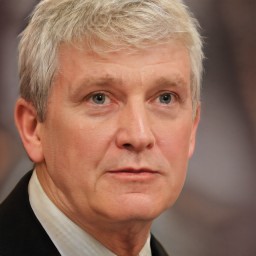}
            &
            \includegraphics[width=0.15\textwidth]
            {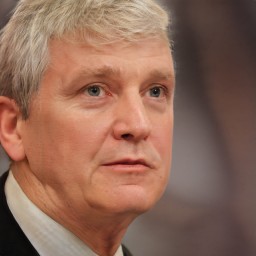}
            &
            \includegraphics[width=0.15\textwidth]
            {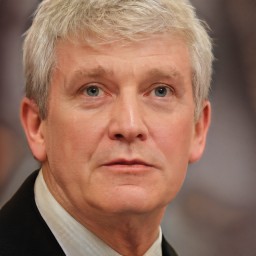}
            \\
            &
            &
            &
            \raisebox{2.5\height}{\rotatebox[origin=c]{90}{-Age}}
            &
            \includegraphics[width=0.15\textwidth]
            {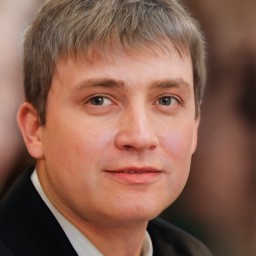}
            &
            \includegraphics[width=0.15\textwidth]
            {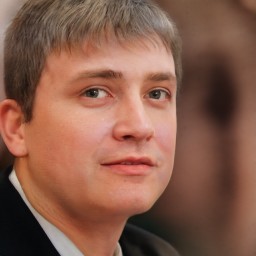}
            &
            \includegraphics[width=0.15\textwidth]
            {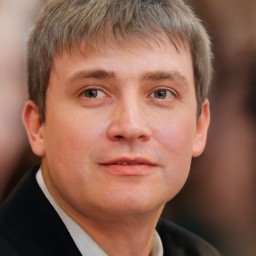}
            \\
\end{tabular}
    }
    \captionof{figure}{Editing result for age attribute. By utilizing tri-plane features, our method preserves identity more and also generates both realistic and view-consistent editing.}
    \label{supp:editing2}
\end{table*}

\begin{table*}[h]
\centering
\setlength{\tabcolsep}{0pt}
\renewcommand{\arraystretch}{0}
\begin{tabular}{c@{\hskip 0.3cm}ccccc}
    \raisebox{2.5\height}{\rotatebox[origin=c]{90}{Input}}
    &
    \includegraphics[width=0.18\textwidth]{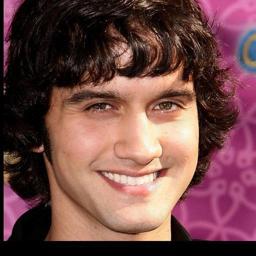}
    &
    \includegraphics[width=0.18\textwidth]{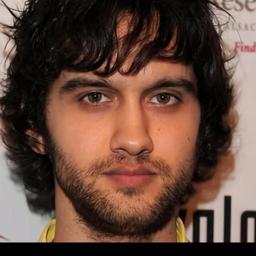}
    &
    \includegraphics[width=0.18\textwidth]{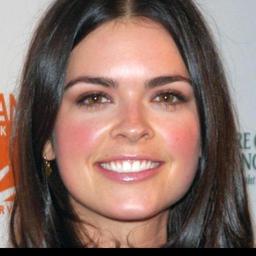}
    &
    \includegraphics[width=0.18\textwidth]{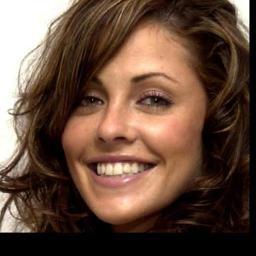}
    &
    \includegraphics[width=0.18\textwidth]{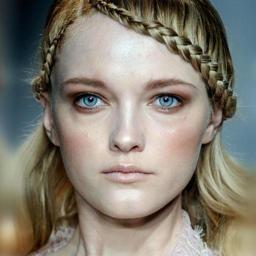}
    \\[0.3cm]
    \raisebox{1.2\height}{\rotatebox[origin=c]{90}{$\mathcal{W}+$ opt. \cite{karras2020analyzing}}}
    &
    \includegraphics[width=0.18\textwidth]{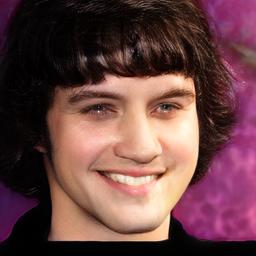}
    &
    \includegraphics[width=0.18\textwidth]{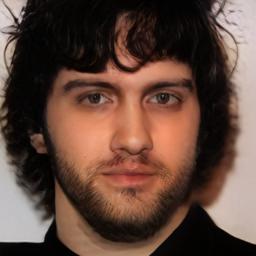}
    &
    \includegraphics[width=0.18\textwidth]{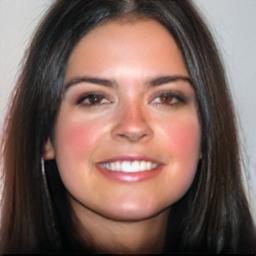}
    &
    \includegraphics[width=0.18\textwidth]{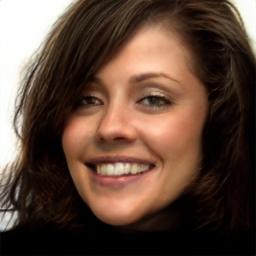}
    &
    \includegraphics[width=0.18\textwidth]{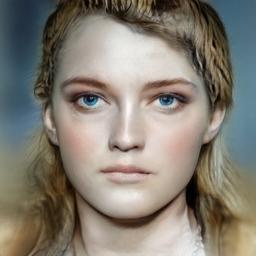}
    \\[0cm]
    \raisebox{2.0\height}{\rotatebox[origin=c]{90}{PTI~\cite{roich2022pivotal}}}
    &
    \includegraphics[width=0.18\textwidth]{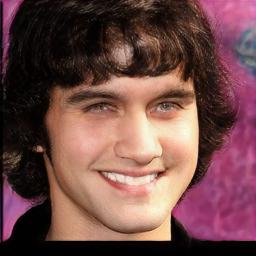}
    &
    \includegraphics[width=0.18\textwidth]{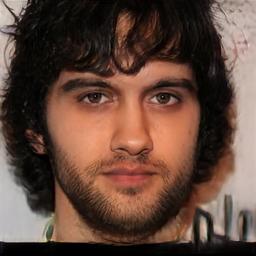}
    &
    \includegraphics[width=0.18\textwidth]{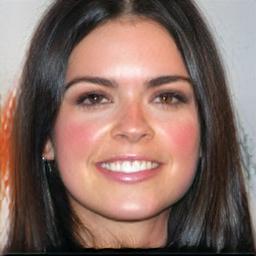}
    &
    \includegraphics[width=0.18\textwidth]{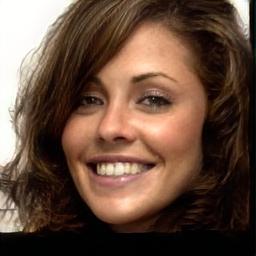}
    &
    \includegraphics[width=0.18\textwidth]{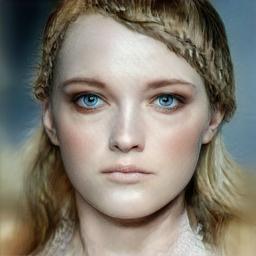}
    \\
\end{tabular}
\captionof{figure}{Additional qualitative comparison on image reconstruction.}
\label{supp:visualcomp1}
\vspace{-0.3cm}
\end{table*}

\begin{table*}[]
\centering
\setlength{\tabcolsep}{0pt}
\renewcommand{\arraystretch}{0}
\begin{tabular}{c@{\hskip 0.3cm}ccccc}
    \raisebox{3.0\height}{\rotatebox[origin=c]{90}{Input}}
    &
    \includegraphics[width=0.18\textwidth]{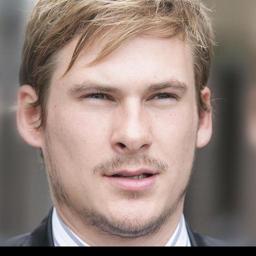}
    & 
    \includegraphics[width=0.18\textwidth]{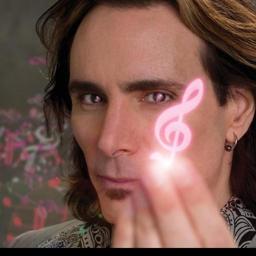}
    &
    \includegraphics[width=0.18\textwidth]{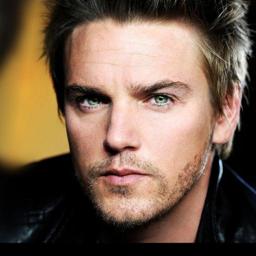}
    &
    \includegraphics[width=0.18\textwidth]{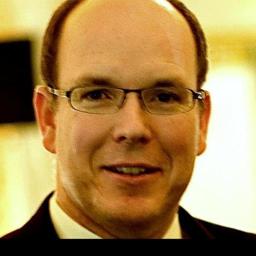}
    &
    \includegraphics[width=0.18\textwidth]{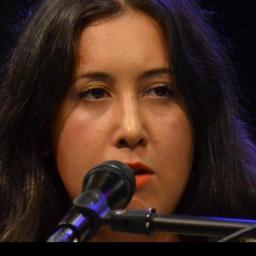}
    \\[0.3cm]
    \raisebox{1.2\height}{\rotatebox[origin=c]{90}{$\mathcal{W}+$ opt. \cite{karras2020analyzing}}}
    &
    \includegraphics[width=0.18\textwidth]{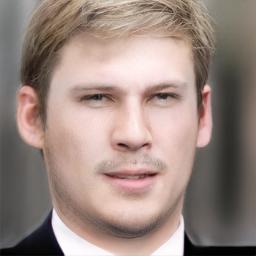}
    & 
    \includegraphics[width=0.18\textwidth]{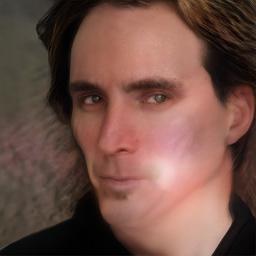}
    &
    \includegraphics[width=0.18\textwidth]{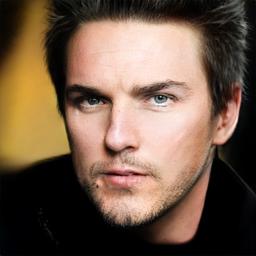}
    &
    \includegraphics[width=0.18\textwidth]{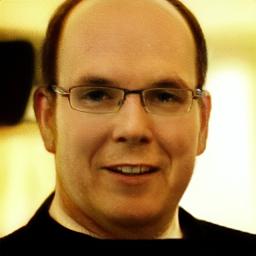}
    &
    \includegraphics[width=0.18\textwidth]{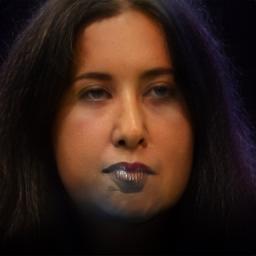}
    \\
    \raisebox{2.0\height}{\rotatebox[origin=c]{90}{PTI~\cite{roich2022pivotal}}}
    &
    \includegraphics[width=0.18\textwidth]{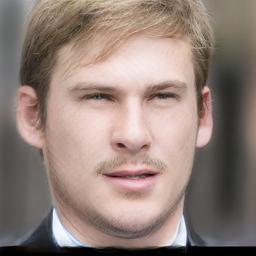}
    & 
    \includegraphics[width=0.18\textwidth]{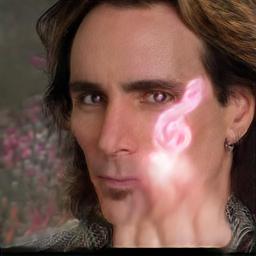}
    &
    \includegraphics[width=0.18\textwidth]{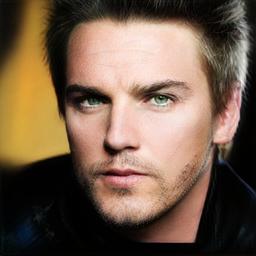}
    &
    \includegraphics[width=0.18\textwidth]{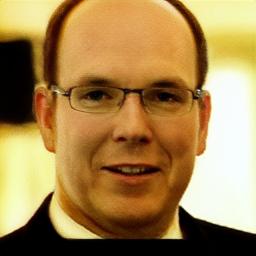}
    &
    \includegraphics[width=0.18\textwidth]{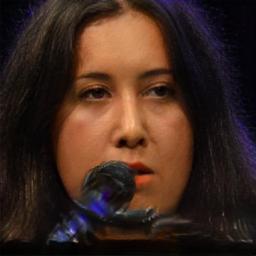}
    \\
\end{tabular}
\captionof{figure}{Additional qualitative comparison on image reconstruction.}
\label{supp:visualcomp2}
\vspace{-0.3cm}
\end{table*}

\begin{table*}[]
\centering
\setlength{\tabcolsep}{0pt}
\renewcommand{\arraystretch}{0}
\begin{tabular}{c@{\hskip 0.3cm}ccccc}
    \raisebox{2.5\height}{\rotatebox[origin=c]{90}{Input}}
    &
    \includegraphics[width=0.18\textwidth]{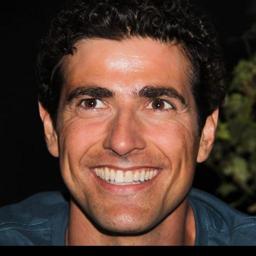}
    &
    \includegraphics[width=0.18\textwidth]{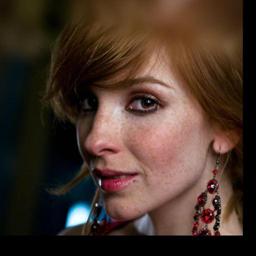}
    &
    \includegraphics[width=0.18\textwidth]{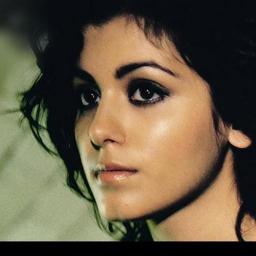}
    &
    \includegraphics[width=0.18\textwidth]{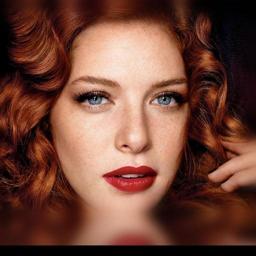}
    &
    \includegraphics[width=0.18\textwidth]{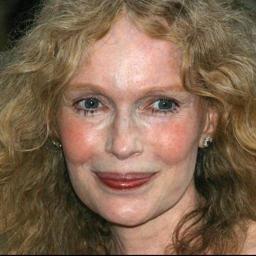}
    \\[0.3cm]
    \raisebox{1.2\height}{\rotatebox[origin=c]{90}{$\mathcal{W}+$ opt. \cite{karras2020analyzing}}}
    &
    \includegraphics[width=0.18\textwidth]{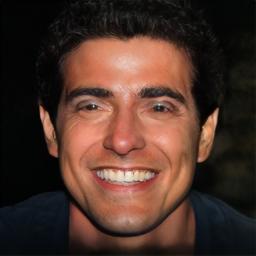}
    &
    \includegraphics[width=0.18\textwidth]{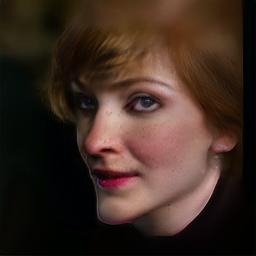}
    &
    \includegraphics[width=0.18\textwidth]{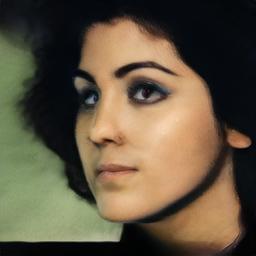}
    &
    \includegraphics[width=0.18\textwidth]{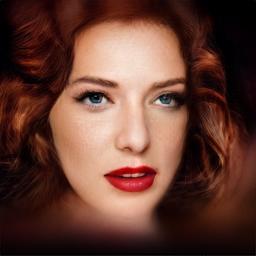}
    &
    \includegraphics[width=0.18\textwidth]{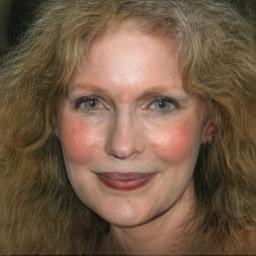}
    \\
    \raisebox{1.7\height}{\rotatebox[origin=c]{90}{PTI~\cite{roich2022pivotal}}}
    &
    \includegraphics[width=0.18\textwidth]{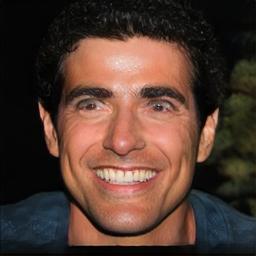}
    &
    \includegraphics[width=0.18\textwidth]{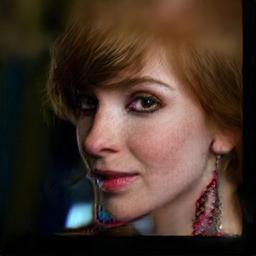}
    &
    \includegraphics[width=0.18\textwidth]{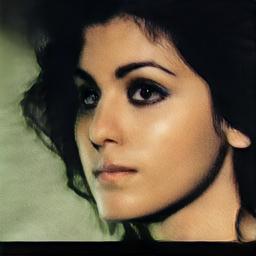}
    &
    \includegraphics[width=0.18\textwidth]{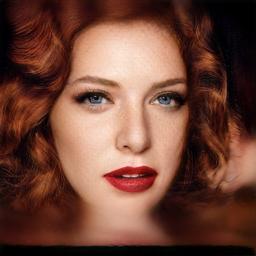}
    &
    \includegraphics[width=0.18\textwidth]{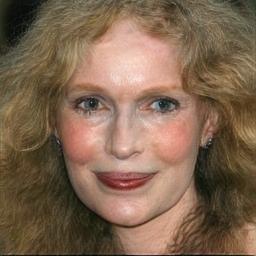}
    \\
\end{tabular}
\captionof{figure}{Additional qualitative comparison on image reconstruction.}
\label{refinement-hard:fig}
\vspace{-0.3cm}
\end{table*}

\begin{table*}[]
\centering
\setlength{\tabcolsep}{0pt}
\renewcommand{\arraystretch}{0}
\begin{tabular}{c@{\hskip 0.3cm}ccccc}
    \raisebox{2.5\height}{\rotatebox[origin=c]{90}{Input}}
    &
    \includegraphics[width=0.18\textwidth]{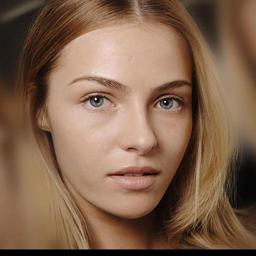}
    &
    \includegraphics[width=0.18\textwidth]{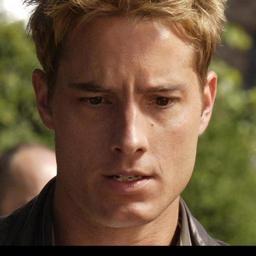}
    & 
    \includegraphics[width=0.18\textwidth]{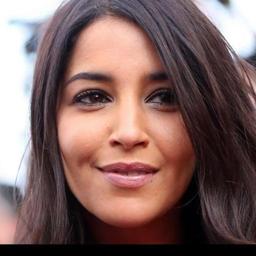}
    &
    \includegraphics[width=0.18\textwidth]{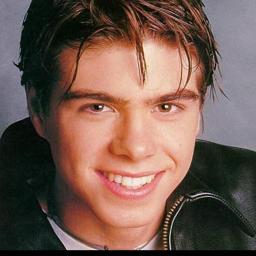}
    &
    \includegraphics[width=0.18\textwidth]{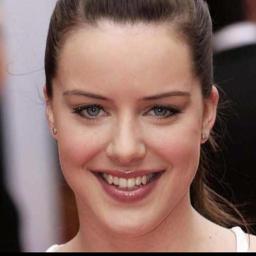}
    \\[0.3cm]
    \raisebox{1.2\height}{\rotatebox[origin=c]{90}{$\mathcal{W}+$ opt. \cite{karras2020analyzing}}}
    &
    \includegraphics[width=0.18\textwidth]{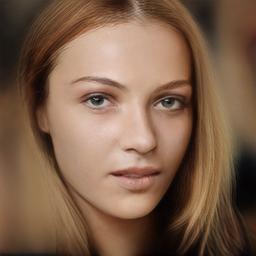}
    &
    \includegraphics[width=0.18\textwidth]{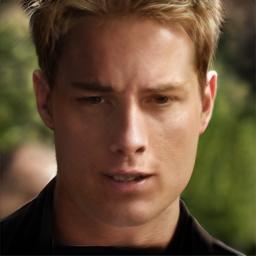}
    & 
    \includegraphics[width=0.18\textwidth]{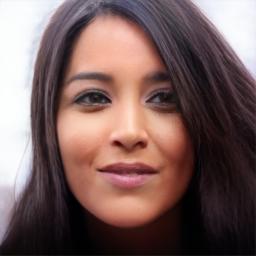}
    &
    \includegraphics[width=0.18\textwidth]{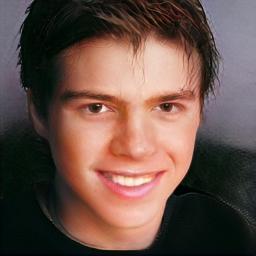}
    &
    \includegraphics[width=0.18\textwidth]{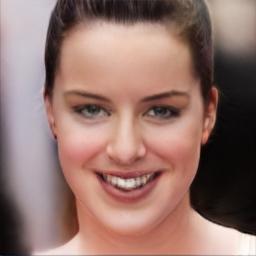}
    \\
    \raisebox{1.7\height}{\rotatebox[origin=c]{90}{PTI~\cite{roich2022pivotal}}}
    &
    \includegraphics[width=0.18\textwidth]{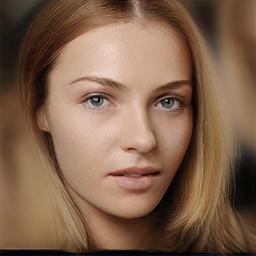}
    &
    \includegraphics[width=0.18\textwidth]{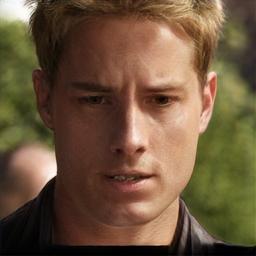}
    & 
    \includegraphics[width=0.18\textwidth]{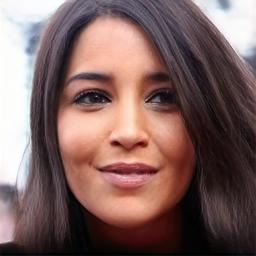}
    &
    \includegraphics[width=0.18\textwidth]{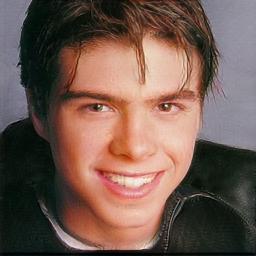}
    &
    \includegraphics[width=0.18\textwidth]{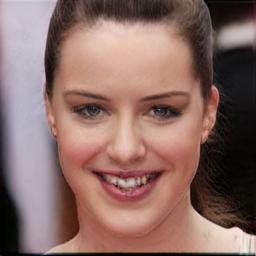}
    \\
\end{tabular}
\captionof{figure}{Additional qualitative comparison on image reconstruction.}
\label{refinement-easy:fig}
\vspace{-0.3cm}
\end{table*}

\begin{table*}[]
\centering
\setlength{\tabcolsep}{0pt}
\renewcommand{\arraystretch}{0}
\begin{tabular}{cc@{\hskip 0.3cm}ccccc}
    \raisebox{2.5\height}{\rotatebox[origin=c]{90}{Input}}
    &
    \includegraphics[width=0.18\textwidth]{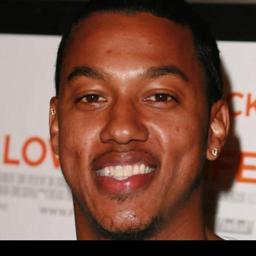}
    &
    \raisebox{1.2\height}{\rotatebox[origin=c]{90}{$\mathcal{W}+$ opt. \cite{karras2020analyzing}}}
    &
    \includegraphics[width=0.18\textwidth]{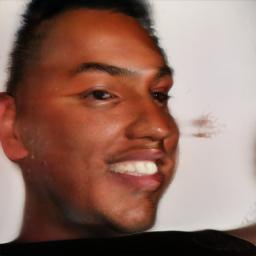}
    & 
    \includegraphics[width=0.18\textwidth]{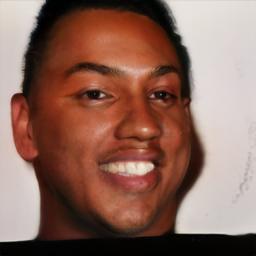}
    & 
    \includegraphics[width=0.18\textwidth]{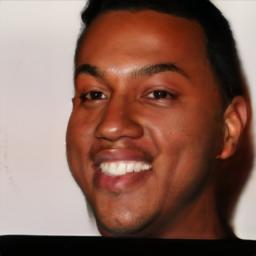}
    & 
    \includegraphics[width=0.18\textwidth]{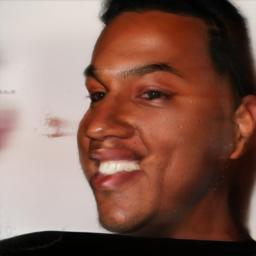}
    \\
    \raisebox{2.5\height}{\rotatebox[origin=c]{90}{Input}}
    &
    \includegraphics[width=0.18\textwidth]{images/supplement/rotation/original/28451.jpg}
    &
    \raisebox{2.0\height}{\rotatebox[origin=c]{90}{PTI~\cite{roich2022pivotal}}}
    &
    \includegraphics[width=0.18\textwidth]{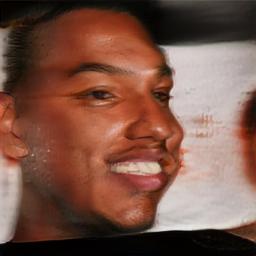}
    & 
    \includegraphics[width=0.18\textwidth]{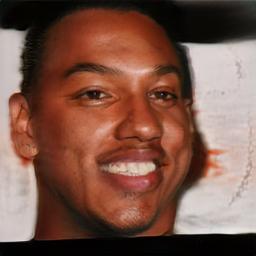}
    & 
    \includegraphics[width=0.18\textwidth]{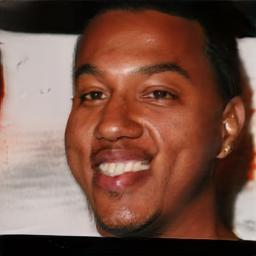}
    & 
    \includegraphics[width=0.18\textwidth]{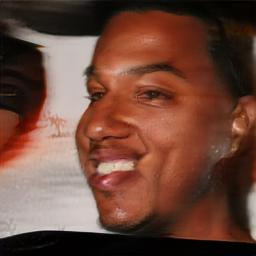}
    \\
    \raisebox{2.5\height}{\rotatebox[origin=c]{90}{Input}}
    &
    \includegraphics[width=0.18\textwidth]{images/supplement/rotation/original/28451.jpg}
    &
    \raisebox{1.5\height}{\rotatebox[origin=c]{90}{Pose Opt.~\cite{ko20233d}}}
    &
    \includegraphics[width=0.18\textwidth]{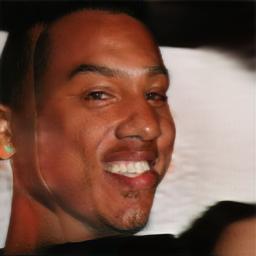}
    & 
    \includegraphics[width=0.18\textwidth]{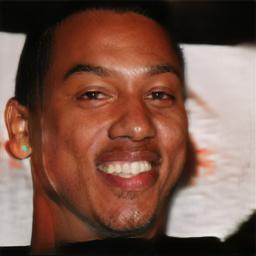}
    & 
    \includegraphics[width=0.18\textwidth]{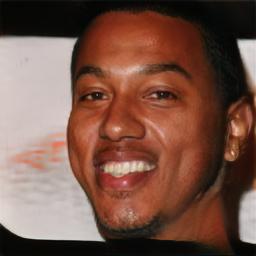}
    & 
    \includegraphics[width=0.18\textwidth]{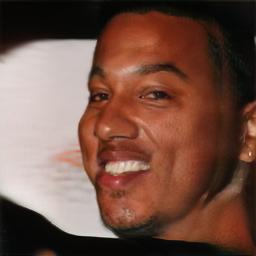}
    \\
\end{tabular}
\captionof{figure}{Additional qualitative evaluation on novel view rendering of yaw angle -0.6, -0.3, 0.3, and 0.6 radians.}
\label{supp:visualcomp4}
\vspace{-0.3cm}
\end{table*}

\begin{table*}[]
\centering
\setlength{\tabcolsep}{0pt}
\renewcommand{\arraystretch}{0}
\begin{tabular}{cc@{\hskip 0.3cm}ccccc}
    \raisebox{2.5\height}{\rotatebox[origin=c]{90}{Input}}
    &
    \includegraphics[width=0.18\textwidth]{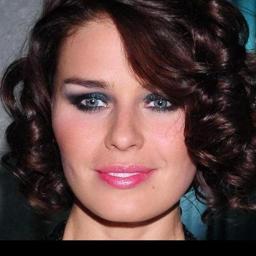}
    &
    \raisebox{1.2\height}{\rotatebox[origin=c]{90}{$\mathcal{W}+$ opt. \cite{karras2020analyzing}}}
    &
    \includegraphics[width=0.18\textwidth]{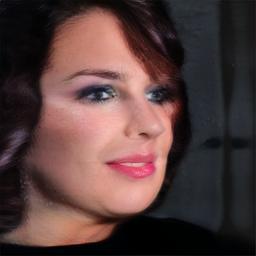}
    & 
    \includegraphics[width=0.18\textwidth]{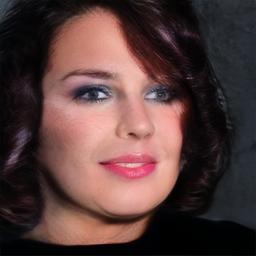}
    & 
    \includegraphics[width=0.18\textwidth]{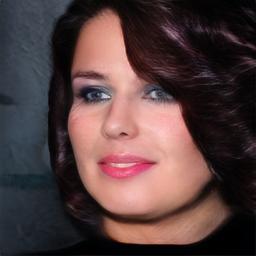}
    & 
    \includegraphics[width=0.18\textwidth]{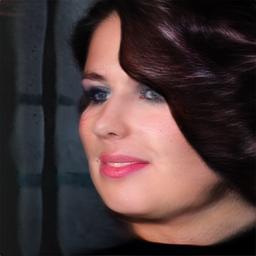}
    \\
    \raisebox{2.5\height}{\rotatebox[origin=c]{90}{Input}}
    &
    \includegraphics[width=0.18\textwidth]{images/supplement/rotation/original/16531.jpg}
    &
    \raisebox{2.0\height}{\rotatebox[origin=c]{90}{PTI~\cite{roich2022pivotal}}}
    &
    \includegraphics[width=0.18\textwidth]{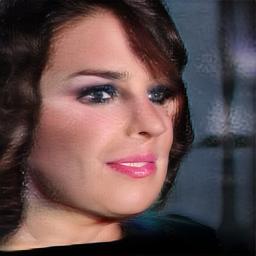}
    & 
    \includegraphics[width=0.18\textwidth]{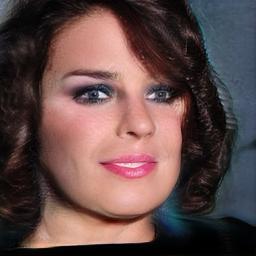}
    & 
    \includegraphics[width=0.18\textwidth]{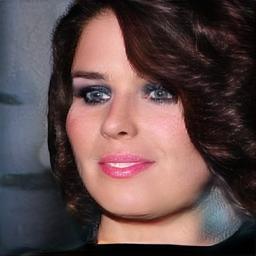}
    & 
    \includegraphics[width=0.18\textwidth]{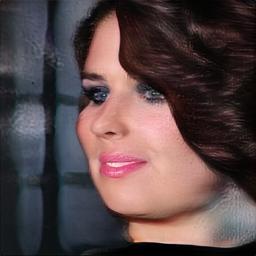}
    \\
    \raisebox{2.5\height}{\rotatebox[origin=c]{90}{Input}}
    &
    \includegraphics[width=0.18\textwidth]{images/supplement/rotation/original/16531.jpg}
    &
    \raisebox{1.5\height}{\rotatebox[origin=c]{90}{Pose Opt.~\cite{ko20233d}}}
    &
    \includegraphics[width=0.18\textwidth]{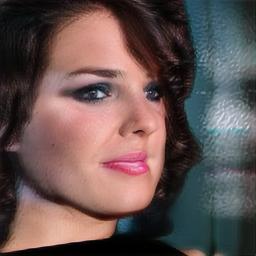}
    & 
    \includegraphics[width=0.18\textwidth]{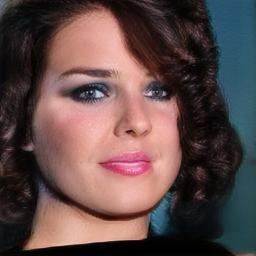}
    & 
    \includegraphics[width=0.18\textwidth]{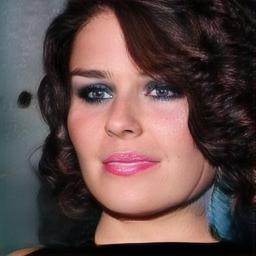}
    & 
    \includegraphics[width=0.18\textwidth]{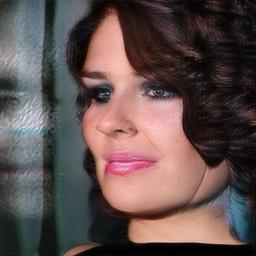}
    \\
\end{tabular}
\captionof{figure}{Additional qualitative evaluation on novel view rendering of yaw angle -0.6, -0.3, 0.3, and 0.6 radians.}
\label{supp:visualcomp5}
\vspace{-0.3cm}
\end{table*}

\begin{table*}[]
\centering
\setlength{\tabcolsep}{0pt}
\renewcommand{\arraystretch}{0}
\begin{tabular}{cc@{\hskip 0.3cm}ccccc}
    \raisebox{2.5\height}{\rotatebox[origin=c]{90}{Input}}
    &
    \includegraphics[width=0.18\textwidth]{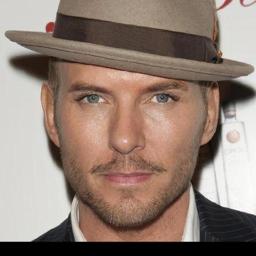}
    &
    \raisebox{1.2\height}{\rotatebox[origin=c]{90}{$\mathcal{W}+$ opt. \cite{karras2020analyzing}}}
    &
    \includegraphics[width=0.18\textwidth]{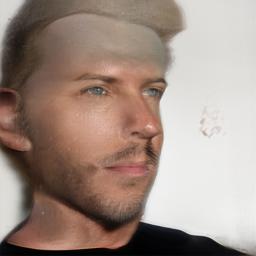}
    & 
    \includegraphics[width=0.18\textwidth]{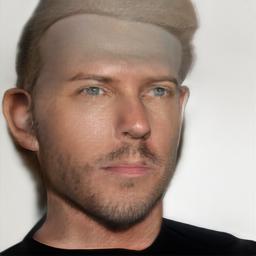}
    & 
    \includegraphics[width=0.18\textwidth]{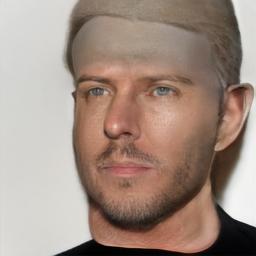}
    & 
    \includegraphics[width=0.18\textwidth]{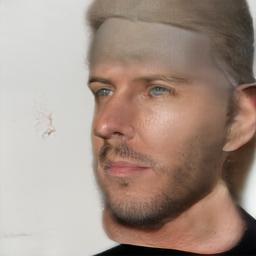}
    \\
    \raisebox{2.5\height}{\rotatebox[origin=c]{90}{Input}}
    &
    \includegraphics[width=0.18\textwidth]{images/supplement/rotation/original/12253.jpg}
    &
    \raisebox{2.0\height}{\rotatebox[origin=c]{90}{PTI~\cite{roich2022pivotal}}}
    &
    \includegraphics[width=0.18\textwidth]{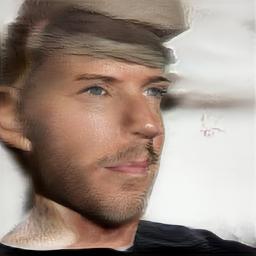}
    & 
    \includegraphics[width=0.18\textwidth]{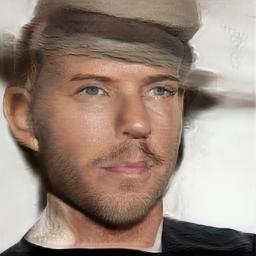}
    & 
    \includegraphics[width=0.18\textwidth]{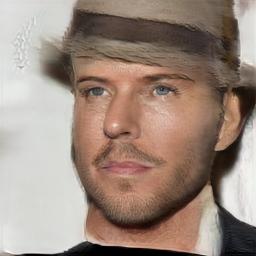}
    & 
    \includegraphics[width=0.18\textwidth]{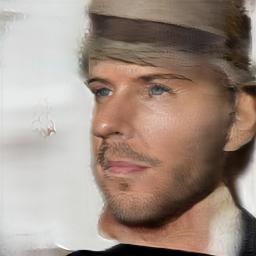}
    \\
    \raisebox{2.5\height}{\rotatebox[origin=c]{90}{Input}}
    &
    \includegraphics[width=0.18\textwidth]{images/supplement/rotation/original/12253.jpg}
    &
    \raisebox{1.5\height}{\rotatebox[origin=c]{90}{Pose Opt.~\cite{ko20233d}}}
    &
    \includegraphics[width=0.18\textwidth]{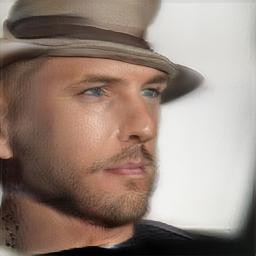}
    & 
    \includegraphics[width=0.18\textwidth]{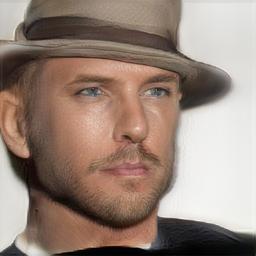}
    & 
    \includegraphics[width=0.18\textwidth]{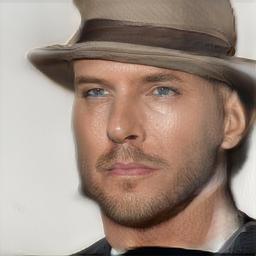}
    & 
    \includegraphics[width=0.18\textwidth]{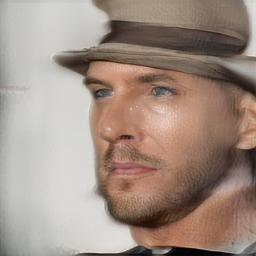}
    \\
\end{tabular}
\captionof{figure}{Additional qualitative evaluation on novel view rendering of yaw angle -0.6, -0.3, 0.3, and 0.6 radians.}
\label{supp:visualcomp3}
\vspace{-0.3cm}
\end{table*}

\begin{table*}[]
\centering
\setlength{\tabcolsep}{0pt}
\renewcommand{\arraystretch}{0}
\begin{tabular}{cc@{\hskip 0.3cm}ccccc}
    \raisebox{2.5\height}{\rotatebox[origin=c]{90}{Input}}
    &
    \includegraphics[width=0.18\textwidth]{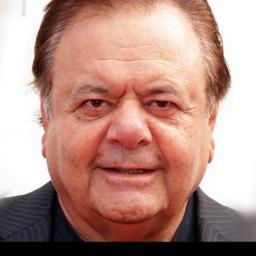}
    &
    \raisebox{1.2\height}{\rotatebox[origin=c]{90}{$\mathcal{W}+$ opt. \cite{karras2020analyzing}}}
    &
    \includegraphics[width=0.18\textwidth]{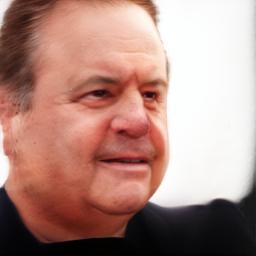}
    & 
    \includegraphics[width=0.18\textwidth]{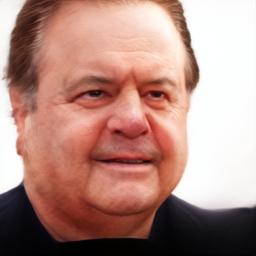}
    & 
    \includegraphics[width=0.18\textwidth]{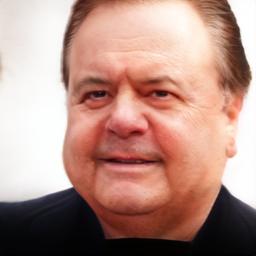}
    & 
    \includegraphics[width=0.18\textwidth]{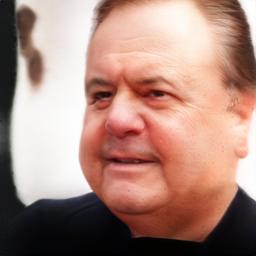}
    \\
    \raisebox{2.5\height}{\rotatebox[origin=c]{90}{Input}}
    &
    \includegraphics[width=0.18\textwidth]{images/supplement/rotation/original/18703.jpg}
    &
    \raisebox{2.0\height}{\rotatebox[origin=c]{90}{PTI~\cite{roich2022pivotal}}}
    &
    \includegraphics[width=0.18\textwidth]{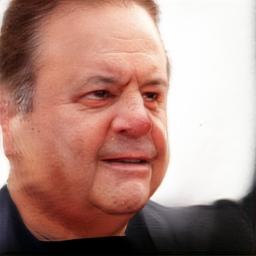}
    & 
    \includegraphics[width=0.18\textwidth]{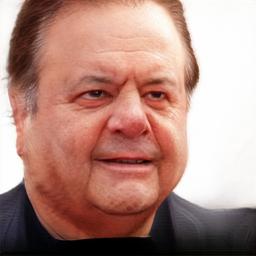}
    & 
    \includegraphics[width=0.18\textwidth]{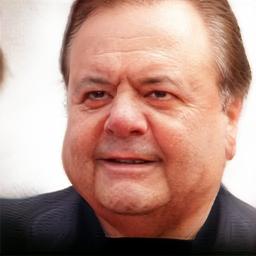}
    & 
    \includegraphics[width=0.18\textwidth]{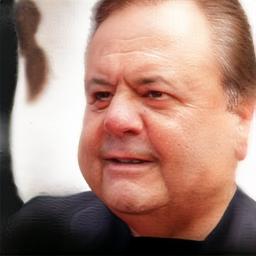}
    \\
    \raisebox{2.5\height}{\rotatebox[origin=c]{90}{Input}}
    &
    \includegraphics[width=0.18\textwidth]{images/supplement/rotation/original/18703.jpg}
    &
    \raisebox{1.5\height}{\rotatebox[origin=c]{90}{Pose Opt.~\cite{ko20233d}}}
    &
    \includegraphics[width=0.18\textwidth]{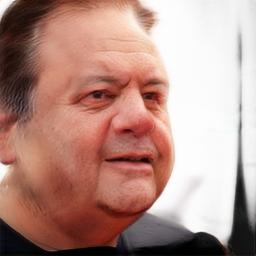}
    & 
    \includegraphics[width=0.18\textwidth]{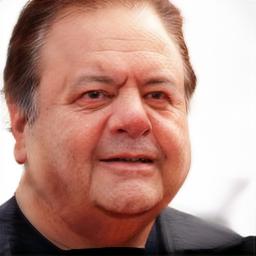}
    & 
    \includegraphics[width=0.18\textwidth]{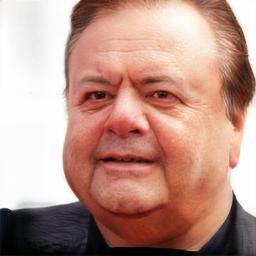}
    & 
    \includegraphics[width=0.18\textwidth]{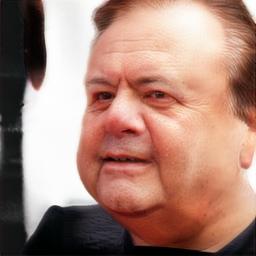}
    \\
\end{tabular}
\captionof{figure}{Additional qualitative evaluation on novel view rendering of yaw angle -0.6, -0.3, 0.3, and 0.6 radians.}
\label{supp:visualcomp6}
\vspace{-0.3cm}
\end{table*}

\begin{table*}[]
\centering
\setlength{\tabcolsep}{0pt}
\renewcommand{\arraystretch}{0}
\begin{tabular}{cc@{\hskip 0.3cm}ccccc}
    \raisebox{2.5\height}{\rotatebox[origin=c]{90}{Input}}
    &
    \includegraphics[width=0.18\textwidth]{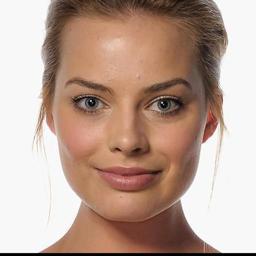}
    &
    \raisebox{1.2\height}{\rotatebox[origin=c]{90}{$\mathcal{W}+$ opt. \cite{karras2020analyzing}}}
    &
    \includegraphics[width=0.18\textwidth]{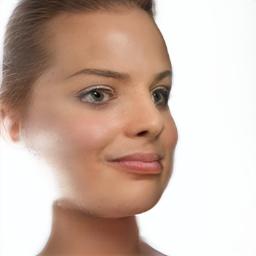}
    & 
    \includegraphics[width=0.18\textwidth]{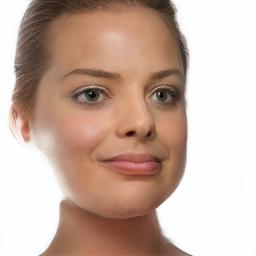}
    & 
    \includegraphics[width=0.18\textwidth]{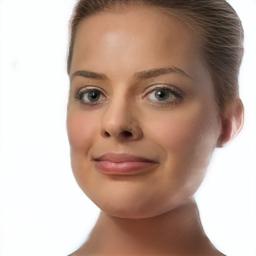}
    & 
    \includegraphics[width=0.18\textwidth]{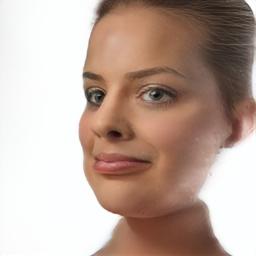}
    \\
    \raisebox{2.5\height}{\rotatebox[origin=c]{90}{Input}}
    &
    \includegraphics[width=0.18\textwidth]{images/supplement/rotation/original/26015.jpg}
    &
    \raisebox{2.0\height}{\rotatebox[origin=c]{90}{PTI~\cite{roich2022pivotal}}}
    &
    \includegraphics[width=0.18\textwidth]{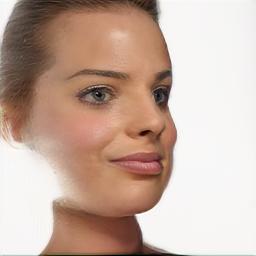}
    & 
    \includegraphics[width=0.18\textwidth]{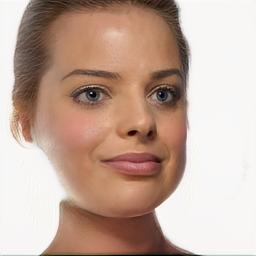}
    & 
    \includegraphics[width=0.18\textwidth]{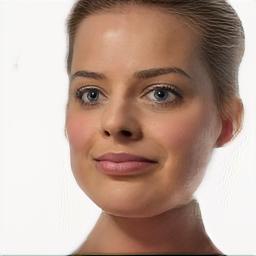}
    & 
    \includegraphics[width=0.18\textwidth]{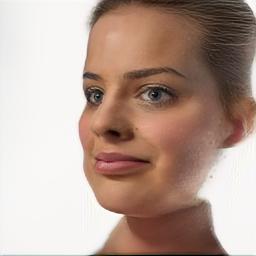}
    \\
    \raisebox{2.5\height}{\rotatebox[origin=c]{90}{Input}}
    &
    \includegraphics[width=0.18\textwidth]{images/supplement/rotation/original/26015.jpg}
    &
    \raisebox{1.5\height}{\rotatebox[origin=c]{90}{Pose Opt.~\cite{ko20233d}}}
    &
    \includegraphics[width=0.18\textwidth]{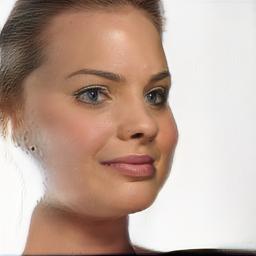}
    & 
    \includegraphics[width=0.18\textwidth]{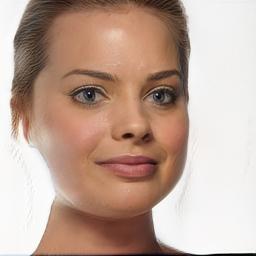}
    & 
    \includegraphics[width=0.18\textwidth]{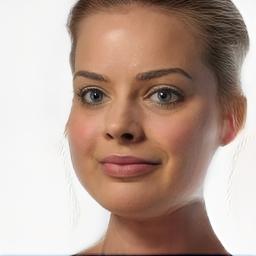}
    & 
    \includegraphics[width=0.18\textwidth]{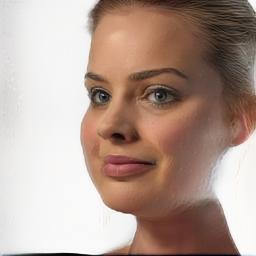}
    \\
\end{tabular}
\captionof{figure}{Additional qualitative evaluation on novel view rendering of yaw angle -0.6, -0.3, 0.3, and 0.6 radians.}
\label{supp:visualcomp7}
\vspace{-0.3cm}
\end{table*}

\begin{table*}[b!]
    \vspace{-0.3cm}
    \centering
    \setlength\tabcolsep{0pt}
        \begin{tabular}{cccccc}
            &
            Input
            &
            w/o $\mathcal{L}_m$
            &
            No $2^{nd}$ branch
            &
            $\mathcal{L}_m=0.5$
            &
            Ours
            \\
            \raisebox{1.5\height}{\rotatebox[origin=c]{90}{Same View}}\,\,
            &
            \includegraphics[width=0.18\textwidth]{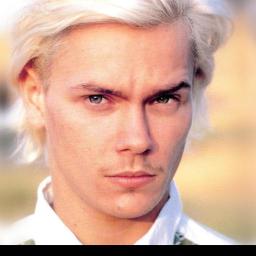}
            &
            \includegraphics[width=0.18\textwidth]{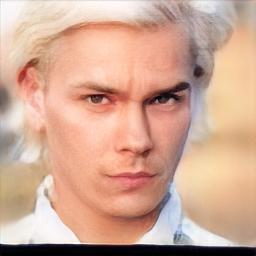}
            &
            \includegraphics[width=0.18\textwidth]{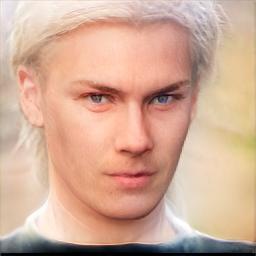}
            &
            \includegraphics[width=0.18\textwidth]{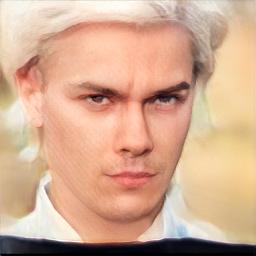}
            &
            \includegraphics[width=0.18\textwidth]{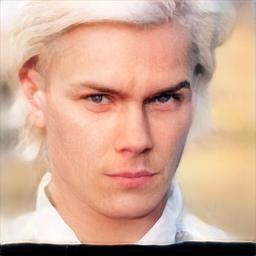}
            \\
            &
            \multicolumn{1}{r}{\raisebox{6.5\height}{\small Novel View}}\,\,
            &
            \includegraphics[width=0.18\textwidth]{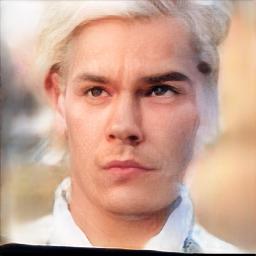}
            &
            \includegraphics[width=0.18\textwidth]{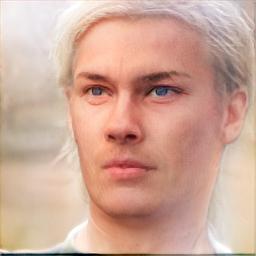}
            &
            \includegraphics[width=0.18\textwidth]{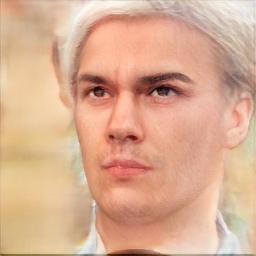}
            &
            \includegraphics[width=0.18\textwidth]{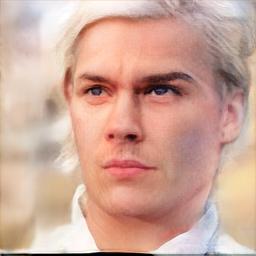}
            \\
\end{tabular}%
        \vspace{1cm}
        \begin{tabular}{cccccc}
            \raisebox{1.5\height}{\rotatebox[origin=c]{90}{Same View}}\,\,
            &
            \includegraphics[width=0.18\textwidth]{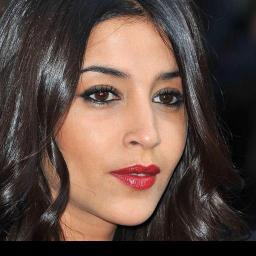}
            &
            \includegraphics[width=0.18\textwidth]{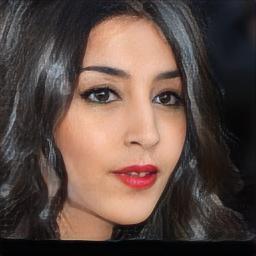}
            &
            \includegraphics[width=0.18\textwidth]{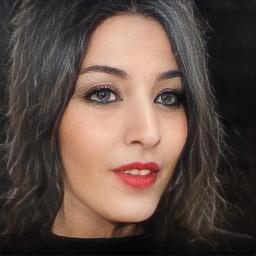}
            &
            \includegraphics[width=0.18\textwidth]{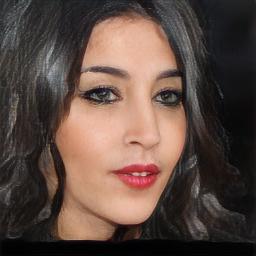}
            &
            \includegraphics[width=0.18\textwidth]{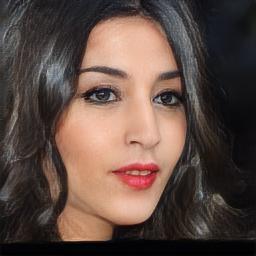}
            \\
            &
            \multicolumn{1}{r}{\raisebox{6.5\height}{\small Novel View}}\,\,
            &
            \includegraphics[width=0.18\textwidth]{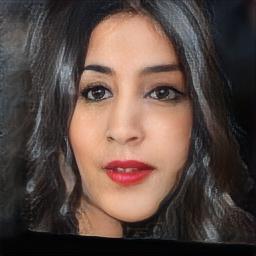}
            &
            \includegraphics[width=0.18\textwidth]{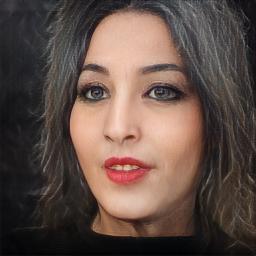}
            &
            \includegraphics[width=0.18\textwidth]{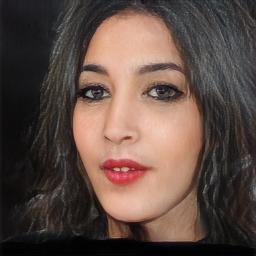}
            &
            \includegraphics[width=0.18\textwidth]{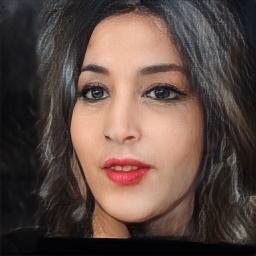}
            \\
            &
            \multicolumn{1}{r}{\raisebox{6.5\height}{\small Geometry}}\,\,
            &
            \includegraphics[width=0.18\textwidth]{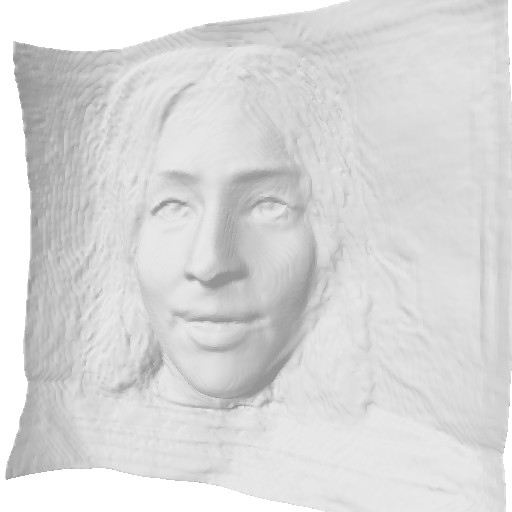}
            &
            \includegraphics[width=0.18\textwidth]{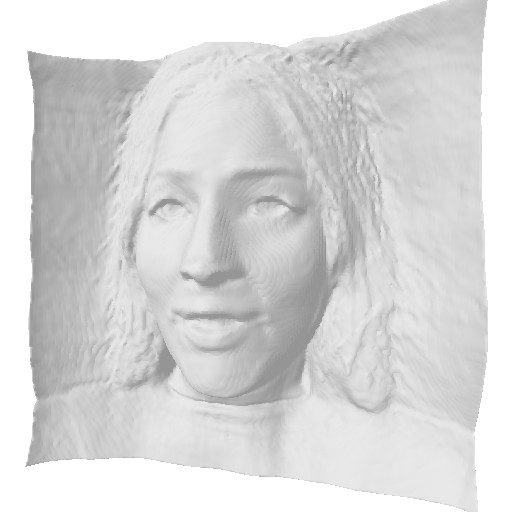}
            &
            \includegraphics[width=0.18\textwidth]{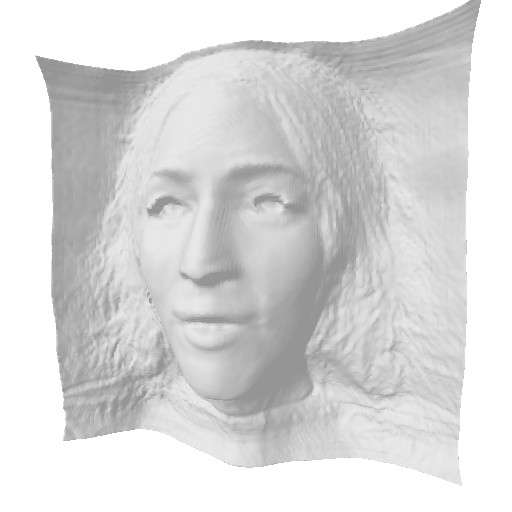}
            &
            \includegraphics[width=0.18\textwidth]{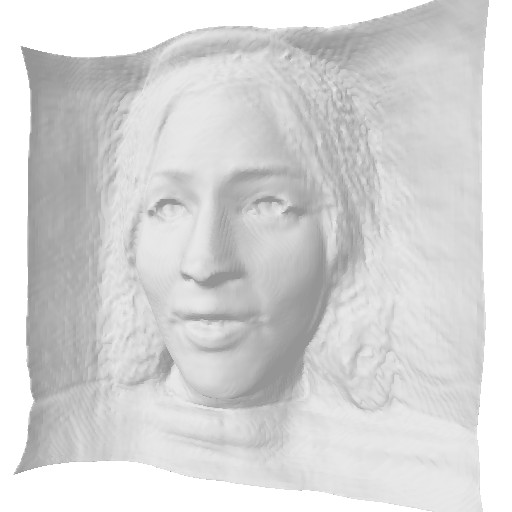}
            \\
        \end{tabular}%
\captionof{figure}{
        Additional qualitative ablation study for the loss, and architecture change.
    }
\label{fig:ablation-supp1}
\vspace{-0.3cm}
\end{table*}

\begin{table*}[b!]
    \vspace{-0.3cm}
    \centering
    \setlength\tabcolsep{0pt}
        \begin{tabular}{cccccc}
            &
            Input
            &
            w/o $\mathcal{L}_m$
            &
            No $2^{nd}$ branch
            &
            $\mathcal{L}_m=0.5$
            &
            Ours
            \\
            \raisebox{1.5\height}{\rotatebox[origin=c]{90}{Same View}}\,\,
            &
            \includegraphics[width=0.18\textwidth]{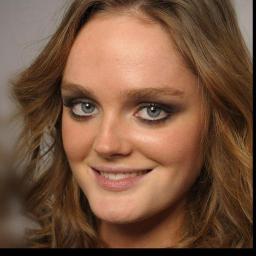}
            &
            \includegraphics[width=0.18\textwidth]{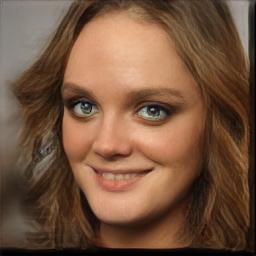}
            &
            \includegraphics[width=0.18\textwidth]{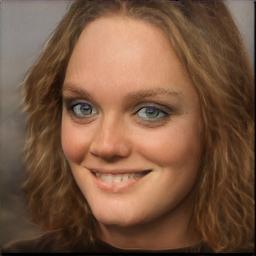}
            &
            \includegraphics[width=0.18\textwidth]{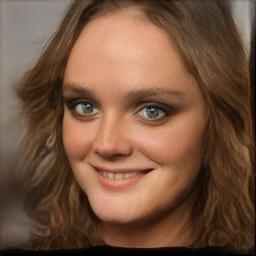}
            &
            \includegraphics[width=0.18\textwidth]{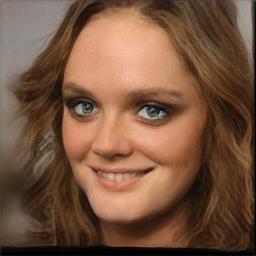}
            \\
            &
            \multicolumn{1}{r}{\raisebox{6.5\height}{\small Novel View}}\,\,
            &
            \includegraphics[width=0.18\textwidth]{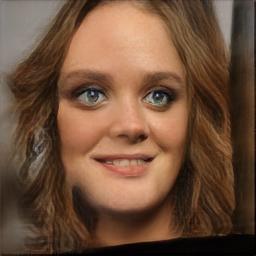}
            &
            \includegraphics[width=0.18\textwidth]{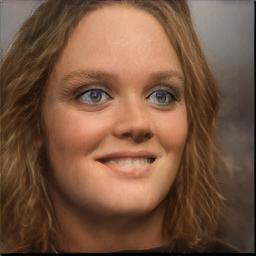}
            &
            \includegraphics[width=0.18\textwidth]{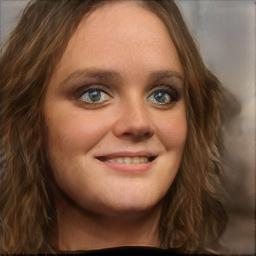}
            &
            \includegraphics[width=0.18\textwidth]{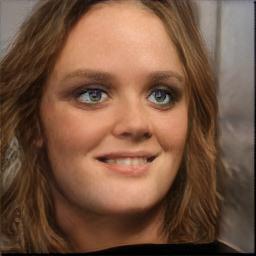}
            \\
\end{tabular}%
        \vspace{1cm}
        \begin{tabular}{cccccc}
            \raisebox{1.5\height}{\rotatebox[origin=c]{90}{Same View}}\,\,
            &
            \includegraphics[width=0.18\textwidth]{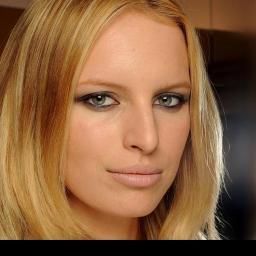}
            &
            \includegraphics[width=0.18\textwidth]{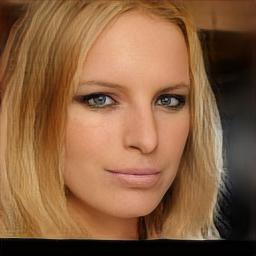}
            &
            \includegraphics[width=0.18\textwidth]{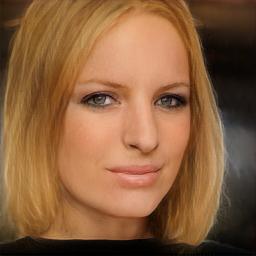}
            &
            \includegraphics[width=0.18\textwidth]{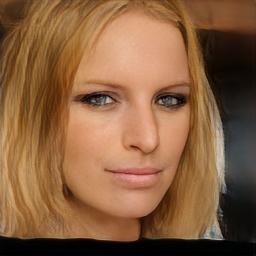}
            &
            \includegraphics[width=0.18\textwidth]{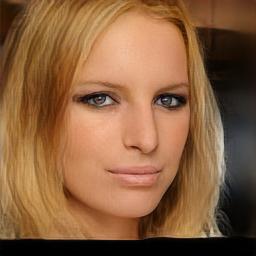}
            \\
            &
            \multicolumn{1}{r}{\raisebox{6.5\height}{\small Novel View}}\,\,
            &
            \includegraphics[width=0.18\textwidth]{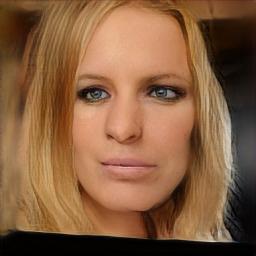}
            &
            \includegraphics[width=0.18\textwidth]{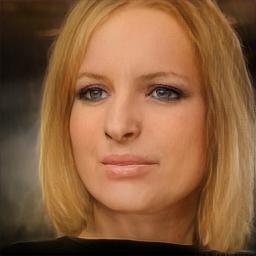}
            &
            \includegraphics[width=0.18\textwidth]{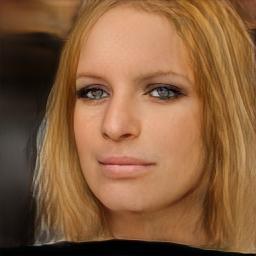}
            &
            \includegraphics[width=0.18\textwidth]{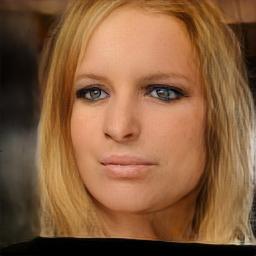}
            \\
            &
            \multicolumn{1}{r}{\raisebox{6.5\height}{\small Geometry}}\,\,
            &
            \includegraphics[width=0.18\textwidth]{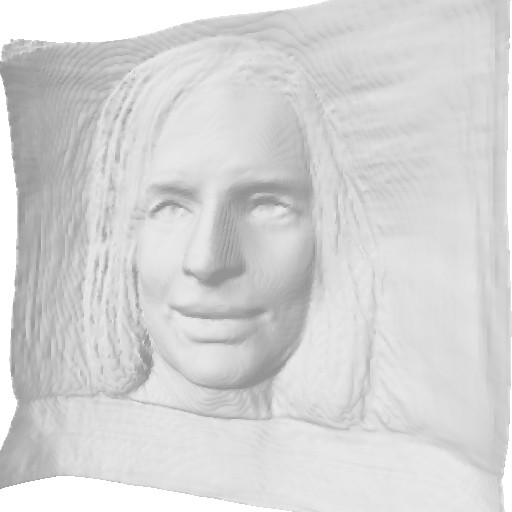}
            &
            \includegraphics[width=0.18\textwidth]{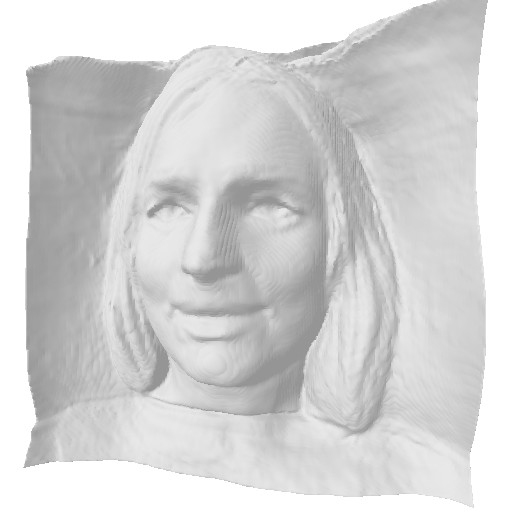}
            &
            \includegraphics[width=0.18\textwidth]{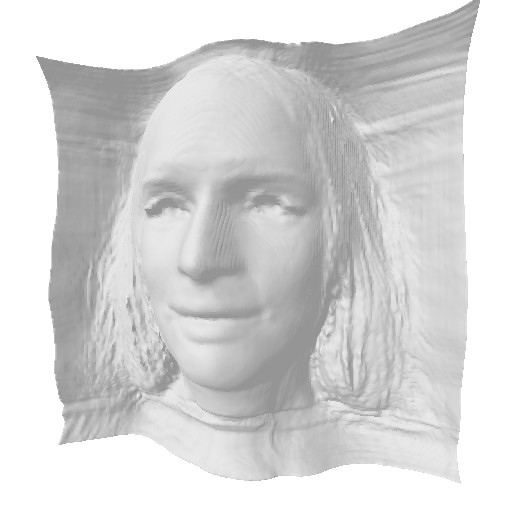}
            &
            \includegraphics[width=0.18\textwidth]{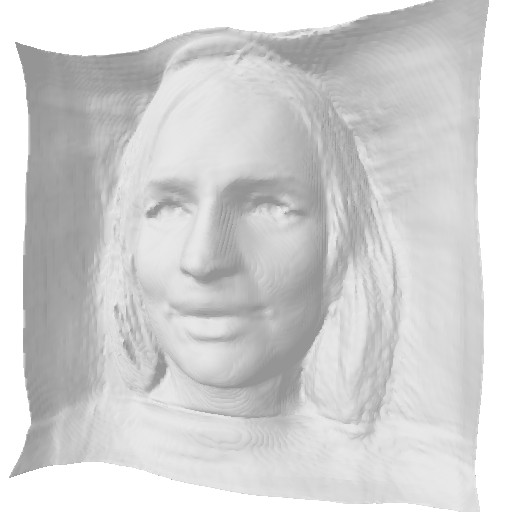}
            \\
        \end{tabular}%
\captionof{figure}{
        Additional qualitative ablation study for the loss, and architecture change.
    }
\label{fig:ablation-supp2}
\vspace{-0.3cm}
\end{table*}